\documentclass[11pt]{article}
\usepackage{fullpage}

\usepackage[utf8]{inputenc} 
\usepackage[T1]{fontenc}    
\usepackage{hyperref}       
\usepackage{url}            
\usepackage{booktabs}       
\usepackage{amsfonts}       
\usepackage{nicefrac}       
\usepackage{microtype}      
\usepackage{tcolorbox}
\usepackage{diagbox}

\usepackage{enumerate}
\usepackage[shortlabels]{enumitem}
\usepackage{subfig,float}
\usepackage[ruled]{algorithm2e}
\usepackage{graphicx} 
\usepackage{caption}
\usepackage{amsmath}
\usepackage{amsthm}
\usepackage{amssymb}
\usepackage{tikz}
\usepackage{tablefootnote}
\usepackage{multirow}
\usepackage{enumerate}
\usepackage{color}
\usepackage{xcolor}
\usepackage{natbib}
 \bibpunct[, ]{(}{)}{,}{a}{}{,}%

\usepackage{makecell}
\usepackage{tabularx}

\usetikzlibrary{arrows}

\allowdisplaybreaks[4]

\usepackage{mathrsfs}

\usepackage{hyperref}
\usepackage{bm,todonotes}

\allowdisplaybreaks

\newtheorem{theorem}{Theorem}[section]
\newtheorem{lemma}{Lemma}[section]
\newtheorem{corollary}{Corollary}[section]

\newtheorem{assumption}{Assumption}[section]

\newtheorem{remark}{Remark}[section]

\usepackage{authblk}

\definecolor{cvprblue}{rgb}{0.21,0.49,0.74}
\hypersetup{
    breaklinks=true,
    colorlinks=true,
    citecolor=cvprblue,
    linkcolor=purple,
    urlcolor=cyan,
}

\usepackage{xspace}
\def\rvc{{\mathbf{c}}}
\def\rvz{{\mathbf{z}}}
\def\rmC{{\mathbf{C}}}

\usepackage{import}
\usepackage{bbm}

\def\ours{\texttt{PPFL}\xspace}
\def\rmA{\mathbf{A}}

\def\rmU{\mathbf{U}}
\def\rmV{\mathbf{V}}

\def\BFW{\mathbf{W}}
\def\BFc{\mathbf{c}}
\def\BFC{\mathbf{C}}
\def\BFone{\mathbf{1}}
\def\BFx{\mathbf{x}}
\def\BFb{\mathbf{b}}
\def\BFL{\mathbf{L}}
\def\BFD{\mathbf{D}}
\def\BFI{\mathbf{I}}
\def\BFz{\mathbf{z}}
\def\BFs{\mathbf{s}}

\newcommand{\R}{\mathbb{R}}
\def\ppflone{\texttt{PPFL}1\xspace}
\def\ppfltwo{\texttt{PPFL}2\xspace}

\DeclareMathOperator*{\argmin}{\arg\min}

\def\Halmos{\mbox{\quad$\square$}}

\title{
\ours: A Personalized Federated Learning Framework for Heterogeneous Population
}

\author[1]{Hao Di}
\author[2]{Yi Yang}
\author[1]{Haishan Ye\thanks{Corresponding author. Email: \texttt{yehaishan@xjtu.edu.cn}}}
\author[1]{Xiangyu Chang}

\affil[1]{School of Management, Xi'an Jiaotong University}
\affil[2]{W.P. Carey School of Business, Arizona State University}

\begin{document}
\date{}
\maketitle
\stepcounter{footnote}
\footnotetext[2]{This paper has been accepted for publication in INFORMS Journal on Computing. DOI: \url{https://doi.org/10.1287/ijoc.2023.0376}}

\begin{abstract}%
Personalization aims to characterize individual preferences and is widely applied across many fields.
However, conventional personalized methods operate in a centralized manner, potentially exposing raw data when pooling individual information.
In this paper, with privacy considerations, we develop a flexible and interpretable personalized framework within the paradigm of federated learning, called \texttt{PPFL} (Population Personalized Federated Learning).
By leveraging ``canonical models" to capture fundamental characteristics of a heterogeneous population and employing ``membership vectors" to reveal clients' preferences, \texttt{PPFL} models heterogeneity as clients' varying preferences for these characteristics. 
This approach provides substantial insights into client characteristics, which are lacking in existing Personalized Federated Learning (PFL) methods.
Furthermore, we explore the relationship between \texttt{PPFL} and three main branches of PFL methods: clustered FL, multi-task PFL, and decoupling PFL, and demonstrate the advantages of \texttt{PPFL}.
To solve \texttt{PPFL} (a non-convex optimization problem with linear constraints), we propose a novel random block coordinate descent algorithm and establish its convergence properties.
We conduct experiments on both pathological and practical data sets, and the results validate the effectiveness of \texttt{PPFL}.
\end{abstract}

\section{Introduction}\label{sec:introduction}

Personalization is based on customizing models or algorithms to fit individuals' traits and preferences. 
It has been widely applied in fields such as marketing \citep{smith2022optimal}, precision medicine \citep{bonkhoff2022precision}, and engineering processes \citep{lin2017collaborative} to offer tailored products or services to heterogeneous clients based on the information collected from them.

An intuitive approach to achieve personalization for heterogeneous clients is to estimate models individually, which requires sufficient data.
However, in real-world scenarios, data may be unevenly distributed among clients, with some having a large volume of data while others may have only scarce data.
As a result, it is challenging to build an accurate model based on measurements from a single client alone.
Hence, conventional personalization methods, like the Mixed-Effect Model (MEM) \citep{verbeke1996linear} and Mixture of Experts (MoE) \citep{feffer2018mixture}, are initiated in a centralized manner, where data are centrally collected, stored, and processed. 

However, in most industries, data exist as isolated islands \citep{tan2022towards}. 
A lack of interconnection between data owners makes the collection and centralized storage of data time-consuming and expensive. 
Moreover, these data are often deemed sensitive and regarded as a source of confidential and proprietary information \citep{yang2020privacy}, making data aggregation less feasible in practice.
From a legal perspective, relevant laws and regulations, such as the General Data Protection Regulation in the European Union, set rules restricting data sharing and storage. 
Therefore, learning personalized models while keeping the training data localized becomes a crucial problem.

\begin{figure}
\centering
\includegraphics[scale=0.28]{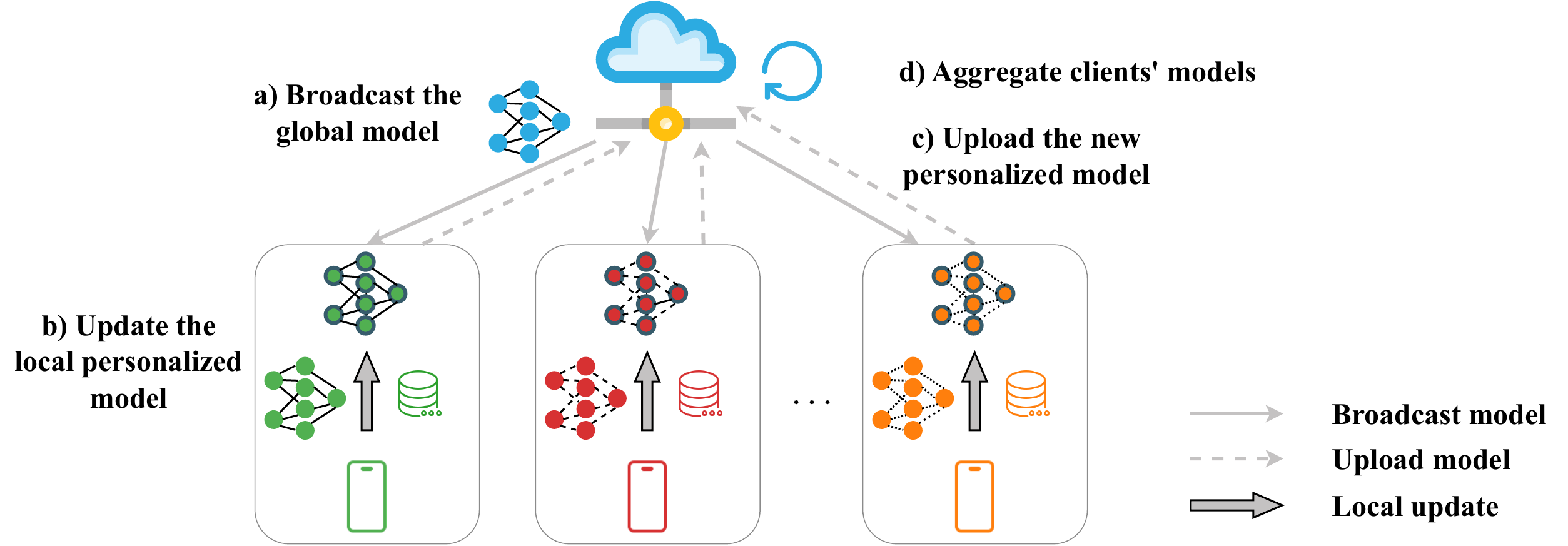}
\caption{A typical personalized federated learning diagram. 
}
\label{fig:diagram of pfl}
\end{figure}

To enable collaborative training of machine learning models among distributed clients without exchanging or centralizing their raw data, a new learning paradigm named Federated Learning (FL) is introduced by \citet{mcmahan2017communication}. 
Instead of collecting data from clients, in the FL framework, the central server aggregates models learned by each client and broadcasts a global model for clients.
However, real-world applications often face a rather heterogeneous population, where each client has diverse underlying patterns~\citep{lin2017collaborative}. 
Therefore, learning a single global model may introduce significant bias when serving clients with heterogeneous data distributions~\citep{hanzely2023personalized}. 
To fill this gap, researchers investigate incorporating personalization into the FL framework, referred to as personalized federated learning (PFL), and consider it a viable solution to address the challenges of heterogeneous populations in FL \citep{li2020federated}.
In the typical PFL paradigm, as illustrated in Figure \ref{fig:diagram of pfl}, each client maintains a personalized model locally, and the central server orchestrates the collaboration among clients as in the vanilla FL: a) At the beginning of each round, the central server broadcasts information (e.g., the global model) to clients to guide their personalized models; b) Clients update their local models using stored data and received information; c) After completing the local update iteration, clients transmit the updated models back to the central server; and d) The central server aggregates models to generate the next round's broadcast information.

\subsection{Related Work}
Assuming there are $M$ clients in the system, the vanilla FL approach \citep{mcmahan2017communication} aims to minimize the weighted sum of each client's empirical loss, i.e., 
$
    \min_{\theta} \sum_{i=1}^M p_i f_i(\theta),
$
where $f_i(\theta)$ represents the empirical loss of the $i$-th client, $\theta$ is the model parameter, and $p_i$ denotes the weight (importance) assigned to the $i$-th client's loss.
To construct personalized models and mitigate the heterogeneity in FL, there are three main streams of PFL methods related to this work.

\textbf{Clustered FL}. Clustered FL is a widely discussed approach that assumes clients belong to different clusters based on their underlying behavior patterns.
Hence, it classifies clients into different clusters and provides personalized models at the group level. 
The process mainly comprises two primary steps: client identification and personalized model optimization. 
Client identification aims to assign each client to the most suitable cluster. Then, personalized model optimization provides the optimal solution for each cluster.
Generally, the objective function of clustered FL is given as follows:
\begin{equation}\label{eq1}
    \min_{\theta_1, \dots, \theta_K} \sum_{k=1}^K \sum_{i\in \mathcal{S}_k} p_i f_i(\theta_k),  
\end{equation}
where $\theta_k$ and $\mathcal{S}_k$ represent the personalized model for the $k$-th cluster and the collection of clients belonging to it, respectively.
Existing literature has focused on client identification through diverse approaches, including analyzing the variance in uploaded gradients \citep{sattler2020clustered}, assessing the similarity of clients' feature maps \citep{liu2021pfa}, and evaluating empirical losses \citep{ghosh2020efficient}. Additionally, some researchers propose classifying each local data point rather than each client \citep{marfoq2021federated}.

Although the main advantage of clustered FL is that it explicitly indicates group information and enhances the interpretability of the population's heterogeneous behaviors, the random partitioning of clients into clusters and the non-differentiable nature of the cluster identification step pose significant challenges in guaranteeing convergence \citep{ghosh2020efficient}. 
Moreover, the cluster identification process requires the server to send all $K$ models to each client at every broadcasting round, allowing them to evaluate and select the most suitable model for local training and updates. 
Consequently, the communication cost scales as $\mathcal{O}(Kd)$ ($d$ is the model dimension), which is $K$ times higher than that of vanilla FL.

\textbf{Multi-task PFL}. In this method, clients' models are trained on their local data sets, and the server sends information to clients to guide the development of their personalized models.
Consequently, this approach corresponds to the following optimization problem:
\begin{equation}
    \label{eq:multi-task PFL}
    \min_{\theta_1, \dots, \theta_M, \BFW} \sum_{i=1}^M p_i f_i(\theta_i) + \mathcal{R}(\Theta, \BFW),
\end{equation}
where $\theta_i$ represents the personalized model in the $i$-th client, $\Theta:=\{\theta_i\}_{i=1}^M$ denotes the collection of all personalized models, $\BFW$ measures the relationships among clients, and $\mathcal{R}$ is a regularization function which plays a crucial role in establishing collaboration mechanisms.
For example, \citet{t2020personalized} and \citet{li2021ditto} employ $\ell_2$ regularization with $\BFW$ as the identity matrix, i.e., $\mathcal{R}(\Theta, \BFW):=\sum_{i=1}^M\Vert \theta_i - \theta\Vert^2$, to encourage local personalized models to approach the global model $\theta$ ($\theta := \sum_{i=1}^M p_i \theta_i$).
Besides, Laplacian regularization, $\mathcal{R}(\Theta, \BFW):= \sum_{i=1}^M\sum_{j=1}^M w_{ij}\Vert\theta_i - \theta_j \Vert^2$, is also widely adopted \citep{smith2017federated} to promote pairwise collaboration among clients with similar local distributions. 

Although multi-task PFL is straightforward, using regularization to refine local model updates, the method provides limited insight into the heterogeneous client behaviors compared to Clustered FL. 
Moreover, since inadequate local data leads to performance degradation~\citep{kamp2023federated}, this method becomes less effective when many clients have small local data sets.

\textbf{Decoupling PFL}. 
Decoupled PFL splits the base model into two components: locally retained parameters that capture local data distributions and global parameters shared across clients.
The objective function is represented as follows: 
\begin{equation}
\label{eq:decoupling approach}
    \min_{\theta, v_1, \dots, v_M} \sum_{i=1}^M p_i f_i(\theta, v_i),
\end{equation}
where $v_i$ denotes the private component of the $i$-th client, and $\theta$ represents the remaining shared model.
In this approach, various strategies are proposed for segmenting local private layers and the common global module, such as the local representation module \citep{Liang2020ThinkLA}, the local head layers \citep{collins2021exploiting}, and the local private adapter \citep{pillutla2022federated}.
In addition, several studies investigate the use of multiple global representations to address client heterogeneity in the contexts of mixed-domain linear regression \citep{zhongfeddar} and online bandit problems \citep{kosolwattana2024fcom}.

Since the personalized component is not engaged in the communication, the decoupling PFL only transmits the remaining global component and thus results in lower communication cost. 
However, the absence of personalized parameters on the server side obscures the identification of characteristics and patterns exhibited across populations and hinders post hoc analysis \citep{murdoch2019definitions}.
Moreover, the lack of collaboration in the personalized component may impair its generalization performance. 
Beyond the three main PFL methods discussed above, other approaches, such as Bayesian PFL, data augmentation, and client selection are discussed in Section O3.2 in the Online Supplement.
Overall, based on the discussions, an appropriate PFL framework should ideally possess the following three properties:
\begin{itemize}
    \item {\textbf{Interpretability:}
    The framework could offer interpretability that reveals clients' similarities and differences to understand the underlying patterns of heterogeneous populations. 
    Interpretability is crucial in PFL, especially in fields like healthcare \citep{dayan2021federated}, where FL methods are actively explored. 
    }
    \item {\textbf{Flexibility:} The framework could integrate the strengths of several existing PFL methods (e.g., clustered FL, multi-task PFL, and decoupling PFL) by strategically configuring its architecture and be implementing it using standard machine learning models.}
    \item {\textbf{Efficient Algorithms with Theoretical Guarantees:}
    The framework could offer an efficient algorithm with at least a sub-linear convergence rate, similar to prevalent PFL methods such as decoupling PFL, under non-convex optimization settings. 
    These efficient algorithms could address the challenge of complex sampling algorithms in Bayesian PFL and retain more local data for data-based PFL. 
    }
\end{itemize} 

\subsection{Contribution}

We propose a novel personalized federated learning framework called \texttt{PPFL} (Population Personalized Federated Learning), which possesses the advantages mentioned above. 
\texttt{PPFL} is built on the critical assumption that heterogeneous behavior patterns can be encoded by a set of latent models.
Notably, latent models, widely accepted for handling such heterogeneous patterns, have been widely used in the centralized setting, e.g., \citet{feng2020learning} for linear models and \citet{kim2022integrating} for Bayesian models.
Inspired by these works, \ours leverages \textit{canonical models}, a class of latent models, to represent the heterogeneous population characteristics, where each canonical model corresponds to a distinct behavior pattern.

Note that a client may exhibit a mixture of these patterns.
Thus, we introduce a \textit{membership vector} $\BFc_i=(c_{i1},c_{i2},\dots,c_{iK})^\top$ for the $i$-th client, where $c_{ik}$ quantifies the degree to which the $k$-th canonical model can represent the $i$-th client's characteristics.
Mathematically, these membership vectors and canonical models can be regarded as coefficients and a basis system, respectively, for spanning the modeling space of heterogeneous populations.

Although introducing membership vectors helps build personalized models, it complicates the optimization process of \ours and results in a constrained optimization problem, as these vectors are must lie within the simplex space, i.e., $\sum_{k=1}^K c_{ik} = 1$ and $c_{ik} \geq 0$ for all $i$.
Furthermore, including the membership vectors $\BFc_i$ in the PFL setting results in a non-convex objective function that fails to satisfy the jointly convex property \citep{smith2017federated}, making simultaneous optimization of all variables challenging.
In addition, similar to existing latent modeling works \citep{feng2020learning,lin2018selective}, the Laplacian regularization is enforced on the membership vectors to facilitate the collaboration among clients with similar data distributions and improve the generalization ability.
However, the non-separable Laplacian regularization couples the membership vectors, making their local updates more complex.
To overcome these hurdles, a federated version of the random block coordinate descent (RBCD) algorithm is employed, along with its convergence analysis in the non-convex setting, to estimate the membership vectors and canonical models efficiently.
Finally, to validate the convergence properties and showcase the advantages of \ours, we perform experiments on both synthetic and real-world data sets, covering pathological and practical scenarios.
Our contributions can be summarized as follows:

\begin{itemize}
   \item \textbf{Interpretability}: The \texttt{PPFL} framework introduces the canonical model and membership vectors into the FL paradigm for handling heterogeneous population patterns, which can learn the membership vectors and canonical models simultaneously.
    Compared to the clustered FL, \texttt{PPFL} eliminates the need for the client identification step, significantly reducing computational complexity.
    \item \textbf{Flexibility}: The \texttt{PPFL} framework allows a flexible implementation that can incorporate clustered FL, multi-task PFL, and decoupling PFL. 
    Additionally, \texttt{PPFL} supports two different structures for personalization.
    These two structures effectively extend collaborative learning~\citep{lin2017collaborative,feng2020learning} and provide another form of MoE by introducing flexible integration and a combination of canonical models within the FL setting, further demonstrating \ours's versatility.

    \item \textbf{Algorithms with Theoretical Guarantees}: 
    We propose a novel RBCD algorithm to solve the constrained non-convex optimization problem in \ours.
    Compared to the clustered FL, decoupling PFL, and collaborative learning, which lack convergence rate guarantees,
    the proposed RBCD algorithm can achieve a sub-linear convergence rate ($\mathcal{O}(\frac{1}{\sqrt{T}})$) if using full gradient information. 
    To our knowledge, this is the first result of RBCD algorithms in FL for non-convex optimization problems with linear constraints.
\end{itemize}

\section{Population Personalized Federated Learning}
\label{sec:method}

\subsection{Formulation of \texttt{PPFL}}\label{subsec:formulation}
\begin{figure}[t]
  \caption{Two possible structures of \texttt{PPFL}, where $\theta_{\text{com}}$ is used as an extractor for illustration.
(a) The $K$ canonical models process the input from $\theta_{\text{com}}$ in parallel and a weighted sum of their outputs is computed;
(b) A personalized model $\theta_i=\sum_{k=1}^K c_{ik}\theta_k$ is constructed via a linear combination of canonical models, which then processes the input from $\theta_{\text{com}}$.}
    \centering
    \subfloat[]{\label{fig:form 1}\includegraphics[scale=0.5]{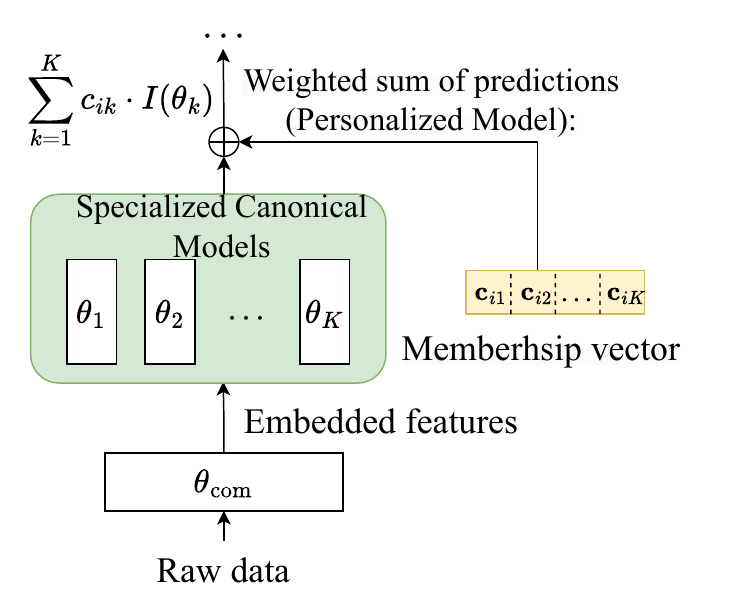}}\hspace{5pt}
    \subfloat[]{\label{fig:form 2}\includegraphics[scale=0.5]{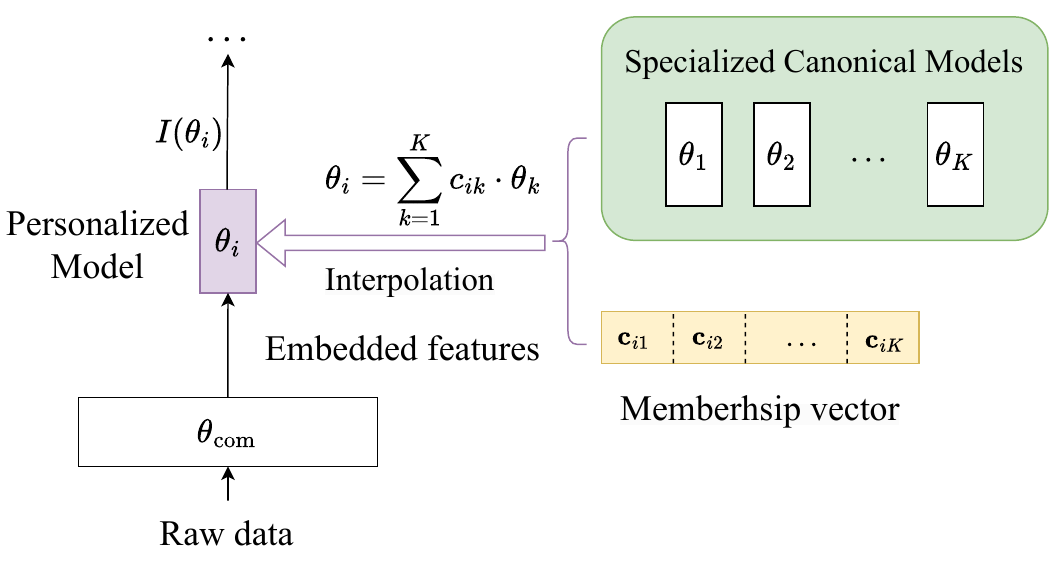}}
    \label{fig:structure}
\end{figure}

We consider an FL system with $M$ clients indexed by $i$, and $[M]$ denotes the index set $\{1,2,\dots,M\}$.
Assuming there are $K$ distinct behavior patterns among clients, we introduce the following \ours framework:
\begin{align}
    \label{eq:origin problem}
\min_{\theta_{\text{com}}, \{\theta_k\}_{k=1}^K,\{\BFc_i\}_{i=1}^M} \quad & \sum_{i=1}^M p_i f_i(\theta_{\text{com}},\{\theta_k\}_{k=1}^K,\BFc_i) + \frac{\lambda}{2}\sum_{i=1}^M\sum_{j=1}^M w_{ij}\Vert\BFc_i -\BFc_j \Vert^2,\\ \label{Eq:constrain1}
\textrm{s.t.} \quad & \BFone^\top\BFc_i=1,\; \forall i\in [M],\\ \label{Eq:constrain2}
  &\BFc_i \geq 0,\; \forall i \in [M],
\end{align}
where $w_{ij}$ is an element of the affinity matrix $\BFW \in \mathbb{R}^{M\times M}$, and $\lambda\geq 0$ serves as a hyperparameter.
In Eq. (\ref{eq:origin problem}), the objective function contains three distinct model parameters: $\theta_{\text{com}}$, $\{\theta_{k}\}_{k=1}^{K}$, and $\{\BFc_i\}_{i=1}^M$.
We will elucidate each of them individually.
\begin{itemize}
    \item The \ours framework introduces $K$ distinct canonical models $\{\theta_k\}_{k=1}^K$, which are shared across clients.
      As shown in Figure \ref{fig:structure}, these $K$ canonical models operate in parallel: accepting the same input (embedded features from $\theta_{\text{com}}$ or raw data) but producing different outputs due to their distinct parameters.
      The canonical models are designed to specialize in and represent distinct characteristics or behavior patterns present in the input data.
    \item The \ours framework utilizes the shared global parameter $\theta_{\text{com}}$ to extract common representations and generate generalizable predictions, similar to the global module in decoupling PFL approaches.
      As depicted in Figure \ref{fig:structure}, $\theta_\text{com}$ is shared across the $K$ distinct canonical models $\{\theta_k\}_{k=1}^K$, ensuring that all canonical models receive the same embedded feature distribution to promote the development of diverse specialized canonical models.
    \item The \ours framework employs a $K$-dimensional membership vector $\BFc_i$ for each client, whose elements reflect the degree to which canonical models can represent the client. 
    Moreover, we require the membership vector $\BFc_i$ to reside in a simplex space for interpretability (see Eqs. (\ref{Eq:constrain1}) and (\ref{Eq:constrain2})).
\end{itemize}

Based on this framework, we can assess the characteristics of each client and derive insights into the heterogeneous population through the membership vector, where an element with a higher value indicates a stronger similarity between the client's behavior and the corresponding captured characteristics.
Furthermore, by incorporating the shared global module $\theta_{\text{com}}$ and adopting a decoupling design akin to decoupling PFL methods, \ours can alleviate local data scarcity at certain individual clients \citep{Liang2020ThinkLA}.
Moreover, since the size of the canonical model is small, the communication cost increases at a slower rate than the linear growth in clustered FL.
Laplacian regularization encourages the membership vectors of clients with similar local distributions to be close to each other and captures intrinsic relationships among clients.

As shown in Figures \ref{fig:form 1} and \ref{fig:form 2}, using the feature extractor $\theta_{\text{com}}$ (e.g., convolution layers) as an illustrative example, \ours offers flexible implementation and supports two potential architectures.
In the first architecture (Figure~\ref{fig:form 1}), the weighted sum $\sum_{k=1}^K \BFc_{ik} I(\theta_k)$ is computed by multiplying each canonical model's output $I(\theta_k)$ by $c_{ik}$, then passed into the subsequent layer.
Meanwhile, the second architecture in Figure~\ref{fig:form 2} manipulates the parameter space and operates by constructing personalized parameters ${\theta}_i = \sum_{k=1}^K c_{ik} \theta_k$ through a linear combination before applying $I(\cdot)$.
These two architectures differ in how they combine client-specific parameters $\mathbf{c}_i$ and canonical models $\theta_k$.
As discussed in Section O3.1 in the Online Supplement, with an appropriate choice of the hyperparameter $\lambda$ and architecture, the proposed method can establish connections with various existing PFL methods.

\subsection{Implementation of \texttt{PPFL} with Different Canonical Models}\label{subsec:implementation}
In the following paragraphs, we implement \texttt{PPFL} for two widely used models: Generalized Linear Model (GLM) and Deep Neural Networks (DNNs). 
Subsequently, we discuss connections to prominent centralized personalized methods specific for these two models, demonstrating \texttt{PPFL}'s flexibility.

\subsubsection{Generalized Linear Model} \label{sec:glm} 
In the GLM, the response variable $y$ is assumed to follow a specific distribution from the exponential family and is related to the feature vector $\BFx\in \mathbb{R}^{d}$ through a link function $g$, and the expectation of $y$ is $ \mathbb{E}[y] = g^{-1}(\BFx^\top \mathbf{b}).$

Due to the simple formulation of GLM, the parameter $\theta_{\text{com}}$ is dispensable when adopting \ours to develop personalized GLMs.
Based on the architectures in Figure~\ref{fig:form 1} and~\ref{fig:form 2}, given the membership vector and $\{\theta_k\}_{k=1}^K$, we derive two potential formulations for the prediction of the $l$-th data point in client $i$ as follows:

\noindent\begin{minipage}{0.4\textwidth}
\begin{equation}
\hat{y}_{il} = \sum\nolimits_{k=1}\nolimits^K c_{ik} g^{-1}(\BFx_{il}^\top\theta_k)\label{eq:GLM form1}
\end{equation} 
    \end{minipage}%
    \begin{minipage}{0.2\textwidth}\vspace{0.2cm}\centering
    and 
    \end{minipage}%
    \begin{minipage}{0.4\textwidth}
    \vspace{-0.15cm}
\begin{equation}
\hat{y}_{il} = g^{-1}(\BFx_{il}^\top\theta\BFc_i)\label{eq:GLM form2},
\end{equation}
    \end{minipage}\vskip1em
\noindent
where $\theta=[\theta_1,\theta_2,\dots,\theta_K]\in \mathbb{R}^{d\times K}$ is a stacked matrix of all canonical models. 
    
Using either Eqs. \eqref{eq:GLM form1} or \eqref{eq:GLM form2}, we can apply \texttt{PPFL} for GLM personalization. 
Moreover, the problem formulation of \texttt{PPFL} is also connected to other personalized GLM variants, such as collaborative learning \citep{lin2017collaborative, feng2020learning} and the widely used General Linear Mixed Model (GLMM). 
In Section O3.4 in the Online Supplement, we provide detailed formulations of GLM for \ours and \citet{feng2020learning}, contrasting these approaches to highlight the flexibility of \ours.

Compared to existing studies \citep{lin2017collaborative, feng2020learning, kim2022integrating} employing latent models, \ours fundamentally differs in its distributed nature, which involves storing and processing clients' data locally rather than centralizing it on the central server.
This characteristic significantly influences the algorithm's design and subsequent convergence analysis, making it infeasible to derive a closed-form solution as in \citet{lin2017collaborative}.
Therefore, a novel RBCD-based optimization method is developed in \ours, which offers an explicit convergence rate in contrast to a stationary point guarantee in \citet{lin2017collaborative}.
Compared to \citet{kosolwattana2024fcom}, which utilizes canonical models for online bandit problems with $\ell_2$ regularization, \ours leverages Laplacian regularization to effectively fuse data from similar clients \citep{lin2017collaborative}, facilitating the development of personalized models.
Furthermore, we demonstrate that \ours can accommodate more complex models, including deep models (as shown in Section \ref{sec:experiment}), extending beyond the linear models typically used in previous studies.

\subsubsection{General Linear Mixed Model}
GLMM, a well-known personalized variant of GLM, incorporates both fixed effects $\mathbf{b}_f$ and random effects $\mathbf{b}_r$ in the predictor $
    \mathbb{E}[y] = g^{-1}(\BFx^\top \BFb_f+\BFx^\top \BFb_r),
$
where $\BFx\in \mathbb{R}^{d}$ is the feature vector.
The fixed effect $\mathbf{b}_f$, inherited from GLM, captures the tendency among the population.
The random effect $\mathbf{b}_r$, typically assumed to follow a zero-mean multivariate Gaussian distribution with a learnable covariance matrix $G$, captures the heterogeneity among individuals and varies across clients.

To explicitly demonstrate the connection between \ours and GLMM, we show that maximizing the log-likelihood of GLMM can achieve the same objective function as \ours, as justified in Theorem \ref{thm:relationship with MEM}, and the detailed proof is provided in Section O1.2 in the Online Supplement.

\begin{theorem}
\label{thm:relationship with MEM}
Suppose that the random effects $\mathbf{b}_r$ are generated from a normal distribution $\mathcal{N}(0,G)$, then the objective function of \texttt{PPFL} in Eq. \eqref{eq:origin problem} is equivalent to the objective function of GLMM, when $G=\theta\theta^\top$, all elements of $\BFW$ are equal to $1$, and $p_i = \frac{n_i}{n}$ with $n_i$ being the size of the local data set and $n=\sum_{i=1}^M n_i$.
\end{theorem}

\begin{remark}
   Theorem \ref{thm:relationship with MEM} provides a Bayesian perspective to interpret the \texttt{PPFL} method: the objective function of GLMM is equivalent to that of \ours, learning the canonical models $\{\theta_k\}_{k=1}^K$ is equivalent to learning the shared covariance matrix $G$ in GLMM. 
    This implies that these canonical models can serve the function of borrowing statistical knowledge across clients, similar to the role of hierarchical models \citep{yue2024federated}.
    Furthermore, the membership vector $\rvc$ can also serve as the random effect, and details are provided in Section O1.2 in the Online Supplement.
    Besides, instead of the homogeneous scenario in GLMM (the random variables $\mathbf{b}_r$ of clients follow the same normal distribution $\mathcal{N}(0, G)$, corresponding to an all-one matrix $\mathbf{W}$ in \texttt{PPFL}), the similarity matrix $\mathbf{W}$ in \texttt{PPFL} can take different values to represent more general scenarios.
    \end{remark}

Notably, \citet{lin2017collaborative} and \citet{feng2020learning} demonstrate equivalence between their methods and the linear mixed model and logistic mixed model, respectively, under the conditions outlined in Theorem \ref{thm:relationship with MEM}. 
We extend results to the general GLMM problem, establishing the equivalence in this common setting. Furthermore, we also justify the rationale behind setting $G=\theta\theta^\top$. 
A detailed discussion of these aspects can be found in Section O1.2 in the Online Supplement.
\begin{figure}[t]
    \caption{The diagram of MoE. The gate network generates probability weights over these $K$ experts.}
    \centering
    \includegraphics[scale=0.5]{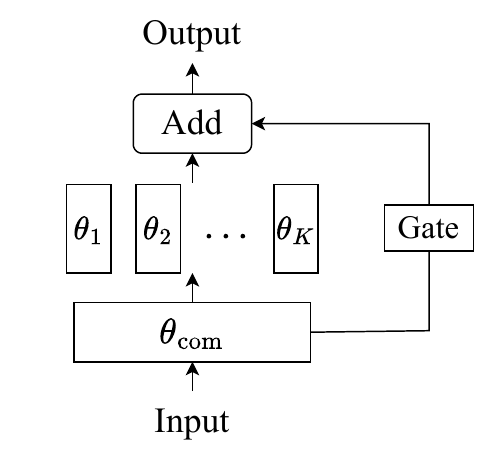}
    \label{fig:MoE}
\end{figure}

\subsubsection{Deep Neural Networks}
\label{sec:nn}

DNNs are widely adopted across multiple industries due to their ability to automatically extract features and achieve strong performance.
As shown in Figures~\ref{fig:form 1} and~\ref{fig:form 2}, \ours can be seamlessly integrated into DNNs, where the layers/modules can be regarded as canonical models.
Notably, the structure of \texttt{PPFL} illustrated in Figure~\ref{fig:form 1} is similar to that of MoE \citep{shazeer2017outrageously} shown in Figure \ref{fig:MoE}.
The MoE module comprises multiple parallel expert networks (denoted as $\theta_k$ for simplicity) and a gate network (typically a learnable softmax function).
As illustrated in Figure \ref{fig:MoE}, input features are processed by both the expert networks and the gate network.
Each expert network $\theta_k$ computes its prediction $f_k(\theta_k; x_k)$, while the gate network outputs probability weights over the $K$ experts.
Hence, the final output of the MoE module is a weighted sum of the predictions from the $K$ parallel expert networks, i.e., $
    \sum_{k=1}^K \text{Softmax}(\theta_g\mathbf{x})_k \cdot f_k(\theta_k; x_k)$, 
where $\theta_g$ is the learnable parameters of the gate network.
While MoE and \ours are designed to specialize in different underlying subgroups, a key distinction lies in their weight vector calculations, reflecting different underlying ideas behind these two approaches.

Specifically, in MoE, the $K$-dimensional weight vector is calculated based on the input data point $\mathbf{x}$ through the gate function, i.e., $\text{Softmax}(\theta_g\mathbf{x})$.
In contrast, the membership vector $\mathbf{c}_i$ is independent of the input data point and treated as a client attribute in \ours.
In summary, MoE utilizes the gate network to route each data point to the most relevant prototype, while \texttt{PPFL} designs the canonical models and membership vectors to capture the clients' characteristics.
Consequently, \ours provides enhanced interpretability compared to MoE, as its membership vectors directly reveal clients’ preferences.
Moreover, in addition to the linear combination of predictions as in MoE, \ours can also implement a composite personalized model by spanning these canonical models as presented in Figure \ref{fig:form 2}, potentially enabling richer representations and demonstrating the flexibility of the method.

\section{Algorithm for \ours}
\label{section: algorithm}

This section will design an efficient algorithm to solve \texttt{PPFL} and analyze its convergence property.

\subsection{Algorithm Design}
Let $\theta$ be the combination of the shared parameters $\theta_{\text{com}}$ and $\{\theta_k\}_{k=1}^K$, $\BFC =[\BFc_1^\top,\dots,\BFc_M^\top]^\top \in \mathbb{R}^{KM}$ denote the concatenated membership vectors, $\BFL = (\BFD- \BFW) \otimes \BFI \in \mathbb{R}^{KM \times KM}$ be a Laplacian matrix ($\otimes$ denotes the Kronecker product), and $\BFD$ be a diagonal matrix with the elements $d_{ii} = \sum_{j=1}^M w_{ij}$.
Let $\rvz=\{\theta, \rmC\}$ denote the collection of model parameters, indexed by $h\in\{1, 2\}$, such that $\rvz_1=\theta$ and $\rvz_2=\rmC$.
We reformulate \texttt{PPFL} framework in Eq.~\eqref{eq:origin problem} as
\begin{equation}
\label{eq:reformulated problem}
\begin{aligned}
\min_{\theta, \BFC} \quad & F(\theta,\BFC)=\sum_{i=1}^M\, p_i f_i(\theta, \BFc_i) + \lambda \BFC^\top\BFL\BFC, \quad\quad
\textrm{s.t.} \quad \BFC \in \mathcal{C},
\end{aligned}
\end{equation}
where the feasible set $\mathcal{C}\subset \mathbb{R}^{KM}$ corresponds to the Cartesian product of $M$ simplex spaces.

There are three key challenges in addressing the above optimization problem.
First, the jointly convex property of $F$ (convexity in both $\theta$ and $\BFC$) is a strong condition and hardly guaranteed even for the commonly used basic models (e.g., GLM), making simultaneous optimization of $\theta$ and $\BFC$ impractical \citep{smith2017federated}.
Second, the local loss function $f_i$ contains its membership vector $\BFc_i$ restricted to a simplex space, which results in a constrained optimization in an FL setting. 
Third, the non-separable Laplacian regularization usually complicates the process since it involves dependencies on other membership vectors.

To cope with these issues, we propose an RBCD algorithm in the FL setting, outlined in Algorithm~\ref{alg:ppfl algorithm}, which decouples variables and optimizes one block at a time while fixing the others. 
Thus, the proposed algorithm can efficiently solve the optimization problem with multiple variables without the jointly convex property. 

\begin{algorithm}[t]
\caption{RBCD-based PPFL algorithm.}
\label{alg:ppfl algorithm}
\SetKwFunction{LG}{Local Gradient$(\BFz_h^t, \eta_{h, t})$}
\SetKwBlock{LGtheta}{\textbf{Local Gradient}$(\theta^t, \eta_t)$}{}
\SetKwBlock{LGC}{\textbf{Local Gradient}$(\BFC^t, \eta_t)$}{}
  \SetAlgoLined
  \KwIn{the initial state $\BFz^{0}$, the number of canonical models $K$, the number of communication rounds $T$, the number of local steps $E$, the step sizes $\{\eta_t\}_{T}$, $\eta_{1, t} = \eta_t / E$, $\eta_{2, t} = \eta_t$, the Laplacian matrix $\BFL$ with its hyperparameter $\lambda$, the smoothness parameters $L_1$ and $L_2$, and probabilities $\rho_h$ with $\sum_h \rho_h= 1$.}
  Set the initial state $\BFz^{0}$ for each client\;
  \For {$t=0,1,\dots,T - 1$} {
  The server randomly selects a block $\BFz_h^{t}$ with probability $\rho_h$\;
  $G(\BFz_h^t, \xi) \leftarrow$ \LG~based on Algorithm O2 in the Online Supplement;
  
  The server computes $\BFz_h^{t+1}$ based on Eq. \eqref{eq:update of theta} or \eqref{eq:update of c}\;
  The server synchronizes clients with the updated block $\BFz_h^{t+1}$\;
  }
  \KwOut{Set $\Bar{\BFz} =\BFz^{t'}$ according to the probability distribution}
 \begin{equation}
\label{eq:output probability}
    \begin{aligned}
    \mathbb{P}(t'=t) = \frac{\eta_t\min\limits_{h\in\{1,2\}}\rho_h(1-\gamma_{h,t}L_h)}{\sum_{t=1}^T \eta_t\min_{h=1,2}\rho_h(1-\gamma_{h,t}L_h)},
    \end{aligned}
\end{equation}
where $\gamma_{1, t} = 4\eta_t$ and $\gamma_{2, t} = \frac{\eta_t}{2}$.
\end{algorithm}

Specifically, we utilize the mirror descent method to develop a unified optimization framework. 
Given $\BFz_h^t$, the descent direction $G$, and the distance-generating function $\omega_h$, $\BFz_{h}^{t+1}$ is obtained via
$
    \BFz_h^{t+1} = P_h(\BFz_h^t, G, \eta)=\argmin_{\BFz_h \in\mathcal{Z}_h}\langle G, \BFz_h-\BFz_h^{t}\rangle + \frac{1}{\eta} D_h(\BFz_h, \BFz_h^{t}),
$
where $D_h(\BFz_h, \BFz_h^t) = \omega_h(\BFz_h) -\omega_h(\BFz_h^t) -\langle\nabla \omega_h(\BFz_h^t), \BFz_h-\BFz_h^t \rangle$ is the Bregman distance.

For $\theta$, we employ the $\ell_2$ norm as the distance-generating function (i.e., $\omega_1(\theta) = \frac{1}{2}\Vert \theta \Vert_2^2$).
Since the size of the parameter $\theta$ is usually large, considering the communication cost, we follow the vanilla FL performing multiple local steps.
Thus, given the stochastic gradient at the $\tau$-th local step in the $i$-th client $\nabla_{\theta} f_i(\theta_{i, \tau}^t; \BFc_i, \xi_{i, \tau}^t)$, we derive $\theta_{i, \tau + 1}^t$ as follows:
\begin{equation}
       \theta_{i, \tau + 1}^{t} = 
    \arg\min_{\theta} \langle \nabla_{\theta} f_i(\theta_{i, \tau}^t; \BFc_i, \xi_{i, \tau}^t), \theta - \theta_{i, \tau}^t\rangle + \frac{1}{2\eta_{1, t}} (\Vert \theta\Vert_2^2 - \Vert \theta_{i, \tau}^t\Vert_2^2 - \langle\theta_{i, \tau}^t, \theta - \theta_{i, \tau}^t\rangle),
    \label{eq:update of theta in each client}
\end{equation}
where $\BFc_{i}$ is fixed and the data point $\xi_{i, \tau}^t$ is uniformly randomly selected from the $i$-th client.

With the first-order optimality condition for the unconstrained problem, Eq. \eqref{eq:update of theta in each client} reduces to
$
\theta_{i,\tau+1}^{t} = \theta_{i, \tau}^t - \eta_{1, t} \nabla_{\theta} f_{i}(\theta_{i, \tau}^t; \BFc_i, \xi_{i, \tau}^t),
$
which is the same as the stochastic gradient descent. Hence, updating the parameter based on Eq. \eqref{eq:update of theta in each client} is equivalent to the local update in the \texttt{FedAvg} method \citep{mcmahan2017communication}. We obtain the $\theta^{t+1}$ as follows:
\begin{equation}
    \theta^{t+1} = P_1(\theta^t, G(\theta^t, \xi), \eta) = \theta^t - \eta_{1, t}\sum_{i=1}^M p_i \sum_{\tau=0}^{E-1} \nabla_{\theta} f_i(\theta_{i, \tau}^t; \BFc_i, \xi_{i, \tau}^t),
    \label{eq:update of theta}
\end{equation}
where $G(\theta^t, \xi^t) = \sum_{i=1}^M p_i \sum_{\tau=0}^{E-1} \nabla_{\theta} f_i(\theta_{i, \tau}^t; \BFc_i, \xi_{i, \tau}^t)$ is the accumulated stochastic gradient.

For $\BFc_i$ falling within the simplex space, the entropy function $\omega_2(\BFc) = \BFc^\top\log \BFc$ is commonly adopted as the distance-generating function.
We utilize the Majorization-Minimization method \citep{tuck2019distributed} to further address the non-separable Laplacian regularization.
Specifically, since the affinity matrix $\BFW$ is usually positive semi-definite \citep{nader2019positive}, at time $t$, an upper bound on the objective function \citep{ziko2020laplacian} in Eq. \eqref{eq:reformulated problem} is given by
    \begin{equation}
        \sum_{i=1}^M p_i f_i(\theta, \BFc_i) + \lambda \BFC^\top (\BFD\otimes \BFI) \BFC - \lambda [(\BFC^t)^\top (\BFW \otimes \BFI) \BFC^t +2 ((\BFW \otimes \BFI)\BFC^t)^\top(\BFC - \BFC^t)].
        \label{eq:surrogate function}
    \end{equation}
    
Let $S_t(\mathbf{C};\theta)$ denote the above surrogate function. Note that the surrogate function $S_t(\mathbf{C};\theta)\geq F(\mathbf{C};\theta)$ $\forall \mathbf{C} \in \mathcal{C}$ conditioned on $\theta$ and $S_t(\mathbf{C}^t;\theta) = F(\mathbf{C}^t;\theta)$. Then, each client's problem can be solved in parallel: 
$
    \min_{\BFc_i\in \Delta^K} p_i f_i(\theta, \BFc_i) + \lambda d_{ii} \langle \BFc_i, \BFc_i\rangle - 2\lambda\langle \sum_{j=1}^M w_{ij} \BFc_j^t, \BFc_i - \BFc_i^t \rangle,
$
where $\Delta^K$ denotes the simplex space.
Since the size of the $K$-dimensional vector is relatively small, we perform only one mirror descent step for $\BFc$ in the above problem, i.e., 
\begin{equation}
    \BFc_i^{t+1} = \arg\min_{\BFc_i \in \Delta^K} \langle p_i \nabla_{\BFc}f_i(\BFc_{i}^t; \theta, \xi_{i}^t) + 2\lambda (\BFL \BFC^t)_i, \BFc_i - \BFc_i^t\rangle + \frac{1}{\eta} D_2(\BFc_i, \BFc_i^t),
    \label{eq:update of c_i in each client}
\end{equation}
where $(\BFL\BFC^t)_i = d_{ii} \BFc_i^t - \sum_{j=1}^M w_{ij} \BFc_j^t$ is a slice of the vector $\BFL\BFC^t$ from index $(i-1)*K$ to $i*K - 1$. Through Eq. \eqref{eq:update of c_i in each client} and the function $\omega_2$, we have
$
 \BFc_{i}^{t+1} = \BFc_i^t \cdot \frac{\exp(-\eta (\nabla_{\BFc}f_i(\BFc_{i}^t; \theta, \xi_{i}^t) + 2\lambda (\BFL \BFC^t)_i))}{\langle \BFc_i^t, \exp(-\eta( \nabla_{\BFc}f_i(\BFc_{i}^t; \theta, \xi_{i}^t) + 2\lambda (\BFL \BFC^t)_i)) \rangle}.
$

Hence, for the vector $\BFC$, we update all membership vectors simultaneously using Eq. \eqref{eq:update of c_i in each client}, i.e., 
$
\BFC^{t+1} =  \scalebox{0.7}{\(\begin{bmatrix}
        \BFc_1^{t+1}\\
        \vdots\\
        \BFc_M^{t+1}
    \end{bmatrix}\)} =\scalebox{0.7}{\(\begin{bmatrix}
     \argmin\limits_{\BFc_1\in \Delta^K} \langle G_{\BFc_1}(\BFC^t,\xi_1^t), \BFc_1-\BFc_1^t\rangle + \frac{1}{\eta} D_2(\BFc_1,\BFc_1^t)\\
    \vdots\\
    \argmin\limits_{\BFc_M\in \Delta^K} \langle G_{\BFc_M}(\BFC^t,\xi_M^t), \BFc_M-\BFc_M^t\rangle + \frac{1}{\eta} D_2(\BFc_M,\BFc_M^t)
    \end{bmatrix}\)},
$
with $G(\BFC^t, \xi)=\scalebox{0.7}{\( \begin{bmatrix}        G_{\BFc_1}(\BFC^t, \xi_1^t)\\
        \vdots\\
        G_{\BFc_M}(\BFC^t, \xi_M^t)\end{bmatrix}\)}=\scalebox{0.7}{\( \begin{bmatrix} 
    p_1 \nabla_{\BFc} f_1(\BFc_1^t;\theta, \xi_1^t)\\
    \vdots\\
    p_M \nabla_{\BFc} f_M(\BFc_M^t;\theta, \xi_M^t)
    \end{bmatrix}\)} + 2\lambda \BFL \BFC^t.$
Finally, we obtain $\BFC^{t+1}$ as follows:
\begin{equation}
    \label{eq:update of c}
    \BFC^{t+1}
    =\scalebox{1}{\(\begin{bmatrix}
    \BFc_1^t \cdot \exp(-\eta G_{\BFc_1}(\BFC^t, \xi_1^t)) / \langle \exp(-\eta G_{\BFc_1}(\BFC^t, \xi_1^t)), \BFc_1^t\rangle\\
     \vdots\\
    \BFc_M^t \cdot \exp(-\eta G_{\BFc_M}(\BFC^t, \xi_M^t)) / \langle\exp(-\eta G_{\BFc_M}(\BFC^t, \xi_M^t)), \BFc_M^t\rangle\\
    \end{bmatrix}\)}.
\end{equation}

In summary, in Algorithm~\ref{alg:ppfl algorithm}, the server first samples a block $\BFz_h^t$ with probability $\rho_h$ and broadcasts it to clients. 
Then, clients perform the corresponding local update (refer to Algorithm~O2 in Section O2 in the Online Supplement). 
Subsequently, the server updates the selected block with uploaded gradients using Eq. \eqref{eq:update of theta} or \eqref{eq:update of c} and then synchronizes it with clients.
Finally, the server outputs the value $\BFz^{t'}$ from the trajectory $\{\BFz^t\}_{t=1}^{T}$ following the probability distribution in Eq. \eqref{eq:output probability}.

\begin{remark}
In contrast to previous studies, the proposed RBCD-based \texttt{PPFL} algorithm offers a solution to a more general FL setting. 
Unlike the centralized RBCD algorithm \citet{dang2015stochastic}, it is specifically designed within the FL framework, which enables unconstrained blocks to perform multiple local steps for improved communication efficiency \citep{mcmahan2017communication}.
Compared to previous BCD approaches in the FL framework, the proposed algorithm partitions the parameters into blocks, diverging from the method of treating each client as a block utilized by \citet{wu2021federated}, and enables a flexible update strategy tailored to each block of parameters.
Furthermore, the stochastic RBCD-based \texttt{PPFL} algorithm does not necessitate the stringent jointly convex and strongly convex assumptions as in \citet{hanzely2023personalized}. Instead, it is designed for a more general setting, requiring only the smoothness assumption and being capable of handling linear constraints.
\end{remark}

\subsection{Convergence Analysis}
\label{sec:convergence}

In this section, we will demonstrate the convergence properties of the proposed algorithm for the framework in Eq.~\eqref{eq:reformulated problem}.
Throughout the rest of this paper, $\Vert \cdot \Vert$ denotes the $\ell_2$ norm if not specified.
Without loss of generality, we assume the objective function $F$ is well-bounded by $F^*$ and make the following assumptions:
\begin{assumption}[Smoothness] 
\label{assumption:smoothness}
For the $i$-th client, $i\in[M]$, there exist constants $L_1$ and $L_2$ such that:
\begin{itemize}
    \item Given any $\BFc_i \in \Delta^K$, $\Vert \nabla_{\theta}f_i(\theta', \BFc_i) - \nabla_{\theta}f_i(\theta, \BFc_i)\Vert \leq L_1\Vert \theta' - \theta\Vert$ for any feasible $\theta'$ and $\theta$;
    \item 
    Given any $\theta$, $\Vert p_i\nabla_{\BFc}f_i(\theta, \BFc_i') - p_i\nabla_{\BFc}f_i(\theta, \BFc_i)+ 2\lambda d_{ii}(\BFc_i' - \BFc_i)\Vert_{\infty} \leq L_2\Vert \BFc_i' - \BFc_i\Vert_1$ for any  $\mathbf{c}$, $\mathbf{c}'\in\Delta^K$.
\end{itemize}
\end{assumption}

\begin{assumption}[Unbiased Stochastic Gradient $\&$ Bounded Variance] 
\label{assumption:bounded variance}
For any $\tau \in [0,E-1]$, $i\in[M]$, and $t\in[0,T-1]$, we assume that the stochastic gradient estimated at the random data point $\xi_{i, \tau}^t$ is unbiased, i.e.,
\begin{itemize}
    \item $\mathbb{E}_{\xi_{i, \tau}^t}[\nabla_{\theta} f_i(\theta;\BFc_i, \xi_{i,\tau}^t)]=\nabla_{\theta} f_i(\theta;\BFc_i)$, for any given $\BFc_i\in \Delta^K$ and any feasible $\theta$;
    \item $\mathbb{E}_{\xi_i^t}[\nabla_{\BFc} f_i(\BFc_i;\theta, \xi_i^t)]= \nabla_{\BFc} f_i(\BFc_i; \theta)$, for any given $\theta$ and any $\BFc_i\in \Delta^K$.
\end{itemize}
Furthermore, there exist constants $\sigma_1^2$ and $\sigma_2^2$ such that
\begin{itemize}
    \item $\mathbb{E}_{\xi_{i, \tau}^t}\Vert \nabla_{\theta} f_i(\theta;\BFc_i, \xi_{i,\tau}^t) - \nabla_{\theta} f_i(\theta;\BFc_i)\Vert^2 \leq \sigma_1^2$, for any given $\BFc_i \in \Delta^K$ and any feasible $\theta$;
    \item $\mathbb{E}_{\xi_i^t}\Vert \nabla_{\BFc} f_i(\BFc_i;\theta, \xi_{i}^t) - \nabla_{\BFc} f_i(\BFc_i;\theta)\Vert^2 \leq \sigma_2^2$, for any given $\theta$ and any $\BFc_i \in \Delta^K$.
\end{itemize}
\end{assumption}

\begin{assumption}[Bounded Gradient Diversity] There exists a constant $\delta^2$ such that for any given $\BFC\in\mathcal{C}$ and any feasible $\theta$, we have
\label{assumption:Gradient Diversity}
$
    \sum_{i=1}^M p_i \Vert \nabla_{\theta} f_i(\theta;\BFc_i) - \nabla_{\theta} F(\theta, \BFC)\Vert^2 \leq \delta^2.
$
\end{assumption}

Note that Assumptions \ref{assumption:smoothness} and \ref{assumption:bounded variance} are commonly applied in the optimization field. 
From Assumption \ref{assumption:smoothness}, we can further infer that the function $F$ is also $L_1$-smooth in $\theta$ for all $\BFC\in \mathcal{C}$, and the surrogate function defined in Eq. \eqref{eq:surrogate function} is $L_2$-smooth in $\BFC$ for any given $\theta$.
Assumption \ref{assumption:Gradient Diversity} is widely accepted in the FL framework \citep{pillutla2022federated}, where $\delta^2$, serving as a non-i.i.d. metric, measures the heterogeneity level across clients.

In non-convex optimization, the gradient norm is commonly utilized as a convergence criterion, where a gradient norm of $0$ implies that a stationary point is obtained. 
Hence, for the unconstrained parameter $\theta$, we use the gradient norm $\Vert \nabla_{\theta}F(\theta, \BFC)\Vert$ to indicate whether $\theta$ has converged with a fixed $\BFC$.
For the constrained parameter, following \citet{dang2015stochastic}, the distance between the current parameter and the updated one, $\Vert\mathcal{G}(\BFz_h^t, \BFz_h^{t+1}, \eta)\Vert = \frac{1}{\eta}\Vert(\BFz^t_h- \BFz^{t+1}_h)\Vert$, replaces the gradient norm.
Thus, for the parameter $\BFC\in\mathcal{C}$, we utilize the criterion 
$\Vert\mathcal{G}(\BFC^{t}, \BFC^{t+1}_+, \eta)\Vert_1 = \frac{1}{\eta}\Vert(\BFC^t - \BFC^{t+1}_+)\Vert_1$,
where $\BFC_{+}^{t+1}$ is composed of 
\begin{equation}
\BFc_{i, +}^{t+1}:= \arg\min_{\BFc_i \in \Delta^K} \langle p_i \nabla_{\BFc}f_i(\BFc_{i}^t; \theta) + 2\lambda (\BFL \BFC^t)_i, \BFc_i - \BFc_i^t\rangle + \frac{1}{\eta} D_2(\BFc_i, \BFc_i^t),
\label{eq:update of c with correct gradient}
\end{equation}
where the full gradient $\nabla_{\BFc}f_i(\BFc_{i}^t; \theta)$ replaces $\nabla_{\BFc}f_i(\BFc_{i}^t; \theta, \xi_{i}^t)$ in Eq. \eqref{eq:update of c_i in each client}.
Furthermore, we define 
$
    G^t = \scalebox{0.6}{\( \begin{bmatrix}
        \nabla_{\theta}F(\theta^t, \BFC^t)\\
        \mathcal{G}(\BFC^t, \BFC^{t+1}_+, \eta)
    \end{bmatrix}\)},
$
with the norm $\Vert G^t \Vert^2 = \Vert \nabla_{\theta}F(\theta^t, \BFC^t)\Vert^2 + \Vert\mathcal{G}(\BFC^t, \BFC^{t+1}_+, \eta)\Vert_1^2$.

\begin{remark}
In \citet{dang2015stochastic}, the smoothness property of the distance generating function $\omega$ is required to ensure the validity of $\Vert\mathcal{G}(\BFz_h^t, \BFz_h^{t+1}, \eta)\Vert$. For the distance generating function $\omega_2(\mathbf{c}_i)=\mathbf{c}_i^\top \log(\mathbf{c}_i)$ employed for the parameter $\mathbf{c}_i$, the smoothness property does not hold if any element $c_{ik} =0$. To remedy this problem and maintain the validity of $\Vert\mathcal{G}(\BFz_h^t, \BFz_h^{t+1}, \eta)\Vert$, we add some constant term such as $10^{-6}$ to prevent any element of $\mathbf{c}_i$ from becoming $0$.
\end{remark}

Under the above assumptions and convergence criterion, we present the convergence properties of Algorithm \ref{alg:ppfl algorithm} in Theorem \ref{thm:convergence property}.
The detailed proof is provided in Section O1.1 in the Online Supplement.
\begin{theorem}
\label{thm:convergence property}
Under Assumptions \ref{assumption:smoothness}, \ref{assumption:bounded variance} and \ref{assumption:Gradient Diversity}, if the learning rate $\eta_t$ with $t\in[0, T]$ satisfies $\eta_t=\min\{\frac{1}{32EL_1}, \frac{2}{L_2}\}$ and $\eta_t=E\eta_{1, t} = \eta_{2, t}$, and $\bar \BFz :=\BFz^{t'}$ with the probability given in Eq.~\eqref{eq:output probability}, then we have
\begin{equation}
    \mathbb{E}[\Vert \bar{G}\Vert^2] \leq \frac{2\Delta_F+2\rho_2(M+1)\sum_{t=0}^{T-1}\eta_t\sigma_2^2 + \rho_1\sum_{t=0}^{T} \eta_t^2(\sigma_1^2 + \delta^2)}{ \sum_{t=0}^{T-1} \eta_t\min_{h\in\{1, 2\}}\rho_h(1-\gamma_{h,t}L_h)},\label{eq:thm1 result}
\end{equation}
where $\gamma_{1, t} = 4\eta_t$, $\gamma_{2, t} = \frac{\eta_t}{2}$, $\Delta_F=F(\BFz^0) - F^*$, and $\bar{G}:=G^{t'}$.
\end{theorem}

The numerator in the inequality \eqref{eq:thm1 result} contains two distinct sources of error. The first error term $2\rho_2(M+1)\sum_{t=0}^{T-1}\eta_t\sigma_2^2$ arises from the sampling error during gradient computation for $\mathbf{c}$, while the second term $\rho_1\sum_{t=0}^{T} \eta_t^2(\sigma_1^2 + \delta^2)$ includes both the sampling error and divergence parameter $\delta^2$. In the following Corollary \ref{cor:convergence rate with a constant learning rate}, we demonstrate the convergence property with a constant learning rate.

\begin{corollary}
\label{cor:convergence rate with a constant learning rate}
    With the same assumptions as in Theorem \ref{thm:convergence property}, considering the constant learning rate $\eta_t = \frac{1}{\sqrt{T}}$, $T \geq \max\{1024E^2L_1^2, \frac{L_2^2}{4}\}$, we obtain that 
    \begin{equation}
        \mathbb{E}[\Vert \bar{G}\Vert^2] = \mathcal{O}\Big(\frac{2\Delta_F + \rho_1 (\sigma_1^2 + \delta^2)}{ \sqrt{T}} + 2\rho_2(M+1)\sigma_2^2\Big).
        \label{eq:corollary 1}
    \end{equation}
\end{corollary}

From Eq. \eqref{eq:corollary 1}, if $\sigma_2^2 = 0$ with learning rate $\eta_t = \frac{1}{\sqrt{T}}$, Algorithm \ref{alg:ppfl algorithm} achieves the convergence rate of $\mathcal{O}(\frac{1}{\sqrt{T}})$.
To this end, we can calculate the full gradient for $\mathbf{c}$, avoiding the error arising from the random sampling step. 
When $\sigma_1^2=\sigma_2^2=\delta=0$, Algorithm \ref{alg:ppfl algorithm} achieves the same $\mathcal{O}(\frac{1}{T})$ convergence rate as \texttt{FedADMM} \citep{zhou2023federated}, as supported by Theorem \ref{thm:convergence property} and detailed in Corollary O1 in Section O1.1 in the Online Supplement.

Besides, we propose a new communication-efficient algorithm, Algorithm O1 in Section O2 in the Online Supplement, which updates $\mathbf{c}_i$ with full gradient and $\theta$ with stochastic gradient alternatively and saves some synchronization costs.
Employing the same lemmas, we can derive the same sub-linear convergence rate $\mathcal{O}(\frac{1}{\sqrt{T}})$.
This can be verified by the following Theorem \ref{thm: sub-linear convergence}.

\begin{theorem}
\label{thm: sub-linear convergence}
Under Assumptions \ref{assumption:smoothness}, \ref{assumption:bounded variance} and \ref{assumption:Gradient Diversity}, when taking Algorithm O1 in the Online Supplement and calculating the full gradient for $\mathbf{c}_i$, $i\in [M]$, i.e., $\sigma_2^2=0$, if the learning rate $\eta$ satisfies:
$
    \eta = \min\{\frac{1}{16L_1}, \frac{1}{L_2}, (\frac{4\Delta_F}{\sigma_{1, 1}T})^{1/2}, (\frac{4\Delta_F}{\sigma_{1, 2}T})^{1/3}\},
$
then we have 
\begin{equation}
\label{eq:corollary inequality}
    \begin{aligned}
        \frac{1}{T} \sum_{t=0}^{T-1} \mathbb{E}[\Vert G(\BFz^t)\Vert^2]\leq \frac{(4\Delta_{F}\sigma_{1,1})^{1/2}}{\sqrt{T}} + \frac{(16\Delta_F^2\sigma_{1, 2}^2)^{1/3}}{T^{2/3}} + \frac{4\Delta_F}{(16L_1+L_2)T},
    \end{aligned}
\end{equation}
where $\Vert G(\BFz^t)\Vert^2=\Vert \nabla_{\theta}F(\theta^{t}, \BFC^{t+1})\Vert^2+\Vert \mathcal{G}(\BFz^{t}, \BFz^{t+1}, \eta)\Vert_1^2$, $\sigma_{1, 1} = 4L_1\sigma_1^2$, and $\sigma_{1, 2} = 128L_1^2 \delta^2$.
\end{theorem}

\begin{remark}
Comparing the proposed RBCD-based PFL algorithm and the alternating minimization variant with the state-of-the-art result presented by \citet{pillutla2022federated}, who employ a decoupling method and demonstrate the sub-linear rate for an alternating minimization algorithm for the non-convex unconstrained problem, we achieve an equivalent sub-linear convergence rate for the non-convex constrained problem if taking the full gradient for the constrained variable. 
\end{remark}

\section{Experiment}
\label{sec:experiment}

In this section, we evaluate \texttt{PPFL} on both pathological and practical data sets and compare it with several baselines, including clustered FL, multi-task PFL, and decoupling PFL methods. 
The code is available at the GitHub repository \citep{ppfl2023}.

\subsection{Pathological Data Sets}
\label{sec:Manipulated}
In the pathological setting, we initially split all data into several pre-specified groups based on their labels, and clients' local data sets are then generated based on these groups. Subsequently, we demonstrate the effectiveness and interpretability of \texttt{PPFL} on these manipulated data sets.
     Moreover, we examine its performance under varying degrees of data heterogeneity. 
Finally, we assess \texttt{PPFL} on an additional covariate-shift data set, where feature distributions vary across clients.

\textbf{Setup.}
We first generate a domain-heterogeneous version of the MNIST and CIFAR-10 data sets following \citet{jeong2022factorized}, which simulates distinct subgroups in the population by restricting clients to hold only one kind of group-specific data. 
Besides, we also create a synthetic data set following \citet{marfoq2021federated},
where the number of samples per class in each client follows a Dirichlet distribution, with all classes available to clients.
There is no predefined clear group structure in this data set. 
Specifically, the created MNIST and CIFAR-10 data sets contain 100 and 80 clients distributed equally across four and three predefined label groups, respectively. 

The synthetic data set contains 300 clients, whose local data are a mixture of three distinct data sources instead of single-source within domain-heterogeneous data sets.
For all data sets, we use 80\% of each client’s data for training and the remaining 20\% for evaluation.
Further details on  the experimental setup and model specifications are provided in Section O4 in the Online Supplement.

We implement \texttt{PPFL} with the two proposed structures in Figures \ref{fig:form 1} and \ref{fig:form 2}, named \texttt{PPFL1} and \texttt{PPFL2}, respectively. 
As a comparison, we consider the purely local training approach, where each client individually trains its own model, and the well-known \texttt{FedAvg} \citep{mcmahan2017communication} as our baselines. 
\ours is also compared to several multi-task PFL methods, including \texttt{pFedMe} \citep{t2020personalized}, \texttt{FedEM} \citep{marfoq2021federated}, the clustered FL method \citep{sattler2020clustered}, and the decoupling PFL, \texttt{FedLG} \citep{Liang2020ThinkLA}. 

\begin{table}[t]
\centering
\caption{Comparison of PPFL with other methods on pathological data sets. The average test accuracy (\%) over five independent trials and the standard deviation (\%) are reported.}
\label{tab:performance on manipulated data sets}
\resizebox{\textwidth}{!}{
\begin{tabular}{lccccccccc}
\hline
& Local & FedAvg & pFedMe & FedEM & Clustered FL & FedLG & PPFL1 & PPFL2 \\ 
\hline
 MNIST     & $96.71_{0.05}$ & $96.64_{0.05}$  & $96.93_{0.06}$  & $\bm{99.04}_{0.03}$ & $98.19_{0.36}$        & $96.95_{0.07}$ & $\bm{99.01}_{0.03}$  & $99.00_{0.03}$                                          \\
CIFAR10   & $54.10_{1.27}$ & $48.98_{0.13}$  & $72.96_{0.28}$  & $84.93_{2.48}$ & $84.73_{0.58}$        & $77.71_{0.30}$ & $\bm{85.33}_{0.29}$  & $85.29_{0.16}$                                                          \\
 Synthetic & $66.13_{0.08}$ & $69.34_{0.00}$  & $70.21_{0.02}$  & $76.25_{0.73}$ & $72.32_{0.29}$        & $69.34_{0.00}$ & $\bm{77.60}_{0.78}$ & $76.39_{0.17}$
\\ 
\hline
\end{tabular}
}
\end{table}

\textbf{Performance Evaluation.} 
For the pathological data sets, the average test accuracy of all methods over five independent trials is reported as shown in Table \ref{tab:performance on manipulated data sets}. 
We can find that the proposed method achieves the best accuracy on MNIST and CIFAR-10 data sets. 
In domain-heterogeneous scenarios, \texttt{FedEM}, Clustered FL, and \texttt{PPFL} significantly outperform the others by taking full advantage of the underlying subgroup structure. 
However, \texttt{PPFL} has a lower variance than \texttt{FedEM} and clustered FL. 
This implies our method has stable performance and is less sensitive to the initial parameter values in complicated neural networks.
Additionally, the proposed method achieves the best performance among all baselines on the Synthetic data set.
Despite the unknown underlying group structure among clients in the synthetic data set, \ours can still outperform all baselines, including \texttt{FedEM}, by appropriately tuning the number of canonical models.

\begin{figure}[t]
\centering
{\subfloat[MNIST]{\includegraphics[scale=0.33]{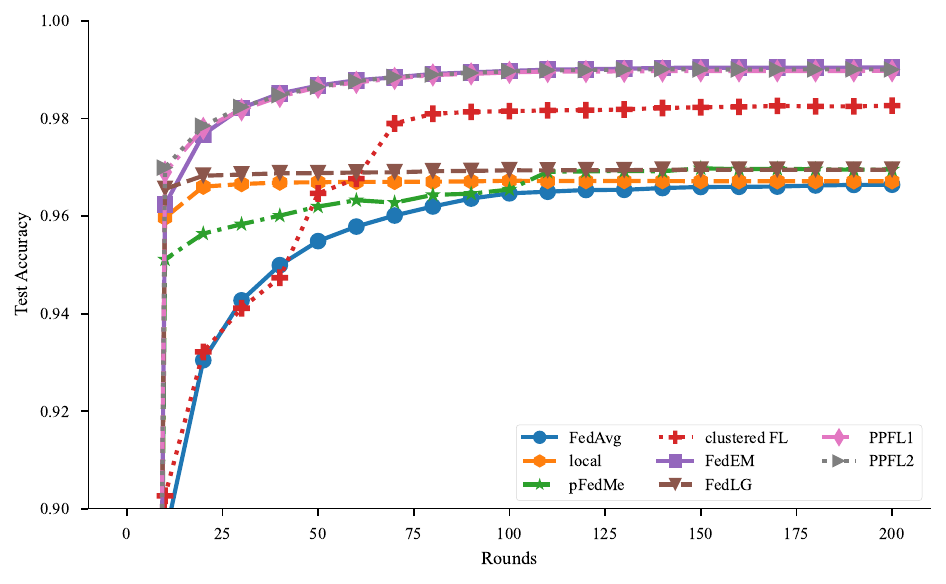}}
\subfloat[CIFAR-10]{\includegraphics[scale=0.33]{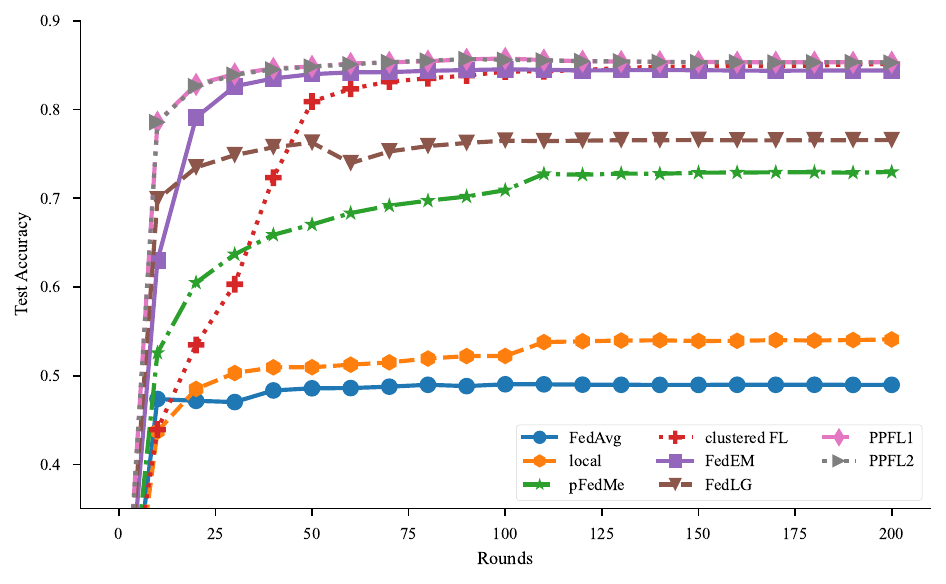}}
\subfloat[Synthetic]{\includegraphics[scale=0.33]{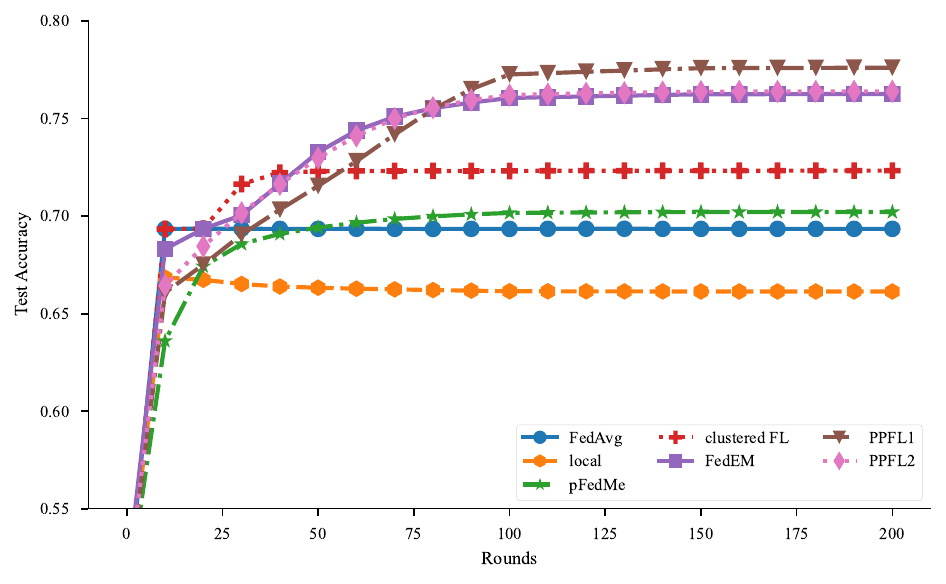}}}
\caption{Average Test Accuracy $\&$ Communication Cost. The average test accuracy is on (a) the MNIST data set, (b) the CIFAR-10 data set, and (c) the synthetic data set. }
\label{fig:curve of manipulated data sets}
\end{figure}

Furthermore, we plot the average accuracy over communication rounds in Figure \ref{fig:curve of manipulated data sets}.
On domain-heterogeneous data sets, \texttt{PPFL} takes fewer communication rounds to converge than all other methods, implying lower communication costs.
On the synthetic data set, \texttt{PPFL} and \texttt{FedEM} take similar communication rounds to reach the same test accuracy.
However, \texttt{PPFL} has lower computational time per round and lower theoretical communication costs compared to \texttt{FedEM}, due to the use of the parameter $\theta_{\text{com}}$ shared among the canonical models.
We present a detailed analysis of the communication costs between \ours and \texttt{FedEM} in Section O5.2 in the Online Supplement, where we demonstrate \ours is both more computation-efficient and communication-efficient than \texttt{FedEM}.
Additionally, results for wall-clock running times and training accuracy are provided in Section O5.1 and Section O5.7 in the Online Supplement, respectively.

\begin{figure}[t]
\caption{(a) Heatmaps of the matrix $\BFC$ during training on the MNIST data set. Each row of the matrix denotes the vector $\BFc_i$ for a specific client. Darker-colored elements have values close to $1$; (b) Heatmaps of the final $\BFC$ matrix for different values of $N$ on the MNIST data set.}
    \centering
    \subfloat[]{\label{fig:C matrix on MNIST}\includegraphics[scale=0.4]{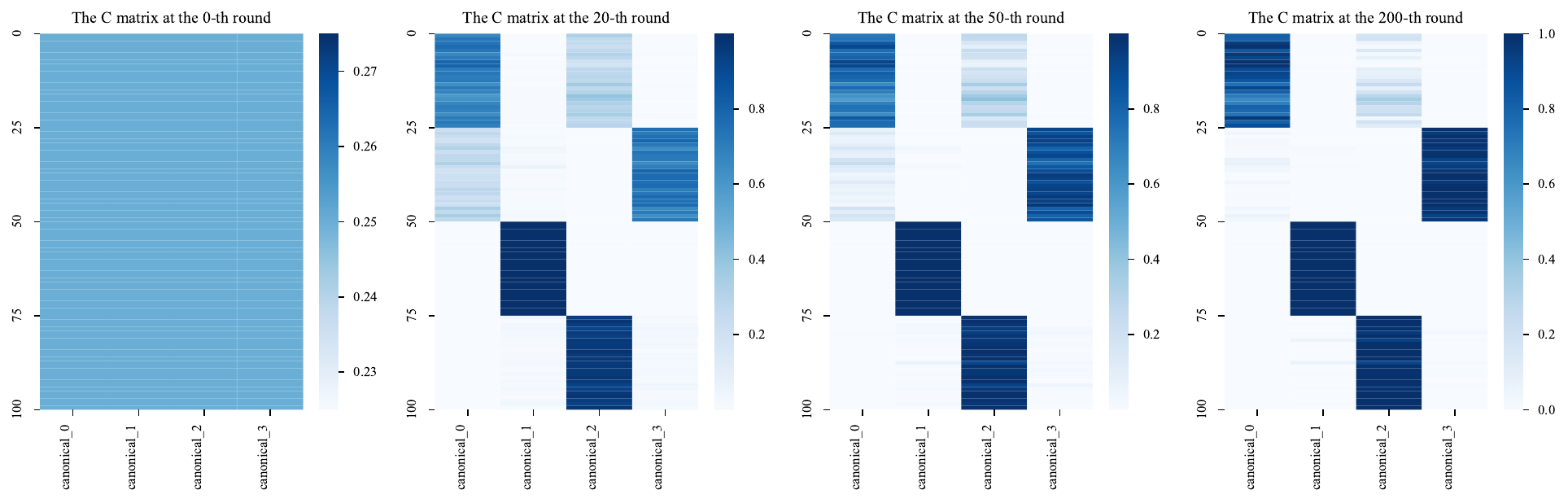}}\\
    \subfloat[]{\label{fig:compare mnist}\includegraphics[scale=0.4]{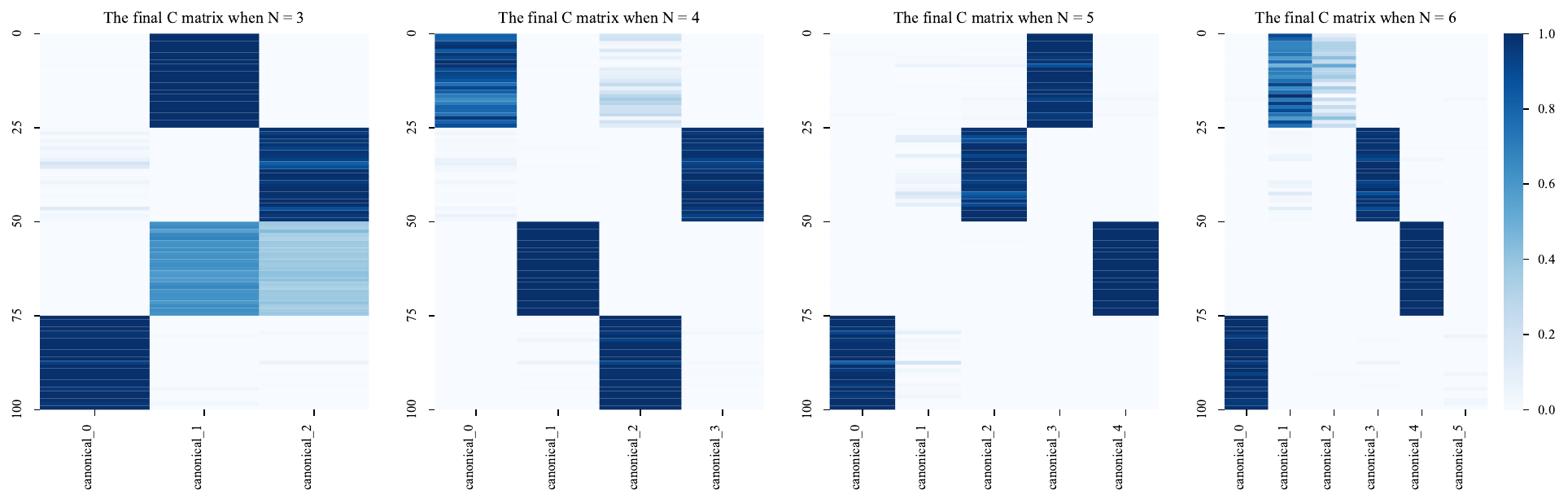}}
    \label{fig:total C MNIST}
\end{figure}

To demonstrate the interpretability of \ours, we analyze the values of the matrix $\BFC$ with $\BFC=[\BFc_1,\dots,\BFc_m]^\top$, throughout the training process on domain-heterogeneous data sets. 
As shown in Figure \ref{fig:C matrix on MNIST}, initially, all elements of $\BFC$ are set to $0.25$.
After $20$ communication rounds, \ours approximately recognizes the subgroups of clients.
The membership vector converges to a unit vector as communication rounds increase.
Furthermore, clients within the same group have similar membership vectors.
This result implies the one-to-one correspondence between canonical models and subgroups, and canonical models can specialize in learning the characteristics of each subgroup, and highlights \ours's effectiveness in capturing clients' characteristics.
The phenomenon where membership vectors gradually approximate the underlying structure is also observed on CIFAR-10 and synthetic data sets, and the results are provided in {Section O5.3} in the Online Supplement.
With these membership vectors on hand, we can effectively infer the characteristics of subgroups and clients, which can further support business decisions such as the market segmentation strategy.

\textbf{Effect of the Number of Canonical Models.} 
Since the number of underlying groups is usually unknown, we examine the influence of the number of canonical models on the group identification performance. 
Specifically, we vary the value of $N$ on the MNIST data set and investigate the heatmaps of the final $\BFC$ matrices.
As shown in Figure \ref{fig:compare mnist}, when $N\geq K$, clients from different groups employ distinct canonical models, whereas for $N<K$ (e.g., $N=3$), they share the same canonical model.
In Section O5.5 in the Online Supplement, we present \ours' results with different $N$ on both the CIFAR-10 and Synthetic data set, where predefined client clusters are absent.
These results suggest that when the ground-truth number of underlying groups is unknown, employing a relatively large number of canonical models can achieve acceptable performance while providing clues for estimating the value of $K$.

\textbf{Effect of Heterogeneity.} 
\label{sec:heterogeneity}
Here, we evaluate the performance of \ours in comparison to baseline methods under varying degrees of data heterogeneity. 
Specifically, we reuse the reconfigured data sources within domain-heterogeneous data sets and distribute samples to clients following a Dirichlet distribution $\text{Dir}(\alpha)$, $\alpha\in \{0.2, 0.5, 1\}$.
Based on the properties of Dirichlet distribution, when $\alpha$ is small, the probability distribution is concentrated near all vertices.
When $\alpha =1$, the Dirichlet distribution reduces to a uniform distribution.
Thus, the smaller value of $\alpha$ corresponds to higher heterogeneity levels among clients.

As shown in Table \ref{tab:performance with varying alpha}, \texttt{PPFL} consistently outperforms the baselines on the MNIST data set across all values of $\alpha$.
On the CIFAR-10 data set, it also achieves the best performance for both $\alpha = 0.5$ and $\alpha = 0.2$.
Note that even though \texttt{FedAvg} achieves the best performance at $\alpha=1$ (i.e., the client distributions are approximately homogeneous) on CIFAR-10, \texttt{PPFL} still ranks as the second-best method.
Additionally, as the value of $\alpha$ decreases, the performance of \texttt{PPFL} improves—a trend also observed in other personalized methods, such as \texttt{FedLG} and \texttt{FedEM}.
These results suggest that \texttt{PPFL} is particularly advantageous in highly and moderately heterogeneous settings. 
Besides, we assess \ours on the covariate-shift data set, with results in Section O5.4 in the Online Supplement confirming its effectiveness in such settings.

\begin{table}[t]
\centering
\caption{Comparison of PPFL with other methods on MNIST and CIFAR10 data sets using different values of $\alpha$ for data partitioning. The average test accuracy ($\%$) over five independent trials and the standard deviation ($\%$) are reported.}
\label{tab:performance with varying alpha}
\renewcommand\arraystretch{1.2}
\resizebox{\textwidth}{!}{
\begin{tabular}{ccccccccccc}
\hline
Data set & $\alpha$ & Local & FedAvg & pFedMe & FedEM & Clustered FL & FedLG & PPFL1 & PPFL2 \\ 
\hline
\multirow{3}{*}{MNIST} & $\alpha=1$   &  $88.73_{0.05}$     &    $97.43_{0.07}$    &   $91.81_{0.06}$     &   $96.71_{0.13}$    &    $90.86_{0.15}$          & $89.08_{0.16}$      &   $97.48_{0.12}$    &    $\mathbf{97.54}_{0.13}$   \\
& $\alpha=0.5$ &   $89.64_{0.05}$    &   $97.33_{0.06}$     &  $92.33_{0.10}$      &   $96.63_{0.12}$    &    $92.56_{0.23}$          &    $90.01_{0.12}$   &   $\mathbf{97.60}_{0.09}$    &  $97.55_{0.11}$     \\
& $\alpha=0.2$ &  $92.41_{0.03}$     &  $97.21_{0.03}$      &   $93.52_{0.08}$     &  $97.16_{0.02}$     &     $94.15_{0.18}$         &    $92.64_{0.09}$   &       $\mathbf{97.81}_{0.08}$ & $97.72_{0.06}$   \\ 
\cline{2-10}
\multirow{3}{*}{CIFAR10}& $\alpha=1$   &  $58.50_{0.29}$     & $\textbf{81.56}_{0.27}$  &  $67.14_{0.14}$      &  $79.97_{0.33}$     &         $69.21_{0.18}$  &   $77.52_{0.14}$    &   $80.91_{0.08}$    &   $81.32_{0.12}$    \\
& $\alpha=0.5$ &    $60.52_{0.43}$   &   $81.43_{0.16}$     &  $67.25_{0.19}$      &    $80.56_{0.20}$   &            $71.21_{0.13}$  &    $78.45_{0.21}$   &    $\textbf{81.78}_{0.08}$   &    $81.59_{0.17}$   \\
& $\alpha=0.2$ &  $67.78_{0.24}$     &  $81.23_{0.27}$      &  $68.84_{0.35}$      &   $82.47_{0.50}$    &            $75.47_{0.24}$  &   $82.28_{0.16}$    &  $\textbf{83.24}_{0.42}$     & $83.11_{0.28}$      \\ \hline
\end{tabular}
}
\end{table}

\subsection{Practical Data Sets}
Different from pathological settings, 
in practical scenarios, the clients' local data sets are generated by themselves, such as the content of their posts on the Internet. 
Therefore, we evaluate \ours on practical data sets to verify its effectiveness on real-world applications.

\label{sec:real data set}

\textbf{Setup.}
We evaluate \ours on Federated EMNIST ($1114$ clients) and StackOverflow ($1000$ clients) data sets, where the true number of underlying groups is unknown and local data sizes vary across clients.
We use ResNet-18 for the EMNIST data set and a 4-layer transformer for the StackOverflow.

We adopt local fine-tuning as a baseline alongside \texttt{FedAvg}.
Given that practical data sets often require large models and potentially require a higher value of $K$, the substantial $dK$ parameter leads to extensive memory requirements for both clustered FL and \texttt{FedEM} methods , which renders their implementation impractical.
Thus, we limit comparisons to the following approaches: Two multi-task PFL methods—Ditto \citep{li2021ditto} and \texttt{pFedMe} \citep{t2020personalized}, and Partial FL \citep{pillutla2022federated}, a decoupling PFL method.
Detailed information on data sets, model specifications, training pipeline, and hyperparameter tuning is provided in Section O4 in the Online Supplement.

\textbf{Performance.}
Table \ref{tab:performance on practical data sets} presents the accuracy of all PFL methods over five independent trials. 
Specifically, on the EMNIST data set, \texttt{PPFL} achieves the best accuracy among all other methods, demonstrating the proposed method's effectiveness in practical recognition tasks.
While on the StackOverflow data set, \texttt{PPFL} exhibits lower predictive accuracy than others.
The suboptimal performance may stem from two reasons: 1) The underlying fundamental properties (group structure) are absent in the StackOverflow data set, which directly relates to the performance degradation, as shown in \citet{chen2022towards} and Section \ref{sec:heterogeneity};
2) This may result from an improper structure and a limited number of canonical models. Following \citet{pillutla2022federated}, we employ the output layer as the canonical model, containing about $10^6$ parameters, which incurs a substantial memory footprint and further restricts the number of canonical models.
Additionally, as shown in Figure O6 in the Online Supplement, data sets with higher homogeneity tend to require a larger number of canonical models.
Hence, using a much smaller number of canonical models to represent the large parameter space for the homogeneous data set may lead to deteriorated performance.

\begin{table}
\centering
\caption{Comparison of PPFL with other methods on practical data sets. The average test accuracy ($\%$) over five trials and the standard deviation ($\%$) are reported.}
\label{tab:performance on practical data sets}
\resizebox{\textwidth}{!}{
\begin{tabular}{cccccccc}
\hline
              & FedAvg & Fintune & Ditto & pFedMe & Partial FL & PPFL1 & PPFL2 \\ \hline
EMNIST        & $93.16$  & $94.14_{0.06}$   & $94.13_{0.06}$ & $94.14_{0.06}$  & $94.07_{0.05}$      & $\bm{94.19}_{0.02}$  & $94.16_{0.02}$   \\
StackOverflow & $23.78$  & $25.15_{0.06}$   & $\bm{25.16}_{0.06}$ & $25.08_{0.06}$  & $25.06_{0.03}$      & $24.21_{0.01}$  & $23.85_{0.01}$  \\ \hline
\end{tabular}}
\end{table}

In summary, results on both pathological and practical data sets show that \texttt{PPFL} achieves the best performance on data sets exhibiting group structures, as well as those with high and moderate levels of heterogeneity.
In such cases, our method can effectively capture and leverage this heterogeneity.
Moreover, even when the latent group structure is not predefined or known, \ours can still achieve considerable performance by selecting an appropriate number of canonical models. 

For the performance difference between \ppflone and \ppfltwo, our intuition is that this difference is rooted in the loss landscapes of these canonical models, and the degree of heterogeneity among clients and the specific architecture of these models significantly shape these landscapes.
A detailed discussion of the differences is presented in Section O5.8 in the Online Supplement.

\section{Conclusion}
\label{sec:conclusion}
This paper presents a novel PFL framework, \texttt{PPFL}, which enhances model-based interpretability compared to existing PFL methods.
Assuming the existence of intrinsic characteristics within the heterogeneous population, we contend that diverse clients exhibit heterogeneity in their preferences for these characteristics.
The proposed method is designed to capture them via canonical models and membership vectors, respectively, which can directly reveal each client's underlying characteristics and aid in understanding the population.
Apart from interpretability, the flexibility of \texttt{PPFL} enables it to leverage the advantages of existing PFL methods while addressing their limitations.
The proposed algorithm can handle the constrained problem with non-separable regularization while ensuring convergence in the non-convex setting.
Results on both pathological and practical data sets validate the interpretability and stability of \ours, demonstrating comparable accuracy even when the underlying characteristics are unknown.

Despite the effectiveness of the proposed method, there are several limitations that warrant further discussion. 
First, as shown in experiments, the prediction accuracy of \ours decreases when the number of canonical models is inappropriate.
Selecting the number of canonical models is non-trivial and depends on the specific data set and model structure, often requiring empirical exploration.
Furthermore, the architecture of the canonical models, such as the specific layer types and the depth of the model, is an empirical question that needs to be carefully addressed based on the problem characteristics.
Additionally, the stochastic noise in the client-specific parameters $\rvc$ can negatively impact the stability and performance of identifying the canonical models, as indicated by the dependence on the stochastic noise term $\sigma_2^2$ in our convergence analysis. 
To mitigate the negative impact of noise, larger batch sizes or deterministic approaches may be necessary, increasing the local computation burden.
Moreover, the proposed algorithm requires full participation, which may cause a high communication cost when handling a large number of clients.
The algorithm with partial participation and multiple local update steps for all blocks and the corresponding convergence properties should be considered in future work.

\bibliography{reference}

\begin{thebibliography}{55}
\providecommand{\natexlab}[1]{#1}
\providecommand{\url}[1]{\texttt{#1}}
\expandafter\ifx\csname urlstyle\endcsname\relax
  \providecommand{\doi}[1]{doi: #1}\else
  \providecommand{\doi}{doi: \begingroup \urlstyle{rm}\Url}\fi

\bibitem[Bonkhoff and Grefkes(2022)]{bonkhoff2022precision}
Anna~K Bonkhoff and Christian Grefkes.
\newblock Precision medicine in stroke: towards personalized outcome predictions using artificial intelligence.
\newblock \emph{Brain}, 145\penalty0 (2):\penalty0 457--475, 2022.

\bibitem[Caldas et~al.(2018)Caldas, Duddu, Wu, Li, Kone{\v{c}}n{\`y}, McMahan, Smith, and Talwalkar]{caldas2018leaf}
Sebastian Caldas, Sai Meher~Karthik Duddu, Peter Wu, Tian Li, Jakub Kone{\v{c}}n{\`y}, H~Brendan McMahan, Virginia Smith, and Ameet Talwalkar.
\newblock Leaf: A benchmark for federated settings.
\newblock \emph{arXiv preprint arXiv:1812.01097}, 2018.

\bibitem[Cao et~al.(2023)Cao, Chen, Fan, Gama, Ong, and Kumar]{cao2023bayesian}
Longbing Cao, Hui Chen, Xuhui Fan, Joao Gama, Yew-Soon Ong, and Vipin Kumar.
\newblock Bayesian federated learning: A survey.
\newblock \emph{arXiv preprint arXiv:2304.13267}, 2023.

\bibitem[Chai et~al.(2020)Chai, Ali, Zawad, Truex, Anwar, Baracaldo, Zhou, Ludwig, Yan, and Cheng]{chai2020tifl}
Zheng Chai, Ahsan Ali, Syed Zawad, Stacey Truex, Ali Anwar, Nathalie Baracaldo, Yi~Zhou, Heiko Ludwig, Feng Yan, and Yue Cheng.
\newblock Tifl: A tier-based federated learning system.
\newblock In \emph{Proceedings of the 29th international symposium on high-performance parallel and distributed computing}, pages 125--136, 2020.

\bibitem[Chen et~al.(2022)Chen, Deng, Wu, Gu, and Li]{chen2022towards}
Zixiang Chen, Yihe Deng, Yue Wu, Quanquan Gu, and Yuanzhi Li.
\newblock Towards understanding the mixture-of-experts layer in deep learning.
\newblock In \emph{Advances in Neural Information Processing Systems}, pages 23049--23062, 2022.

\bibitem[Collins et~al.(2021)Collins, Hassani, Mokhtari, and Shakkottai]{collins2021exploiting}
Liam Collins, Hamed Hassani, Aryan Mokhtari, and Sanjay Shakkottai.
\newblock Exploiting shared representations for personalized federated learning.
\newblock In \emph{International Conference on Machine Learning}, pages 2089--2099, 2021.

\bibitem[Dang and Lan(2015)]{dang2015stochastic}
Cong~D Dang and Guanghui Lan.
\newblock Stochastic block mirror descent methods for nonsmooth and stochastic optimization.
\newblock \emph{SIAM Journal on Optimization}, 25\penalty0 (2):\penalty0 856--881, 2015.

\bibitem[Dayan et~al.(2021)Dayan, Roth, Zhong, Harouni, Gentili, Abidin, Liu, Costa, Wood, Tsai, et~al.]{dayan2021federated}
Ittai Dayan, Holger~R Roth, Aoxiao Zhong, Ahmed Harouni, Amilcare Gentili, Anas~Z Abidin, Andrew Liu, Anthony~Beardsworth Costa, Bradford~J Wood, Chien-Sung Tsai, et~al.
\newblock Federated learning for predicting clinical outcomes in patients with covid-19.
\newblock \emph{Nature Medicine}, 27\penalty0 (10):\penalty0 1735--1743, 2021.

\bibitem[Di et~al.(2025)Di, Yang, Ye, and Xiang]{ppfl2023}
Hao Di, Yi~Yang, Haishan Ye, and Yuchang Xiang.
\newblock {PPFL}: A personalized federated learning framework for heterogeneous population, 2025.
\newblock URL \url{https://github.com/INFORMSJoC/2023.0376}.
\newblock Available for download at https://github.com/INFORMSJoC/2023.0376.

\bibitem[Duan et~al.(2020)Duan, Liu, Chen, Liu, Tan, and Liang]{duan2020self}
Moming Duan, Duo Liu, Xianzhang Chen, Renping Liu, Yujuan Tan, and Liang Liang.
\newblock Self-balancing federated learning with global imbalanced data in mobile systems.
\newblock \emph{IEEE Transactions on Parallel and Distributed Systems}, 32\penalty0 (1):\penalty0 59--71, 2020.

\bibitem[Feffer et~al.(2018)Feffer, Rudovic, and Picard]{feffer2018mixture}
Michael Feffer, Ognjen Rudovic, and Rosalind~W Picard.
\newblock A mixture of personalized experts for human affect estimation.
\newblock In \emph{Machine Learning and Data Mining in Pattern Recognition}, pages 316--330, 2018.

\bibitem[Feng et~al.(2020)Feng, Zhu, Wang, Huang, and Chen]{feng2020learning}
Jingshuo Feng, Xi~Zhu, Feilong Wang, Shuai Huang, and Cynthia Chen.
\newblock A learning framework for personalized random utility maximization (rum) modeling of user behavior.
\newblock \emph{IEEE Transactions on Automation Science and Engineering}, 19\penalty0 (1):\penalty0 510--521, 2020.

\bibitem[Ganin and Lempitsky(2015)]{ganin2015unsupervised}
Yaroslav Ganin and Victor Lempitsky.
\newblock Unsupervised domain adaptation by backpropagation.
\newblock In \emph{International conference on machine learning}, pages 1180--1189. PMLR, 2015.

\bibitem[Ghosh et~al.(2020)Ghosh, Chung, Yin, and Ramchandran]{ghosh2020efficient}
Avishek Ghosh, Jichan Chung, Dong Yin, and Kannan Ramchandran.
\newblock An efficient framework for clustered federated learning.
\newblock In \emph{Advances in Neural Information Processing Systems}, pages 19586--19597, 2020.

\bibitem[Hanzely et~al.(2023)Hanzely, Zhao, and Kolar]{hanzely2023personalized}
Filip Hanzely, Boxin Zhao, and Mladen Kolar.
\newblock Personalized federated learning: A unified framework and universal optimization techniques.
\newblock \emph{Transactions on Machine Learning Research}, 2023.

\bibitem[Hull(1994)]{hull1994database}
Jonathan~J. Hull.
\newblock A database for handwritten text recognition research.
\newblock \emph{IEEE Transactions on pattern analysis and machine intelligence}, 16\penalty0 (5):\penalty0 550--554, 1994.

\bibitem[Jeong and Hwang(2022)]{jeong2022factorized}
Wonyong Jeong and Sung~Ju Hwang.
\newblock Factorized-{FL}: Personalized federated learning with parameter factorization \& similarity matching.
\newblock In \emph{Advances in Neural Information Processing Systems}, pages 35684--35695, 2022.

\bibitem[Kamp et~al.(2023)Kamp, Fischer, and Vreeken]{kamp2023federated}
Michael Kamp, Jonas Fischer, and Jilles Vreeken.
\newblock Federated learning from small datasets.
\newblock In \emph{International Conference on Learning Representations}, 2023.

\bibitem[Kim and Allenby(2022)]{kim2022integrating}
Hyowon Kim and Greg~M Allenby.
\newblock Integrating textual information into models of choice and scaled response data.
\newblock \emph{Marketing Science}, 41\penalty0 (4):\penalty0 815--830, 2022.

\bibitem[Kosolwattana et~al.(2024)Kosolwattana, Wang, Kontar, and Lin]{kosolwattana2024fcom}
Tanapol Kosolwattana, Huazheng Wang, Raed~Al Kontar, and Ying Lin.
\newblock {FCOM}: A federated collaborative online monitoring framework via representation learning.
\newblock \emph{arXiv preprint arXiv:2405.20504}, 2024.

\bibitem[Kotelevskii et~al.(2022)Kotelevskii, Vono, Durmus, and Moulines]{kotelevskii2022fedpop}
Nikita Kotelevskii, Maxime Vono, Alain Durmus, and Eric Moulines.
\newblock Fedpop: A bayesian approach for personalised federated learning.
\newblock In \emph{Advances in Neural Information Processing Systems}, pages 8687--8701, 2022.

\bibitem[LeCun et~al.(1998)LeCun, Bottou, Bengio, and Haffner]{lecun1998gradient}
Yann LeCun, L{\'e}on Bottou, Yoshua Bengio, and Patrick Haffner.
\newblock Gradient-based learning applied to document recognition.
\newblock \emph{Proceedings of the IEEE}, 86\penalty0 (11):\penalty0 2278--2324, 1998.

\bibitem[Li et~al.(2021{\natexlab{a}})Li, Li, and Varshney]{li2021federated}
Chengxi Li, Gang Li, and Pramod~K Varshney.
\newblock Federated learning with soft clustering.
\newblock \emph{IEEE Internet of Things Journal}, 9\penalty0 (10):\penalty0 7773--7782, 2021{\natexlab{a}}.

\bibitem[Li et~al.(2020)Li, Sahu, Talwalkar, and Smith]{li2020federated}
Tian Li, Anit~Kumar Sahu, Ameet Talwalkar, and Virginia Smith.
\newblock Federated learning: Challenges, methods, and future directions.
\newblock \emph{IEEE Signal Processing Magazine}, 37\penalty0 (3):\penalty0 50--60, 2020.

\bibitem[Li et~al.(2021{\natexlab{b}})Li, Hu, Beirami, and Smith]{li2021ditto}
Tian Li, Shengyuan Hu, Ahmad Beirami, and Virginia Smith.
\newblock Ditto: Fair and robust federated learning through personalization.
\newblock In \emph{International Conference on Machine Learning}, pages 6357--6368, 2021{\natexlab{b}}.

\bibitem[Li et~al.(2022)Li, Tong, Anjum, Mohammed, Chen, and Jiang]{li2022federated}
Wentao Li, Jiayi Tong, Md~Monowar Anjum, Noman Mohammed, Yong Chen, and Xiaoqian Jiang.
\newblock Federated learning algorithms for generalized mixed-effects model (glmm) on horizontally partitioned data from distributed sources.
\newblock \emph{BMC Medical Informatics and Decision Making}, 22\penalty0 (1):\penalty0 269, 2022.

\bibitem[Li et~al.(2019)Li, Huang, Yang, Wang, and Zhang]{li2019convergence}
Xiang Li, Kaixuan Huang, Wenhao Yang, Shusen Wang, and Zhihua Zhang.
\newblock On the convergence of fedavg on non-iid data.
\newblock In \emph{International Conference on Learning Representations}, 2019.

\bibitem[Li et~al.(2021{\natexlab{c}})Li, JIANG, Zhang, Kamp, and Dou]{li2020fedbn}
Xiaoxiao Li, Meirui JIANG, Xiaofei Zhang, Michael Kamp, and Qi~Dou.
\newblock Fedbn: Federated learning on non-iid features via local batch normalization.
\newblock In \emph{International Conference on Learning Representations}, 2021{\natexlab{c}}.

\bibitem[Liang et~al.(2020)Liang, Liu, Ziyin, Salakhutdinov, and Morency]{Liang2020ThinkLA}
P.~P. Liang, Terrance Liu, Liu Ziyin, R.~Salakhutdinov, and Louis-Philippe Morency.
\newblock Think locally, act globally: Federated learning with local and global representations.
\newblock \emph{arXiv preprint arXiv:2001.01523}, 2020.

\bibitem[Lin et~al.(2017)Lin, Liu, Byon, Qian, Liu, and Huang]{lin2017collaborative}
Ying Lin, Kaibo Liu, Eunshin Byon, Xiaoning Qian, Shan Liu, and Shuai Huang.
\newblock A collaborative learning framework for estimating many individualized regression models in a heterogeneous population.
\newblock \emph{IEEE Transactions on Reliability}, 67\penalty0 (1):\penalty0 328--341, 2017.

\bibitem[Lin et~al.(2018)Lin, Liu, and Huang]{lin2018selective}
Ying Lin, Shan Liu, and Shuai Huang.
\newblock Selective sensing of a heterogeneous population of units with dynamic health conditions.
\newblock \emph{IISE Transactions}, 50\penalty0 (12):\penalty0 1076--1088, 2018.

\bibitem[Liu et~al.(2021)Liu, Guo, and Chen]{liu2021pfa}
Bingyan Liu, Yao Guo, and Xiangqun Chen.
\newblock {PFA}: Privacy-preserving federated adaptation for effective model personalization.
\newblock In \emph{Web Conference}, pages 923--934, 2021.

\bibitem[Marfoq et~al.(2021)Marfoq, Neglia, Bellet, Kameni, and Vidal]{marfoq2021federated}
Othmane Marfoq, Giovanni Neglia, Aur{\'e}lien Bellet, Laetitia Kameni, and Richard Vidal.
\newblock Federated multi-task learning under a mixture of distributions.
\newblock In \emph{Advances in Neural Information Processing Systems}, pages 15434--15447, 2021.

\bibitem[McMahan et~al.(2017)McMahan, Moore, Ramage, Hampson, and y~Arcas]{mcmahan2017communication}
Brendan McMahan, Eider Moore, Daniel Ramage, Seth Hampson, and Blaise~Aguera y~Arcas.
\newblock Communication-efficient learning of deep networks from decentralized data.
\newblock In \emph{Artificial Intelligence and Statistics}, pages 1273--1282, 2017.

\bibitem[Murdoch et~al.(2019)Murdoch, Singh, Kumbier, Abbasi-Asl, and Yu]{murdoch2019definitions}
W~James Murdoch, Chandan Singh, Karl Kumbier, Reza Abbasi-Asl, and Bin Yu.
\newblock Definitions, methods, and applications in interpretable machine learning.
\newblock \emph{Proceedings of the National Academy of Sciences}, 116\penalty0 (44):\penalty0 22071--22080, 2019.

\bibitem[Nader et~al.(2019)Nader, Bretto, Mourad, and Abbas]{nader2019positive}
Rafic Nader, Alain Bretto, Bassam Mourad, and Hassan Abbas.
\newblock On the positive semi-definite property of similarity matrices.
\newblock \emph{Theoretical Computer Science}, 755:\penalty0 13--28, 2019.

\bibitem[Netzer et~al.(2011)Netzer, Wang, Coates, Bissacco, Ng, et~al.]{netzer2011reading}
Yuval Netzer, Tao Wang, Adam Coates, Alessandro Bissacco, Andrew~Y Ng, et~al.
\newblock Reading digits in natural images with unsupervised feature learning.
\newblock 2011.

\bibitem[Pillutla et~al.(2022)Pillutla, Malik, Mohamed, Rabbat, Sanjabi, and Xiao]{pillutla2022federated}
Krishna Pillutla, Kshitiz Malik, Abdel-Rahman Mohamed, Mike Rabbat, Maziar Sanjabi, and Lin Xiao.
\newblock Federated learning with partial model personalization.
\newblock In \emph{International Conference on Machine Learning}, pages 17716--17758, 2022.

\bibitem[Sattler et~al.(2020)Sattler, M{\"u}ller, and Samek]{sattler2020clustered}
Felix Sattler, Klaus-Robert M{\"u}ller, and Wojciech Samek.
\newblock Clustered federated learning: Model-agnostic distributed multitask optimization under privacy constraints.
\newblock \emph{IEEE Transactions on Neural Networks and Learning Systems}, 32\penalty0 (8):\penalty0 3710--3722, 2020.

\bibitem[Shazeer et~al.(2017)Shazeer, Mirhoseini, Maziarz, Davis, Le, Hinton, and Dean]{shazeer2017outrageously}
Noam Shazeer, Azalia Mirhoseini, Krzysztof Maziarz, Andy Davis, Quoc~V. Le, Geoffrey~E. Hinton, and Jeff Dean.
\newblock Outrageously large neural networks: The sparsely-gated mixture-of-experts layer.
\newblock In \emph{International Conference on Learning Representations}, 2017.

\bibitem[Smith et~al.(2023)Smith, Seiler, and Aggarwal]{smith2022optimal}
Adam~N Smith, Stephan Seiler, and Ishant Aggarwal.
\newblock Optimal price targeting.
\newblock \emph{Marketing Science}, 42\penalty0 (3):\penalty0 476--499, 2023.

\bibitem[Smith et~al.(2017)Smith, Chiang, Sanjabi, and Talwalkar]{smith2017federated}
Virginia Smith, Chao-Kai Chiang, Maziar Sanjabi, and Ameet~S Talwalkar.
\newblock Federated multi-task learning.
\newblock In \emph{Advances in Neural Information Processing Systems}, pages 4424--4434, 2017.

\bibitem[T~Dinh et~al.(2020)T~Dinh, Tran, and Nguyen]{t2020personalized}
Canh T~Dinh, Nguyen Tran, and Josh Nguyen.
\newblock Personalized federated learning with moreau envelopes.
\newblock In \emph{Advances in Neural Information Processing Systems}, pages 21394--21405, 2020.

\bibitem[Tan et~al.(2022)Tan, Yu, Cui, and Yang]{tan2022towards}
Alysa~Ziying Tan, Han Yu, Lizhen Cui, and Qiang Yang.
\newblock Towards personalized federated learning.
\newblock \emph{IEEE Transactions on Neural Networks and Learning Systems}, (Early Access):\penalty0 1--17, 2022.

\bibitem[Tuck et~al.(2019)Tuck, Hallac, and Boyd]{tuck2019distributed}
Jonathan Tuck, David Hallac, and Stephen Boyd.
\newblock Distributed majorization-minimization for {L}aplacian regularized problems.
\newblock \emph{IEEE/CAA Journal of Automatica Sinica}, 6\penalty0 (1):\penalty0 45--52, 2019.

\bibitem[Tuck et~al.(2021)Tuck, Barratt, and Boyd]{tuck2021distributed}
Jonathan Tuck, Shane Barratt, and Stephen Boyd.
\newblock A distributed method for fitting laplacian regularized stratified models.
\newblock \emph{The Journal of Machine Learning Research}, 22\penalty0 (1):\penalty0 2795--2831, 2021.

\bibitem[Verbeke and Lesaffre(1996)]{verbeke1996linear}
Geert Verbeke and Emmanuel Lesaffre.
\newblock A linear mixed-effects model with heterogeneity in the random-effects population.
\newblock \emph{Journal of the American Statistical Association}, 91\penalty0 (433):\penalty0 217--221, 1996.

\bibitem[Wu et~al.(2021)Wu, Scaglione, Wai, Karakoc, Hreinsson, and Ma]{wu2021federated}
Ruiyuan Wu, Anna Scaglione, Hoi-To Wai, Nurullah Karakoc, Kari Hreinsson, and Wing-Kin Ma.
\newblock Federated block coordinate descent scheme for learning global and personalized models.
\newblock In \emph{AAAI Conference on Artificial Intelligence}, pages 10355--10362, 2021.

\bibitem[Yang et~al.(2021)Yang, Huang, Huang, and Chang]{yang2020privacy}
Yi~Yang, Shuai Huang, Wei Huang, and Xiangyu Chang.
\newblock Privacy-preserving cost-sensitive learning.
\newblock \emph{IEEE Transactions on Neural Networks and Learning Systems}, 32\penalty0 (5):\penalty0 2105--2116, 2021.

\bibitem[Yue et~al.(2024)Yue, Kontar, and G{\'o}mez]{yue2024federated}
Xubo Yue, Raed~Al Kontar, and Ana Mar{\'\i}a~Estrada G{\'o}mez.
\newblock Federated data analytics: A study on linear models.
\newblock \emph{IISE Transactions}, 56\penalty0 (1):\penalty0 16--28, 2024.

\bibitem[Zhang et~al.(2023)Zhang, Luo, Wu, He, and Li]{zhang_lightfr_2023}
Honglei Zhang, Fangyuan Luo, Jun Wu, Xiangnan He, and Yidong Li.
\newblock {LightFR}: {Lightweight} {Federated} {Recommendation} with {Privacy}-preserving {Matrix} {Factorization}.
\newblock \emph{ACM Transactions on Information Systems}, 41\penalty0 (4), 2023.

\bibitem[Zhang et~al.(2022)Zhang, Li, Li, Guo, and Shao]{zhang2022personalized}
Xu~Zhang, Yinchuan Li, Wenpeng Li, Kaiyang Guo, and Yunfeng Shao.
\newblock Personalized federated learning via variational bayesian inference.
\newblock In \emph{International Conference on Machine Learning}, pages 26293--26310. PMLR, 2022.

\bibitem[Zhong et~al.(2023)Zhong, He, Ren, Li, and Li]{zhongfeddar}
Aoxiao Zhong, Hao He, Zhaolin Ren, Na~Li, and Quanzheng Li.
\newblock {FedDAR}: Federated domain-aware representation learning.
\newblock In \emph{International Conference on Learning Representations}, 2023.

\bibitem[Zhou and Li(2023)]{zhou2023federated}
Shenglong Zhou and Geoffrey~Ye Li.
\newblock Federated learning via inexact {ADMM}.
\newblock \emph{IEEE Transactions on Pattern Analysis and Machine Intelligence}, 45\penalty0 (8):\penalty0 9699--9708, 2023.

\bibitem[Ziko et~al.(2020)Ziko, Dolz, Granger, and Ayed]{ziko2020laplacian}
Imtiaz Ziko, Jose Dolz, Eric Granger, and Ismail~Ben Ayed.
\newblock Laplacian regularized few-shot learning.
\newblock In \emph{International Conference on Machine Learning}, pages 11660--11670, 2020.

\end{thebibliography}
\bibliographystyle{plainnat}

\newpage
\appendix
\renewcommand{\thesection}{O\arabic{section}}   
\renewcommand{\theequation}{O\arabic{equation}} 
\renewcommand{\thefigure}{O\arabic{figure}}                
\renewcommand{\thetable}{O\arabic{table}}                  
\renewcommand{\thetheorem}{O\arabic{theorem}}              
\renewcommand{\thelemma}{O\arabic{lemma}}                  

\setcounter{equation}{0}
\setcounter{figure}{0}
\setcounter{table}{0}
\setcounter{theorem}{0}
\setcounter{lemma}{0}

\section{Proof of Theorems}

\subsection{Proof of the Theorem 2}
\label{subsec:proof of thm2}

Let $\Delta_{\BFz_h}^t(F) = F(\BFz_h^{t+1}, \BFz_{\setminus h}^t) - F(\BFz_h^t, \BFz_{\setminus h}^t)$ denote the loss decrease resulting from the update of the block $h$, where $\BFz_{\setminus h}^t$ denotes all other blocks at time $t$ except for the $h$-th block $\BFz_h$. 
The proof is derived as follows: First, we derive an upper bound on $\Delta_{\BFz_h}^t(F)$ for each block. 
Next, we take the expectation w.r.t. the block selection process to derive the overall convergence properties of Algorithm 1 in the main text.
Before presenting the proof of Theorem 2, we will outline how to obtain upper bounds on $\Delta_{\theta}^t(F)$ and $\Delta_{\BFC}^t(F)$ respectively using the following lemmas.

\begin{lemma}
\label{lemma:property of theta}
Under Assumptions 1 $\sim$ 3 in the main text, for any $\BFC\in\mathcal{C}$, if $\eta_{1, t} \leq \frac{1}{2EL_1}$, $0\leq t \leq T$, we have
\begin{equation*}
\begin{aligned}
    &\quad \, \mathbb{E}_{\xi^t}[\Delta_{\theta}^{t}(F)] \\
    &\leq -\frac{E\eta_{1, t}}{2}(1-2E\eta_{1, t}L_1-64E^2\eta_{1, t}^2L_{1}^2)\Vert \nabla_{\theta} F(\theta^t, \BFC^t)\Vert^2 +32 E^3\eta_{1, t}^3L_{1}^2\delta^2 + (8E\eta_{1, t} L_{1}+\frac{1}{2})E^2\eta_{1, t}^2L_{1}\sigma_1^2.
\end{aligned}
\end{equation*}
\end{lemma}

\proof{Proof of Lemma \ref{lemma:property of theta}}
Let $G(\theta^t) = \sum_{i=1}^M p_i \sum_{\tau = 0} ^{E-1} \nabla_{\theta} f_i(\theta^t_{i, \tau};\BFc_i^t)$. According to Assumption 1, we have
    \begin{equation}
        \begin{aligned}
            \Delta_{\theta}^t(F) 
        &\leq \langle \nabla_{\theta} F(\theta^t, \BFC^t), \theta^{t+1} - \theta^{t}\rangle + \frac{L_1}{2}\Vert \theta^{t+1} -\theta^{t}\Vert^2\\
        &= -\eta_{1, t} \langle \nabla_{\theta}F(\theta^t, \BFC^t), G(\theta^t)\rangle + \frac{L_{1}}{2} \Vert \theta^{t+1} - \theta^t\Vert^2 -\eta_{1, t} \langle\nabla_{\theta}F(\theta^t, \BFC^t), G(\theta^t, \xi^t) -G(\theta^t)\rangle\\
        &= -\eta_{1, t}\langle\nabla_{\theta}F(\theta^t, \BFC^t), \sum_{i=1}^M p_i \sum_{\tau = 0}^{E-1} \nabla_\theta f_i(\theta_{i, \tau}^t;\BFc_i^t) - \nabla_{\theta} f_i(\theta^t;\BFc_i^t) + \nabla_{\theta} f_i(\theta^t;\BFc_i^t)\rangle + A_1 + A_2\\
    &= -E\eta_{1, t} \Vert \nabla_{\theta}F(\theta^t, \BFC^t)\Vert^2 - \eta_{1, t} \langle\nabla_{\theta}F(\theta^t, \BFC^t), \sum_{i=1}^M p_i \sum_{\tau = 0}^{E-1} \nabla_\theta f_i(\theta_{i, \tau}^t;\BFc_i^t) - \nabla_{\theta} f_i(\theta^t;\BFc_i^t) \rangle + A_1 + A_2\\
    & \overset{(a)}{\leq} -\frac{E\eta_{1, t}}{2}\Vert \nabla_{\theta}F(\theta^t, \BFC^t)\Vert^2 + \frac{\eta_{1, t}}{2E}\Vert \sum_{i=1}^M p_i \sum_{\tau = 0}^{E-1} \nabla_\theta f_i(\theta_{i, \tau}^t;\BFc_i^t) - \nabla_{\theta} f_i(\theta^t;\BFc_i^t) \Vert^2 + A_1 + A_2\\
    & \leq -\frac{E\eta_{1, t}}{2}\Vert \nabla_{\theta}F(\theta^t, \BFC^t)\Vert^2 + \frac{\eta_{1, t}}{2E}\sum_{i=1}^M p_i \Vert\sum_{\tau = 0}^{E-1} \nabla_\theta f_i(\theta_{i, \tau}^t;\BFc_i^t) - \nabla_{\theta} f_i(\theta^t;\BFc_i^t) \Vert^2 + A_1 + A_2\\
    &\leq -\frac{E\eta_{1, t}}{2}\Vert \nabla_{\theta}F(\theta^t, \BFC^t)\Vert^2 + \frac{L_1^2\eta_{1, t}}{2}\sum_{i=1}^M p_i \sum_{\tau = 0}^{E-1} \Vert\theta_{i, \tau}^t - \theta^t\Vert^2 + A_1 + A_2\label{eq:middle of theta without expectation},
        \end{aligned}
    \end{equation}
where we utilize $-ab \leq \frac{E}{2}a^2 + \frac{1}{2E}b^2$ in the inequality $(a)$, employ Jensen's inequality and convexity property of the $\ell_2$ norm in the third inequality, $
    A_1 := \frac{L_1}{2}\Vert \theta^{t+1} - \theta^t\Vert^2,\text{ and }
    A_2 := -\eta_{1, t} \langle\nabla_{\theta}F(\theta^t, \BFC^t), G(\theta^t, \xi^t) -G(\theta^t)\rangle.
$

Let $\xi_{i, [\tau]}^t = \{\xi_{i, 0}^t, \dots, \xi_{i, \tau}^t\}$ and $\xi_{[\tau]}^t = \{\xi_{i, [\tau]}^t\}_M$. Since
\begin{equation*}
    \begin{aligned}
        \mathbb{E}_{\xi^t} [G(\theta^t,\xi^t) - G(\theta^t)] 
    &= \sum_{\tau=0}^{E-1}\mathbb{E}_{\xi_{[\tau-1]}^t}\left[\mathbb{E}_{\xi_\tau^t}\left[\sum_{i=1}^M p_i \nabla_{\theta} f_i(\theta^t_{i, \tau}; \BFc_i^t, \xi_{i, \tau}^t) - p_i \nabla_{\theta} f_i(\theta^t_{i, \tau}; \BFc_i^t)\vert \xi_{[\tau-1]}^t\right]\right]=0,
    \end{aligned}
\end{equation*}
and $\nabla_{\theta}F(\theta^t, \BFC^t)$ is independent of $\xi^t$,
taking the expectation w.r.t. $\xi^t$ on both sides in Eq. \eqref{eq:middle of theta without expectation}, we obtain
\begin{equation*}
    \mathbb{E}_{\xi^t}[\Delta_{\theta}^t(F)] 
   \leq -\frac{E\eta_{1, t}}{2} \Vert \nabla_{\theta} F(\theta^t, \BFC^t)\Vert^2 + \frac{L_1^2\eta_{1, t}}{2}\sum_{i=1}^Mp_i \sum_{\tau=0}^{E-1}\mathbb{E}_{\xi_{i, [\tau-1]}^t}[\Vert\theta_{i, \tau}^t - \theta^t\Vert^2] + \mathbb{E}_{\xi^t}[A_1].
\end{equation*}

Invoking Lemma \ref{lemma:gap in lsgd} and replacing $\mathbb{E}_{\xi^t}[A_1]$ in the above inequality, we derive
\begin{equation}
\begin{aligned}
    \mathbb{E}_{\xi^t}[\Delta_{\theta}^t(F)] &\leq -\frac{E\eta_{1, t}}{2}(1 - 2E\eta_{1, t}L_1)\Vert \nabla_{\theta} F(\theta^t, \BFC^t)\Vert^2  + \frac{E^2\eta_{1, t}^2L_1\sigma_1^2}{2} + \\
    &\quad \, (\frac{L_1^2\eta_{1, t}}{2} + E\eta_{1, t}^2L_1^3)\sum_{i=1}^Mp_i \sum_{\tau=0}^{E-1}\mathbb{E}_{\xi_{i, [\tau-1]}^t}[\Vert\theta_{i, \tau}^t - \theta^t\Vert^2] \\
   &\leq -\frac{E\eta_{1, t}}{2}(1 - 2E\eta_{1, t}L_1)\Vert \nabla_{\theta} F(\theta^t, \BFC^t)\Vert^2 + \eta_{1, t}L_1^2\sum_{i=1}^Mp_i B1 + \frac{E^2\eta_{1, t}^2L_1\sigma_1^2}{2}
   \end{aligned}
   \label{eq:middle of theta with expectation},
\end{equation}
where the last inequality arises from $\eta_{1, t} \leq \frac{1}{2EL_1}$ and $B1:=\sum_{\tau=0}^{E-1}\mathbb{E}_{\xi_{i, [\tau-1]}^t}[\Vert\theta_{i, \tau}^t - \theta^t\Vert^2]$.

For the term $B_1$ in Eq. \eqref{eq:middle of theta with expectation}, using Lemma \ref{lemma:bound on client drift}, we have
\begin{equation*}
\begin{aligned}
    &\quad \mathbb{E}_{\xi^t}[\Delta_{\theta}^t(F)] \\
    &\leq -\frac{E\eta_{1, t}}{2}(1 - 2E\eta_{1, t}L_1)\Vert \nabla_{\theta} F(\theta^t, \BFC^t)\Vert^2 + 16E^3\eta_{1, t}^3L_1^2\sum_{i=1}^M p_i\Vert \nabla_{\theta}f_i(\theta^t;\BFc_i^t) \Vert^2 + (8E\eta_{1, t}L_1 + \frac{1}{2})E^2\eta_{1, t}^2L_1\sigma_1^2\\
    &\leq -\frac{E\eta_{1, t}}{2}(1 - 2E\eta_{1, t}L_1 - 64E^2\eta_{1, t}^2L_1^2)\Vert \nabla_{\theta} F(\theta^t, \BFC^t)\Vert^2 + 32E^3\eta_{1, t}^3L_1^2 \delta^2 + (8E\eta_{1, t}L_1 + \frac{1}{2})E^2\eta_{1, t}^2L_1\sigma_1^2,
\end{aligned}
\end{equation*}
where we replace $\nabla_{\theta}f_i(\theta^t;\BFc_i
^t)$ with $\nabla_{\theta}f_i(\theta^t;\BFc_i
^t) - \nabla_{\theta}F(\theta^t,\BFC^t) + \nabla_{\theta}F(\theta^t,\BFC^t)$ and employ Assumption 3 to generate the last inequality.\Halmos
\endproof

\begin{lemma}
\label{lemma:property of c}
Under Assumptions 1 and 2 in the main text, if the learning rate $\eta_{2, t}$ satisfies $0 < \eta_{2, t} < \frac{2}{L_2}$ for $0\leq t \leq T$, we have 
\begin{equation}
\label{eq:lemma2}
    \mathbb{E}_{\xi^t}[\Delta_{\BFC}^t(F)]\leq  -\frac{\eta_{2, t}}{2}(1-\frac{L_{2}}{2}\eta_{2, t})\Vert\mathcal{G}_2(\BFC^t, \BFC^{t+1}_+, \eta_{2, t}) \Vert ^2_1 + \eta_{2, t}(M+1)\sigma_2^2.
\end{equation}
\end{lemma}

\proof{Proof of Lemma \ref{lemma:property of c}}
Leveraging the properties of $S_t(\mathbf{C};\theta)$ in Eq. (13) and Assumption 1, we begin with
\begin{equation*}
\begin{aligned}
\Delta_{\BFC}^t(F) &\leq S_t(\mathbf{C^{t+1};\theta^t}) - S_t(\mathbf{C^{t};\theta^t}) \\
&\leq \sum_{i=1}^M \langle p_i\nabla_\BFc f_i(\theta^t, \BFc_i^t) + 2\lambda (\BFL\BFC^T)_i,  \BFc_i^{t+1} - \mathbf{c}_i^t\rangle + \frac{L_2}{2}\Vert \BFc_i^{t+1} - \BFc_i^t\Vert_1^2\\
&= \sum_{i=1}^M -\eta_{2, t}\langle p_i\nabla_\BFc f_i(\BFc_i^t;\theta^t) + 2\lambda (\BFL\BFC^t)_i,  
\mathcal{G}(\BFc_i^t, \BFc_{i}^{t+1}, \eta_{2, t})\rangle + \frac{L_2\eta_{2, t}^2}{2}\Vert \mathcal{G}(\BFc_i^t, \BFc_i^{t+1}, \eta_{2, t})\Vert_1^2.\\
\end{aligned}
\end{equation*}

Let $\nabla_{\BFc}g_i(\BFc_i^t;\theta_i^t, \xi_i^t) = p_i\nabla_\BFc f_i(\BFc_i^t;\theta^t, \xi_i^t) + 2\lambda (\BFL\BFC^t)_i$ and $\BFs_i^t = p_i\nabla_\BFc f_i(\BFc_i^t;\theta^t, \xi_i^t) - p_i\nabla_\BFc f_i(\BFc_i^t;\theta^t)$. 
Substituting $\nabla_{\BFc}g_i(\BFc_i^t;\theta_i^t, \xi_i^t)$ and $\BFs_i^t$ into the above inequality, we obtain
\begin{equation}
\begin{aligned}
    &\quad\, \Delta_{\BFC}^t(F) \\
    &\leq  \sum_{i=1}^M -\eta_{2, t}\langle \nabla_{\BFc} g_i(\BFc_i^t;\theta_i^t, \xi_i^t), \mathcal{G}(\BFc_i^t, \BFc_i^{t+1}, \eta_{2, t})\rangle + \frac{L_2}{2}\eta_{2, t}^2 \Vert \mathcal{G}(\BFc_i^t, \BFc_i^{t+1}, \eta_{2, t})\Vert_1^2 + \eta_{2, t}\langle \BFs_i^t, \mathcal{G}(\BFc_i^t, \BFc_{i}^{t+1}, \eta_{2, t})\rangle.
\end{aligned}
    \label{eq:middle result of c}
\end{equation}
According to \citet[Lemma 4.2]{dang2015stochastic} and Eq. (14), for $i\in[M]$, $0\leq t \leq T$, we have
\begin{align}
    \langle \nabla_{\BFc}g_i(\BFc_i^t;\theta^t,\xi_i^t), \mathcal{G}(\BFc_i^t, \BFc_i^{t+1}, \eta_{2, t})\rangle &\geq \Vert\mathcal{G}(\BFc_i^t, \BFc_i^{t+1}, \eta_{2, t}) \Vert^2_1\label{eq:lan1},\\
\Vert \mathcal{G}(\BFc_i^t, \BFc_i^{t+1}, \eta_{2, t}) - \mathcal{G}(\BFc_i^t, \BFc_{i,+}^{t+1}, \eta_{2, t}) \Vert^2_1 &\leq \Vert p_i \nabla_{\BFc}f_i(\BFc_i^t;\theta^t,\xi_i^t) - p_i \nabla_{\BFc}f_i(\BFc_i^t;\theta^t)\Vert^2_{\infty}\label{eq:lan2}.
\end{align}

Combining Eqs. \eqref{eq:middle result of c} and \eqref{eq:lan1}, we derive that
\begin{equation}
    \begin{aligned}
        \Delta_{\BFC}^t(F) \leq \sum_{i=1}^M -\eta_{2, t} (1-\frac{L_2}{2}\eta_{2, t})
    \Vert \mathcal{G}(\BFc_i^t, \BFc_i^{t+1}, \eta_{2, t})\Vert^2_1 + \eta_{2, t}\langle \BFs_i^t, \mathcal{G}(\BFc_i^t, \BFc_{i}^{t+1}, \eta_{2, t})\rangle.
    \label{eq:mid mid of c_i}
    \end{aligned}
\end{equation}

Using the fact that $a^2-2b^2\leq 2(a-b)^2$ and Eq. \eqref{eq:lan2}, the first term in the above is bounded by 
\begin{equation*}
    \begin{aligned}
        \Vert \mathcal{G}(\BFc_i^{t}, \BFc_i^{t+1}, \eta_{2, t})\Vert_1^2 &\geq \frac{1}{2}\Vert \mathcal{G}(\BFc_i^t, \BFc_{i, +}^{t+1},\eta_{2, t})\Vert^2_1 - \Vert \mathcal{G}(\BFc_i^t, \BFc_i^{t+1},\eta_{2, t})- \mathcal{G}(\BFc_i^t, \BFc^{t+1}_{i,+},\eta_{2, t})\Vert^2_\infty\\
    &\geq \frac{1}{2}\Vert \mathcal{G}(\BFc_i^t, \BFc^{t+1}_{i, +},\eta_{2, t})\Vert^2_1 - \Vert p_i\nabla_{\BFc}f_i(\BFc_i^t;\theta^t, \xi^t)-p_i\nabla_{\BFc}f_i(\BFc_i^t;\theta^t)\Vert^2_1\\
    &= \frac{1}{2}\Vert \mathcal{G}(\BFc_i^t, \BFc^{t+1}_{i, +},\eta_{2, t})\Vert^2_1 - p_i^2\Vert \BFs_i\Vert^2_\infty.
    \end{aligned}
\end{equation*}

Besides, for the last term in Eq. \eqref{eq:mid mid of c_i}, we have
\begin{equation*}
\begin{aligned}
    \langle \BFs^t_i, \mathcal{G}(\BFc_i^t, \BFc_i^{t+1},\eta_{2, t})\rangle
    &= \langle \BFs_i^t, \mathcal{G}(\BFc_i^t, \BFc_i^{t+1},\eta_{2, t}) - \mathcal{G}(\BFc_i^t, \BFc_{i, +}^{t+1},\eta_{2, t}) + \mathcal{G}(\BFc_i^t, \BFc^{t+1}_{i, +},\eta_{2, t})\rangle \\
    &\leq \Vert \BFs_t\Vert_{\infty} \Vert\mathcal{G}(\BFc_i^t, \BFc^{t+1}_{i,+},\eta_{2, t}) - \mathcal{G}(\BFc_i^t, \BFc^{t+1}_i,\eta_{2, t}) \Vert_1 + \langle \BFs_i^t, \mathcal{G}(\BFc_i^t, \BFc^{t+1}_{i,+},\eta_{2, t})\rangle\\
    &\leq \langle \BFs_i^t, \mathcal{G}(\BFc_i^t, \BFc^{t+1}_{i, +},\eta_{2, t})\rangle + \Vert \BFs_i^t\Vert^2_{\infty},
\end{aligned}    
\end{equation*}
where the second inequality is using Eq. \eqref{eq:lan2} again.

Hence, plugging above inequality into Eq. \eqref{eq:mid mid of c_i}, we have
\begin{equation*}
    \begin{aligned}
        &\quad \Delta_{\BFC}^t(F)\\
    &\leq  \sum_{i=1}^M -\eta_{2, t} (1-\frac{L_2}{2}\eta_{2, t}) [\frac{1}{2}\Vert \mathcal{G}(\BFc_i^t, \BFc_{i, +}^{t+1}, \eta_{2, t})\Vert_1^2 - p_i^2 \Vert\BFs_i^t\Vert_\infty^2] + \eta_{2, t}\langle \BFs_i^t, \mathcal{G}(\BFc_i^t, \BFc_{i, +}^{t+1}, \eta_{2, t})\rangle + \eta_{2, t}\Vert \BFs_i^t\Vert^2_\infty\\
    &= -\frac{\eta_{2, t}}{2}(1-\frac{L_2}{2}\eta_{2, t})\Vert \mathcal{G}(\BFC^t, \BFC^{t+1}_+, \eta_{2, t})\Vert_1^2 + \eta_{2, t}\sum_{i=1}^M ((p_i^2(1-\frac{L_2}{2}\eta_{2, t}) + 1))\Vert \BFs_i^t\Vert_\infty^2 + \eta_{2, t}\langle \BFs^t, \mathcal{G}(\BFC^t, \BFC_{+}^{t+1}, \eta_{2, t})\rangle
    \end{aligned}
\end{equation*}
which implies the Eq. \eqref{eq:lemma2} by taking the expectation w.r.t. $\xi^t$.\Halmos
\endproof

Now, with the upper bounds $\Delta_{\theta}^t(F)$ and $\Delta_{\BFC}^t(F)$ on hand, we present the proof of Theorem 2 as follows.

\proof{Proof of Theorem 2:} 

According to Algorithm 1 in the main text, at the time $t\in[0, T]$, let $h^t\in\{1, 2\}$ denote the currently selected block and $\Psi^{t}$ represent the $\sigma$-algebra of $\xi^{t}$ and $h^t$, i.e., $\Psi^{t} = \{\xi^{s}, h^t\}_t$.
When $h^t = 1$ at the time $t$, based on Lemma \ref{lemma:property of theta}, we have
\begin{equation*}
    \begin{aligned}
        &\quad \, \mathbb{E}_{\xi^t}[\Delta_{\theta}^t(F)] \\
        &\leq -\frac{E\eta_{1, t}}{2}(1-2E\eta_{1, t}L_1-64E^2\eta_{1, t}^2L_{1}^2)\Vert \nabla_{\theta} F(\theta^t,C^t)\Vert^2 +32 E^3\eta_{1, t}^3L_{1}^2\delta^2 +(8E\eta_{1, t} L_{1}+\frac{1}{2})E^2\eta_{1, t}^2L_{1}\sigma_1^2.
    \end{aligned}
\end{equation*}

Due to $\eta_{1, t} \leq \frac{1}{32EL_1}$, we can simplify the above inequality as 
$
    \mathbb{E}_{\xi^t}[\Delta_{\theta}^t(F)] 
    \leq  -\frac{E\eta_{1, t}}{2}(1-\gamma_{1, t}L_1) \Vert \nabla_{\theta} F(\theta^t,\BFC^t)\Vert^2 + (\delta^2 + \sigma_1^2) E^2\eta_{1, t}^2L_1,
$
with $\gamma_{1, t} = 4E\eta_{1, t}$, which implies
\begin{equation}
    \frac{E\eta_{1, t}}{2}(1-\gamma_{1, t}L_1) \Vert \nabla_{\theta} F(\theta^t,\BFC^t)\Vert^2 \leq -\mathbb{E}_{\xi^t}[\Delta_{\theta}^t(F)] + (\delta^2 + \sigma_1^2) E\eta_{1, t}. \label{eq:bound of theta gradient}
\end{equation}

When $h^t= 2$, following Lemma \ref{lemma:property of c}, with $\gamma_{2, t} = \frac{\eta_{2, t}}{2}$, we have
\begin{equation}
    \frac{\eta_{2, t}}{2}(1-\gamma_{2, t}L_2)\Vert \mathcal{G}(\BFC^t, \BFC^{t+1}, \eta_{2, t})\Vert^2_1 \leq -\mathbb{E}_{\xi^t}[\Delta_{\BFC}^t(F)] + (M+1)\eta_{2, t}\sigma_2^2\label{eq:bound of C gradient}.
\end{equation}

Note that we omit the expectation conditional on $\Psi^{t-1}$ in both Eqs. \eqref{eq:bound of theta gradient} and \eqref{eq:bound of C gradient}. Let $\eta_t = E\eta_{1, t} = \eta_{2, t}$. Multiplying both sides of Eqs. \eqref{eq:bound of theta gradient} and \eqref{eq:bound of C gradient} by $\rho_1$ and $\rho_2$ respectively and summing them up, we have
\begin{equation}
        \rho_1\eta_t(1-\gamma_{1, t}L_1)\Vert \nabla F(\theta^t, C^t)\Vert^2 + \rho_2\eta_t(1-\gamma_{2, t}L_2)\Vert \mathcal{G}(\BFC^t, \BFC^{t+1}, \eta_{2,t})\Vert^2_1 \leq -2\mathbb{E}_{\Psi^{t}}[\Delta^t(F)] + 2\eta_t\sigma^2,\label{eq:mid of combination}
\end{equation}
where $\sigma^2 = \rho_1(\delta^2+\sigma_1^2) + \rho_2(M+1)\sigma_2^2$.

Taking the minimum value between $\rho_1\eta_t(1-\gamma_{1, t}L_1)$ and $\rho_2\eta_t(1-\gamma_{2, t}L_2)$ in Eq. \eqref{eq:mid of combination}, using the definition of $\mathcal{G}^t$, we obtain
\begin{equation}
    \mathbb{E}_{\Psi^{t-1}}\Vert\mathcal{G}^t\Vert^2\min_{h\in\{1, 2\}}\rho_h\eta_t(1-\gamma_{h, t}L_h) \leq -2\mathbb{E}_{\Psi^t}[F(\BFz^{t+1}) - F(\BFz^t)] + 2\eta_t\sigma^2.
\end{equation}

Summing the above inequality up for $t$ from $0$ to $T-1$, we have
\begin{equation}
    \label{eq:Th2}
    \begin{aligned}
        \sum_{t=0}^{T-1}\min_{h\in \{1, 2\}} \rho_h\eta_t(1-\gamma_{h, t}L_h) \mathbb{E}_{\Psi^{T-1}}\Vert \mathcal{G}^t\Vert^2 \leq 2(F(\BFz^0) - F^*) + 2\sum_{t=0}^{T-1}\eta_t\sigma^2.
    \end{aligned}
\end{equation}
Thus, we can conclude Theorem 2 by taking the expectation of Eq. \eqref{eq:Th2} w.r.t. $t'$.
\Halmos
\endproof

 \proof{Proof of the Theorem 3} With Algorithm \ref{alg:variant ppfl}, we have
\begin{equation*}
    F(\theta^{t+1},\BFC^{t+1}) - F(\theta^t,\BFC^t) = \underbrace{F(\theta^{t+1},\BFC^{t+1}) - F(\theta^{t},\BFC^{t+1})}_{D_\theta} + \underbrace{F(\theta^{t},\BFC^{t+1}) - F(\theta^t,\BFC^t)}_{D_{\BFC}}.
\end{equation*}

Using Lemma \ref{lemma:property of theta}, we have
\begin{equation}
    \begin{aligned}
    \mathbb{E}[D_{\theta}] 
    &\leq -\frac{\eta}{2}(1-2\eta L_1 - 64\eta^2 L_1)\Vert \nabla_{\theta}F(\theta^t, \BFC^{t+1})\Vert^2 + 32 \eta^3L_1^2 \delta^2 + (8\eta L_1+\frac{1}{2})\eta^2L_1\sigma_1^2\\
    &\leq \frac{1}{4}\Vert \nabla_{\theta}F(\theta^t, \BFC^{t+1})\Vert^2 + 32 \eta^3L_1^2 \delta^2 + \eta^2L_1\sigma_1^2,
\end{aligned}
\label{eq:eq-theta}
\end{equation}
where the last inequality is caused by the range of $\eta$, i.e.,  $\eta \leq \frac{1}{16L_1}$.

For $D_{\BFC}$, using Lemma \ref{lemma:property of c} and $\sigma^2 = 0$, we have
\begin{equation}
\begin{aligned}
    \mathbb{E}[D_{\BFC}] \leq -\frac{\eta}{2}(1-\frac{\eta}{2}L_2)\Vert \mathcal{G}(\BFC^t, \BFC_{+}^{t+1}, \eta)\Vert_1^2
 \leq -\frac{1}{4}\Vert \mathcal{G}(\BFC^t, \BFC_{+}^{t+1}, \eta)\Vert_1^2,
\label{eq:corrolary-C}
\end{aligned} 
\end{equation}
where $\eta \leq \frac{1}{L_2}$ leads to the last inequality.

Summing Eqs. \eqref{eq:eq-theta} and \eqref{eq:corrolary-C}, we obtain
$
    \mathbb{E}_{\Psi^t}[\Vert \nabla_{\theta}F(\theta^t, \BFC^{t+1})\Vert^2 + \Vert \mathcal{G}(\BFC^t, \BFC_{+}^{t+1}, \eta)\Vert_1^2] \leq \frac{4\mathbb{E}_{\Psi^t}[F(\BFz^{t+1}) - F(\BFz^t)]}{\eta} + 128\eta^2L_1^2\delta^2 + 4\eta L_1\sigma_1^2.
$
Then, summing this inequality over $t$ from $0$ to $T-1$ and then dividing $T$, we have
\begin{equation*}
    \frac{1}{T} \sum_{t=0}^{T-1}\mathbb{E}_{\Psi^{T-1}}[\Vert \nabla_{\theta}F(\theta^t, \BFC^{t+1})\Vert^2 + \Vert \mathcal{G}(\BFC^t, \BFC_{+}^{t+1}, \eta)\Vert_1^2] \leq \frac{4(F(\BFz^0) - F^*)}{\eta T} + 128\eta^2L_1^2\delta^2 + 4\eta L_1\sigma_1^2,
\end{equation*}
which implies Theorem 3 with the predefined learning rate $\eta$.
\Halmos
\endproof

\begin{lemma} \label{lemma:gap in lsgd}
Under Assumptions 1 and 2 in the main text, following Eq. (12), the difference between $\theta^{t+1}$ and $\theta^t$ is bounded by 
\begin{equation*}
    \begin{aligned}
    \mathbb{E}_{\xi}[\Vert \theta^{t+1}-\theta^{t}\Vert^2] 
    \end{aligned}
    \leq E^2\eta_{1, t}^2\sigma_1^2 + 2E\eta_{1, t}^2L_1^2\sum_{i=1}^M p_i \sum_{\tau=0}^{E-1} \mathbb{E}_{\xi_{i, [\tau-1]}^t}\left[\Vert\theta_{i, \tau}^t- \theta^t\Vert^2\right] + 2E^2\eta_{1, t}^2\Vert\nabla_{\theta} F(\theta^t,\BFC^t) \Vert^2.
\end{equation*}
\end{lemma}
\proof{Proof of Lemma \ref{lemma:gap in lsgd}:} With Eq. (12), we get
\begin{equation*}
    \Vert \theta^{t+1}-\theta^{t}\Vert^2
    = \eta_{1, t}^2 \Vert \sum_{i=1}^Mp_i\sum_{\tau=0}^{E-1}\nabla_{\theta} f_i(\theta_{i,\tau}^t; \BFc_i^t, \xi_{i,\tau}^t)\Vert^2 
    \leq E\eta_{1, t}^2 \sum_{\tau=0}^{E-1} \Vert \sum_{i=1}^M p_i \nabla_{\theta} f_i(\theta_{i, \tau}^t;\BFc_i^t, \xi_{i,\tau}^t)\Vert^2.
\end{equation*}
Taking the expectation w.r.t. $\xi^t$ on both sides in the above inequality, we have
\begin{equation*}
    \begin{aligned}
        &\quad \, \mathbb{E}_{\xi^t}[\Vert \theta^{t+1} - \theta^t\Vert^2] \\
    &\leq E\eta_{1, t}^2 \sum_{\tau=0}^{E-1} \mathbb{E}_{\xi_{[\tau-1]}^t}\left[
    \mathbb{E}_{\xi_{\tau}^t}[\Vert \sum_{i=1}^M p_i \nabla_{\theta} f_i(\theta_{i, \tau}^t;\BFc_i^t, \xi_{i,\tau}^t)\Vert^2\vert \xi_{[\tau-1]}^t]\right]\\
    &\overset{(a)}{=} E^2\eta_{1, t}^2\sigma_1^2 + E\eta_{1, t}^2 \sum_{\tau=0}^{E-1} \mathbb{E}_{\xi_{[\tau-1]}^t}\left[
    \Vert \sum_{i=1}^M p_i\nabla_{\theta} f_i(\theta_{i, \tau}^t;\BFc_i^t)\Vert^2\right]\\
    &= E^2\eta_{1, t}^2\sigma_1^2 + E\eta_{1, t}^2 \sum_{\tau=0}^{E-1} \mathbb{E}_{\xi_{[\tau-1]}^t}\left[
    \Vert \sum_{i=1}^M p_i\left(\nabla_{\theta} f_i(\theta_{i, \tau}^t;\BFc_i^t)- \nabla_{\theta} f_i(\theta^t;\BFc_i^t) + \nabla_{\theta} f_i(\theta^t;\BFc_i^t)\right)\Vert^2\right]\\
    &\leq E^2\eta_{1, t}^2\sigma_1^2 + 2E\eta_{1, t}^2 \sum_{\tau=0}^{E-1} \mathbb{E}_{\xi_{[\tau-1]}^t}[\Vert\sum_{i=1}^M p_i \nabla_{\theta} f_i(\theta_{i, \tau}^t;\BFc_i^t)- \nabla_{\theta} f_i(\theta^t;\BFc_i^t)\Vert^2] + 2E^2\eta_{1, t}^2\Vert\nabla_{\theta} F(\theta^t,\BFC^t) \Vert^2\\
    &\overset{(b)}{\leq} E^2\eta_{1, t}^2\sigma_1^2 + 2E\eta_{1, t}^2L_1^2\sum_{i=1}^M p_i \sum_{\tau=0}^{E-1} \mathbb{E}_{\xi_{i, [\tau-1]}^t}\left[\Vert\theta_{i, \tau}^t- \theta^t\Vert^2\right] + 2E^2\eta_{1, t}^2\Vert\nabla_{\theta} F(\theta^t,\BFC^t) \Vert^2,
    \end{aligned}
\end{equation*}
where we use the fact that $\mathbb{E}\Vert \BFx\Vert^2=\Vert \mathbb{E}[\BFx]\Vert^2+E\Vert \BFx-\mathbb{E}[\BFx]\Vert^2$ in (a), and $(b)$ is caused by the convexity of $\ell_2$ norm, Jensen's inequality, and Assumption 1.
\Halmos 
\endproof

\begin{lemma} \label{lemma:bound on client drift} Holding Assumptions 1 $\sim$ 3 in the main text, for the sequence $\{\theta_{i,\tau}^t\}$ generated based on the local stochastic gradient method in Eq. (12), with a fixed learning rate $\eta_{1, t}$, $\eta_{1, t} \leq \frac{1}{2EL_1}$, we have
\begin{equation*}
\sum_{\tau=0}^{E-1} \mathbb{E}_{\xi_{i, \tau-1}}[\Vert \theta^t_{i, \tau} - \theta^t\Vert^2] \leq 16E^3\eta_{1, t}^2\Vert \nabla_{\theta}f_i(\theta^t, \BFc_i^t)\Vert^2 + 8E^3\eta_{1, t}^2\sigma_{1}^2.    
\end{equation*}
\end{lemma}
\proof{Proof of Lemma \ref{lemma:bound on client drift}} Given the current local step $\tau$, $0\leq \tau < E$, we start with
\begin{equation}
\label{eq:middle of heterogeneity}
    \begin{aligned}
    \Vert \theta^t_{i, \tau} - \theta^t\Vert^2 
    &= \Vert \theta^t_{i, \tau - 1} - \theta^t -\eta_{1, t} \nabla_{\theta}f_i(\theta_{i, \tau -1}^t;\BFc_i^t, \xi_{i,\tau-1}^t)\Vert^2\\
    &= \Vert \theta^t_{i, \tau - 1} - \theta^t\Vert^2 - 2 \langle \theta^t_{i, \tau - 1} - \theta^t,
    \eta_{1, t} \nabla_{\theta}f_i(\theta_{i, \tau -1}^t;\BFc_i^t, \xi_{i,\tau-1}^t)
    \rangle + \eta_{1, t}^2\Vert \nabla_{\theta}f_i(\theta_{i, \tau -1}^t;\BFc_i^t, \xi_{i,\tau-1}^t)\Vert^2\\
    &\leq (1+ \frac{1}{2E-1}) \Vert \theta^t_{i, \tau - 1} - \theta^t\Vert^2 + 2E\eta_{1, t}^2 \Vert \nabla_{\theta}f_i(\theta_{i, \tau -1}^t;\BFc_i^t, \xi_{i,\tau-1}^t)\Vert^2 ,        
    \end{aligned}
\end{equation}
where we utilize $-2ab\leq \frac{1}{2E-1}a^2 + (2E-1)b^2$ in the last inequality.

Taking the expectation w.r.t. $\xi^t_{i, [\tau-1]}$ on both sides in Eq. \eqref{eq:middle of heterogeneity}, we get
\begin{equation*}
    \begin{aligned}
    &\quad\, \mathbb{E}_{\xi_{i, \tau-1}}[\Vert \theta^t_{i, \tau} - \theta^t\Vert^2]\\
    &\leq 2E\eta_{1, t}^2 \mathbb{E}_{\xi_{i, [\tau-2]}^t}\left[
    \mathbb{E}_{\xi_{i, \tau-1}^t}[\Vert\nabla_{\theta} f_i(\theta_{i, \tau-1}^t;\BFc_i^t, \xi_{i, \tau-1}^t)\Vert^2|\xi_{i, [\tau-2]}^t]\right] + (1+\frac{1}{2E-1})\mathbb{E}_{\xi_{i, \tau-2}}[\Vert \theta^t_{i, \tau-1} - \theta^t\Vert^2]  \\
    &\overset{(a)}{=} (1+\frac{1}{2E-1})\mathbb{E}_{\xi_{i, \tau-2}}[\Vert \theta^t_{i, \tau-1} - \theta^t\Vert^2] +2E\eta_{1, t}^2 \mathbb{E}_{\xi^t_{i, [\tau-2]}}[\Vert 
    \nabla_{\theta} f_i(\theta_{i, \tau-1}^t;\BFc_i^t)
    \Vert^2] + 2E\eta_{1, t}^2\sigma_1^2\\
    &\leq  (1+\frac{1}{2E-1} + 4E\eta_{1, t}^2L_1^2)\mathbb{E}_{\xi_{i, \tau-2}}[\Vert \theta^t_{i, \tau-1} - \theta^t\Vert^2]  + 4E\eta_{1, t}^2\Vert\nabla_{\theta}f_i(\theta^t;\BFc_i^t)\Vert^2+ 2E\eta_{1, t}^2\sigma_1^2\\
    &\leq (1+\frac{2}{E})\mathbb{E}_{\xi_{i, \tau-2}}[\Vert \theta^t_{i, \tau-1} - \theta^t\Vert^2] + 4E\eta_{1, t}^2\Vert \nabla_{\theta}f_i(\theta^t;\BFc_i^t)\Vert^2 + 2E\eta_{1, t}^2\sigma_1^2,
    \end{aligned}
\end{equation*}
where we use the $\mathbb{E}[\Vert\BFx\Vert^2] = \mathbb{E}[\Vert \BFx - \mathbb{E}[x]\Vert^2] + \Vert\mathbb{E}[\BFx]\Vert^2$ in the equality $(a)$, and the last inequality is induced by $2E-1\geq E$ and $\eta_{1, t}\leq \frac{1}{2EL_1}$.

Let $\mu := 4E\eta_{1, t}^2\Vert \nabla_{\theta}f_i(\theta^t;\BFc_i^t)\Vert^2 + 2E\eta_{1, t}^2\sigma_1^2$. Unrolling the above inequality with the fact that $\theta_{i, 0}^t = \theta^t$, we obtain
\begin{equation}
    \mathbb{E}_{\xi_{i, \tau-1}}[\Vert \theta^t_{i, \tau} - \theta^t\Vert^2] 
    \leq \mu\sum_{s=0}^{\tau -1}(1+\frac{2}{E})^s 
    = \frac{E}{2}\mu (1+\frac{2}{E})^{\tau} 
    \leq \frac{E}{2}\mu (1+\frac{2}{E})^{E} 
    \leq 4E\mu
\label{eq:result of lemma 4}
\end{equation}
where the last inequality uses $(1+a/b)^t \leq e^a$ and $e^2 \leq 8$. Hence, with Eq. \eqref{eq:result of lemma 4}, we can conclude the result by summing over $\tau$.\Halmos
\endproof

\begin{corollary}
\label{cor:rate with ADMM}
    Under the same assumptions in Theorem 2,
    if $\sigma_1 = 0$, $\sigma_2=0$, and $\delta=0$, i.e., using the full data set to compute the gradient under the homogeneous setting,
    considering the constant learning rate with $\eta_t < \min\{\frac{1}{4L_1}, \frac{2}{L_2}, \frac{1}{32EL_1}\}$,  we obtain 
    $
        \mathbb{E}[\Vert \bar{G}\Vert^2] = \mathcal{O}\left(\frac{F(\mathbf{z}^0) - F^*}{T}\right).
    $
\end{corollary}
\begin{remark}
    The convergence rate established in Corollary \ref{cor:rate with ADMM} is directly derived from Theorem 2 by setting $\sigma_1 = \sigma_2 = \delta=0$ and employing a sufficiently small learning rate to ensure the denominator remains positive.
    Notably, this convergence rate is consistent with the one obtained by the \texttt{FedADMM} algorithm proposed by \citet{zhou2023federated}, which establishes convergence under the same conditions.
    However, \texttt{FedADMM} is designed to address the standard separable federated learning objective function, $\min_{\theta}\sum_{i=1}^M p_i f(\theta)$.
    It cannot be directly applied to problem in Eq. (4) due to the non-separable Laplacian regularization term for $\rvc_i$. 
    Adapting \texttt{FedADMM} to address problem in Eq. (4) would require additional techniques \citep{tuck2021distributed} and the design of new communication protocols. 
    Furthermore, in contrast to computing a given number of gradients as in Algorithm 1 in the main text, the \texttt{FedADMM} method necessitates finding the optimal value of $\theta$, which can be non-trivial and computationally demanding. 
    Thus, the \texttt{FedADMM} algorithm requires modifications to address the proposed problem.
\end{remark}

\subsection{Connection with General Linear Mixed Model}
\label{subsec:proof of thm1}

\proof{Proof of Theorem 1:} 
Following the formulation $\mathbb{E}[y] = g^{-1}(\BFx^\top \BFb_f+\BFx^\top \BFb_r)$, the vector of random effect $\BFb_r$ is generated from a normal distribution $\mathcal{N}(0,G)$.
For the GLMM problem, the objective function combines the negative log-likelihood and the log prior of random effect. Using $f_i$ as the negative log-likelihood function of the linear predictor, the objective function is 
$
     \sum_{i=1}^M \frac{n_i}{n} f_i(\BFx^\top \BFb_f+\BFx^\top \BFb_{ir}) + \sum_{i=1}^M \BFb_{ir}^\top G^{-1}\BFb_{ir}.
$

Let $\BFb_i = \BFb_f + \BFb_{ir}$. Rearranging the above formula gives
\begin{equation}
    \label{eq:the GLMM objective function}
    \sum_{i=1}^M \frac{n_i}{n} f_i(\BFx_i^\top\BFb_i) + \sum_{i=1}^M (\BFb_{i} -\BFb_{f})^\top G^{-1}(\BFb_{i}-\BFb_{f}).
\end{equation}

According to Eq. (7), the linear predictor is $\BFx^{\top}\theta\BFc$ and let $\theta\BFc_i=\BFb_i'$. Plugging $\BFc_i=\theta^{+}\BFb_i'$ into Eq. (4), we obtain
$
    \sum_{i=1}^M \frac{n_i}{n}f_i(\BFx_i^\top\BFb_i') + \frac{\lambda}{2} \sum_{i=1}^M\sum_{j=1}^M \Vert \theta^{+}\BFb_i' - \theta^{+}\BFb_j'\Vert^2,
$
where $\theta^{+}$ denotes the pseudo-inverse of $\theta$, $\theta^{+}\BFb_i'=(\theta^\top\theta)^{-1}\theta^{\top}\BFb_i'$.

Rearranging the regularization term, we obtain
\begin{equation}
    \label{eq:similar objective function as MEM}
\sum_{i=1}^M \frac{n_i}{N}f_i(\BFx_i^\top \BFb_i') + \lambda M\sum_{i=1}^M[(\BFb_i'-\BFb_0')^\top\Lambda^{-1}(\BFb_i'-\BFb_0')],
\end{equation}
where $\BFb_0'=\frac{\sum_{i=1}^M \BFb_i'}{M}$ and $\Lambda^{-1}=(\theta^+)^\top\theta^+$. Comparing Eqs. \eqref{eq:the GLMM objective function} and \eqref{eq:similar objective function as MEM}, when $\lambda=\frac{1}{M}$, the objective function of GLMM is equivalent to the objective function of the proposed method.\Halmos
\endproof

In Theorem 1, the identical values of $w_{ij}$ match the fundamental assumption in GLMM: clients' random coefficients are drawn from a common distribution.
While, in \texttt{PPFL}, the unrestricted value of $w_{ij}$ enables it to model realistic relationships among clients, demonstrating the generality and flexibility of \ours over the personalized variant GLMM.

Note that the original equation of GLMM has been simplified, and we have $\mathbb{E}[y]=g^{-1}(\mathbf{x}_b^\top \mathbf{b}_f + \mathbf{x}_r^\top \mathbf{b}_r)$ in most cases, where $\mathbf{x}_f$ and $\mathbf{x}_r$ are distinct fixed and random features respectively. 
For this case, let $\mathbf{x} = [\mathbf{x}_b;\mathbf{x}_r]$ and $\mathbf{b_i} = [\mathbf{b}_f; \mathbf{b_r}]$, and we can still obtain the same form as Eq. \eqref{eq:the GLMM objective function} by expanding and padding the corresponding matrices.
Accordingly, since the coefficient of $\mathbf{x}_f$ is the same for all clients, we can use the consistent constraint on part of the components of the membership vectors (i.e., $\mathbf{c}_{ik} = \mathbf{c}_{jk},\, \forall i, j \in [M]$), where the number of fixed components is the same as the dimension of $\mathbf{x}_f$.
Hence, the equivalence still holds and this simplification does not affect the final result.

Finally, to justify the specification $G=\theta\theta^\top$ in Theorem 1, we leverage the fundamental linear algebra result that any positive definite covariance matrix $G$ in GLMMs admits an eigendecomposition $G = \hat{\theta}\hat{\theta}^\top$ with $\hat{\theta}$ being orthogonal. 
In the ideal case where $K = d$, $\theta$ serves as a decomposition of $G$, and the canonical models are orthogonal to each other.
When $K < d$, the canonical models $\{\theta_k\}_{k=1}^K$ can be viewed as a rank-$K$ approximation of the principal subspace spanned by the eigenvectors of $G$ associated with the $K$ largest eigenvalues.

\section{The variant of \ours’s algorithm}

\label{sec:variant of ppfl}

\begin{algorithm}[t]
\caption{The variant of \texttt{PPFL} algorithm: alternate the update of two blocks.}
\label{alg:variant ppfl}
\SetKwFunction{LGtheta}{Local Gradient$(\theta^t, \eta_1)$}
  \SetAlgoLined
  \KwIn{the initial state $\BFz^0$, the number of canonical models $K$, the number of communication rounds $T$, the number of local steps $E$, the step sizes $\eta$, $\eta_1 = \eta / E$, $\eta_2 = \eta$, and the Laplacian penalty $\BFL$ with its hyperparameter $\lambda$.}
  \For {$t=0,1,\dots,T$} {
  The server broadcasts $\theta^t$ and the vector $\lambda(\BFL\BFC^t)_i$ to client $i$, $i\in [M]$\;
  \For {$i=1,\dots, M$} {
  The client updates the membership vector $\BFc_i^{t+1}$ with the step size $\eta_2$ and $\lambda(\BFL\BFC^t)_i$ based on Eq. (16)\;
  \For {$\tau = 0, \dots, E-1$} {
  Set $\theta_{i, 0}^t = \theta^t$\;
  $\theta_{i, \tau+1}^t = \theta_{i, \tau}^t - \eta_1\nabla_{\theta} f_i(\theta^t;\BFc_i^{t+1}, \xi_{i, \tau}^t)$\;
  }
  The client uploads the accumulated gradient $\theta_{i,E}^t - \theta^t$ and $\BFc_i^{t+1}$ to the server\;
  }
  The server updates $\theta^{t+1}$ based on Eq. (12) and substitutes $\BFC^t$ with $\BFC^{t+1}$\;
  }
  \KwOut{$\theta^T, \BFC^{T}$}
\end{algorithm}

In Algorithm \ref{alg:variant ppfl}, we introduce a communication-efficient variant of Algorithm 1 when the full gradient $\nabla f_i(\mathbf{c}_i^t;\theta)$ is utilized. 
Unlike the random block update in each round of Algorithm 1, Algorithm \ref{alg:variant ppfl} places the update of $\mathbf{c}_i$ in client $i$, prescribes a sequential updating order for the two blocks, and alternately updates them in each round.

Specifically, at the beginning of each round $t$, the server broadcasts $\lambda (\mathbf{LC}^t)_i$ as well as $\theta^t$ to client $i$, where $(\mathbf{LC}^t)_i$ is the $i$-th row of the matrix $\mathbf{LC}^t$.
With $(\mathbf{LC}^t)_i$ and the full gradient $\nabla f_i(\mathbf{c}_i^t;\theta)$, the client first gets $\mathbf{c}_i^{t+1}$ based on Eq. (16). Then, conditioned on $\mathbf{c}_i^{t+1}$, the client executes the same local update of $\theta$ as in Algorithm 1. 
Finally, the clients transmit the accumulated gradient $\theta_{i,E}^t - \theta^t$, along with the updated $\mathbf{c}_i^{t+1}$, to the server to update $\theta^{t+1}$ and $\mathbf{C}^{t+1}$. 
By prescribing the order of updates and alternately updating these blocks, the final synchronization step at the end of each round in Algorithm 1 can be discarded, which improves communication efficiency.

\section{Connections with Existing Methods}

\subsection{Connections with Existing PFL Methods}
\label{sec:connections with existing PFL}

\begin{figure}[t]
    \caption{Illustrative diagram. (a) The replica of $\theta_{\text{com}}$ and the $k$-th canonical model constitute the $k$-th group-specific model in clustered FL; (b) $\theta_{\text{com}}$ and $\{\theta_k\}_K$ remain consistent across clients and are packed together to serve as the shared global module in the decoupling PFL.}
    \centering
    \subfloat[]{\label{fig:CFL}\includegraphics[width=0.34\linewidth]{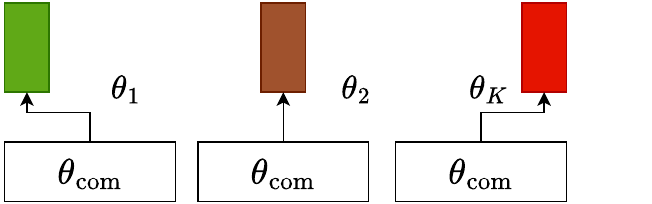}}\hspace{5pt}
    \subfloat[]{\label{fig:partial}\includegraphics[scale=0.4]{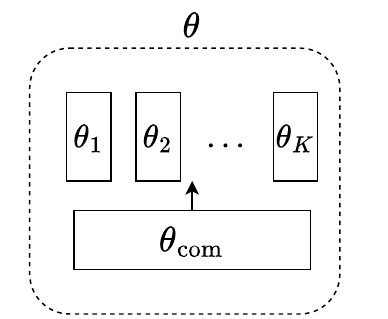}}\hspace{5pt}
    \label{fig:virtual form}
\end{figure}

\textbf{Connections with clustered FL}. 
With an appropriate model structure, we show that the clustered FL is included in the \texttt{PPFL} framework. 
Specifically, as shown in Figure~\ref{fig:CFL}, we create a virtual structure by incorporating a replica of $\theta_{\text{com}}$ into each canonical model, and the combination can be considered as the cluster-specific model in clustered FL.
If $\lambda=0$ and the \textit{max rule} (i.e., $\arg\max_k c_{ik}$) is further applied to find the best cluster for each client, then the original objective function in Eq. (4) becomes as follows:
\begin{equation}
\begin{aligned}\label{eq7}
\min_{\{\theta_k\}_{k=1}^K, \{\BFc_i\}_{i=1}^M} \sum_{k=1}^K\sum_{i\in \mathcal{S}_k} p_i f_{i}(\theta_k),
\end{aligned}
\end{equation}
where, with a slight abuse of notation, $\theta_k$ represents the combination of the $k$-th canonical model and $\theta_{\text{com}}$, and the set $\mathcal{S}_k:=\{i\vert \arg\max_k c_{ik} = k,i\in[M]\}$.

Further, if $\mathbf{c}_{k}$ is treated as non-trainable and determined solely based on the corresponding losses, e.g., ${c}_{ ik} = \frac{1/f_i(\theta_k)}{\sum_{j=1}^K 1/f_i(\theta_j)}$, then Eq. \eqref{eq7} reduces to the conventional model of clustered FL \citep{ghosh2020efficient}, where clients are assigned to the cluster with the lowest loss in each round.
Compared with the objective function in Eq. (1), the formulation in Eq. \eqref{eq7} and the problem in Eq. (4) also include additional variables $\{\mathbf{c}_{i}\}_{i=1}^M$, parameterizing the client identification process.
Specifically, ${c}_{ik}$ captures the alignment between client $i$ and the $k$-th behavior pattern model.
With the help of the variable $\mathbf{c}_i$, the proposed \texttt{PPFL} preserves the continuity of the optimization process, potentially leading to less time for convergence.
Moreover, the proposed \texttt{PPFL} in Eq. (4) enables clients to have alignments with multiple canonical models. 
This allows for capturing within-cluster heterogeneity and customizing each client's model, in contrast to the single-cluster partition in clustered FL.

\begin{algorithm}[t]
\caption{The local gradient computation}
\label{alg:local gradient}
\SetKwBlock{LGtheta}{\textbf{Local Gradient}$(\theta^t, \eta_{1, t})$}{}
\SetKwBlock{LGC}{\textbf{Local Gradient}$(\BFC^t, \eta_{2,t})$}{}
\LGtheta{
 \For{$i=1,2,\dots, M$}{
    Set $\theta_{i, 0}^t = \theta^t$\;
    \For{$\tau = 0,\dots, E-1$}{
    $\theta_{i, \tau+1}^{t} = \theta_{i, \tau}^{t} - \eta_{1,t} \nabla_{\theta} f_i(\theta_{i, \tau}^t; \BFc_i,\xi_{i, \tau}^t)$\;
    }
    Upload the local accumulated gradient $\theta_{i, E}^t - \theta^t$ to the server;
 }
}
\textbf{Return:} the weighted sum of local accumulated gradients $\sum_{i=1}^M p_i (\theta_{i, E}^t - \theta^t)$.

\LGC{
 \For{$i=1,2,\dots, M$}{
 Compute the stochastic gradient $p_i\nabla_{\BFc}f_i(\BFc_i^t;\theta, \xi_i^t)$\;
 Upload the stochastic gradient $p_i\nabla_{\BFc}f_i(\BFc_i^t;\theta, \xi_i^t)$ to the server\;
 }
 The server concatenates the uploaded gradients as \begin{equation*}
     g(\BFC^t) = [p_1\nabla_{\BFc}f_1(\BFc_1^t;\theta, \xi_1^t)^T, \cdots,
 p_M\nabla_{\BFc}f_M(\BFc_M^t;\theta, \xi_M^t)^T]^T;
 \end{equation*}
}
\textbf{Return:}  $g(\BFC^t) + 2\lambda\BFL\BFC^t$.
\end{algorithm}

\textbf{Connections with multi-task PFL}. With the second architecture in Figure 2b, let the personalized model $\theta_i$ in client $i$ comprise both $\theta_{\text{com}}$ and the interpolated term $\sum_{k=1}^Kc_{ik}\theta_k$, i.e., $\theta_i = [\theta_\text{com}; \sum_{k=1}^Kc_{ik}\theta_k]$, where the concatenated parameter $\theta_i$ does not change the dimension of the output. Plugging $\theta_i$ into the objective function in Eq. (4) leads to:
\begin{equation}
\begin{aligned}
\label{eq:multi-task connection}
\min_{\{\theta_i\}_{i=1}^M} \quad & \sum_{i=1}^M p_i f_i(\theta_i) + \frac{\lambda}{2}\sum_{i=1}^M\sum_{j=1}^M w_{ij}\Vert\BFc_i -\BFc_j \Vert^2.\\
\end{aligned}
\end{equation}

Compared to the problem in Eq. (2) with Laplacian regularization, \texttt{PPFL} imposes regularization on $\BFc_i$ instead of $\theta_i$.
Since the values of $\theta_{\text{com}}$ are consistent across clients as well as $\{\theta_k\}_{k=1}^K$, it is reasonable to conduct the pairwise comparison of the various membership vectors.
While, if plugging $\theta_i = [\theta_\text{com}; \sum_{k=1}^Kc_{ik}\theta_k]$ into the objective function in Eq. (2), we can still derive the similar formulation as Eq. \eqref{eq:multi-task connection}, which demonstrates the equivalence between these two objective functions.
Moreover, instead of a similar objective function, \texttt{PPFL} can alleviate the flaws of multi-task PFL.
Since the $K$-dimensional vector has a relatively smaller size than $\theta_{i}$, the reduced regularization of \texttt{PPFL} in Eq. \eqref{eq:multi-task connection} can alleviate the mismatch problem during the calculation of client similarity \citep{jeong2022factorized}.
Besides, the model parameter-sharing mechanism and the low-dimensional client-specific vector in \texttt{PPFL} can reduce the data requirement for local training and mitigate the issue of insufficient local data \citep{Liang2020ThinkLA}.

\textbf{Connections with decoupling PFL}
Since both $\theta_{\text{com}}$ and $\{\theta_k\}_{k=1}^K$ are shared among clients, as depicted in Figure~\ref{fig:partial}, we can use $\theta$ to represent the collection of them, regardless of which architecture is applied.
Then, we reformulate the original objective function in Eq. (4) as:
\begin{equation}
\begin{aligned}
\min_{\theta,\{\mathbf{c}_i\}_{i=1}^M} \quad & \sum_{i=1}^M p_i f_i(\theta, \mathbf{c}_i) + \frac{\lambda}{2} \sum_{i=1}\sum_{j=1}w_{ij}\Vert \BFc_i - \BFc_j\Vert^2,
\end{aligned}
\label{eq:conn_dPFL}
\end{equation}
which matches the objective function in decoupling PFL in Eq. (3) when $\lambda = 0$.
 
Despite the similar objective function, \texttt{PPFL} and the decoupling PFL are guided by distinct underlying principles.
The decoupling PFL emphasizes that the local-reserved module can specialize in the local data distribution without communication.
Moreover, by incorporating Laplacian regularization, clients can utilize other clients' membership vector information, which promotes collaboration among clients and enhances the generalization performance.

\subsection{Other PFL Methods}
In addition to the three PFL approaches presented in the main content, several other methods have been proposed to achieve model personalization.
One notable direction is the Bayesian PFL, which assumes that the clients' personalized parameters follow different distributions.
This approach incorporates Bayesian learning principles into the FL framework.
For example, \citet{kotelevskii2022fedpop} and \citet{li2022federated} utilize Mixed Effect Models (MEM) based on the Bayesian framework to learn the client-specific random effects for personalization.
To estimate the distributions of personalized parameters, Bayesian PFL approaches employ techniques such as variational inference \citep{zhang2022personalized} and Markov chain Monte Carlo sampling \citep{kotelevskii2022fedpop}.
Bayesian PFL approaches have demonstrated promising results on small-scale data sets \citep{zhang2022personalized}.
However, they often suffer from high computational costs and low communication efficiency \citep{cao2023bayesian}, which limits their practical applicability.
Another line of research focuses on data-based PFL methods.
These approaches employ techniques such as data augmentation \citep{duan2020self} and client selection \citep{chai2020tifl} to reduce the heterogeneity among client data distributions.
By alleviating data heterogeneity, these methods can potentially improve the generalization performance of the global model across clients.

\subsection{Connection with Federated Matrix Factorization Problems}
\label{append:matrix factorization}

In this section, we investigate the relationship between \ours and the federated matrix factorization problem, focusing on their respective contributions to constructing personalized models.
We discuss both the similarities and distinctions between them, providing insights into how each approach models personalization.

Matrix factorization is a widely adopted technique with applications across diverse fields, particularly in recommendation systems \citep{li2021federated, zhang_lightfr_2023}. 
Fundamentally, matrix factorization involves decomposing a given matrix, $\rmA \in \R^{p \times q}$, into the product of two lower-dimensional matrices, $\rmU \in \R^{p \times r}$ and $\rmV \in \R^{q \times r}$, with $r<\min\{p, q\}$, such that $\rmA \approx \rmU\rmV^\top$. This decomposition relies on the assumption that the original matrix $\rmA$ has an inherent low-rank structure or latent patterns.
For example, in a movie recommendation system, $\rmA$ may represent user ratings for various movies. 
In this context, $\rmV$ represents user preferences across latent factors (e.g., genre, director, and cast), while $\rmU$ reflects the relevance of these factors to each movie.

In FL, matrix factorization can be adapted to model personalization by decomposing the global parameter matrix into a shared matrix $\rmU$ and a client-specific matrix $\rmV$ \citep{jeong2022factorized}. 
The objective function for federated matrix factorization can be expressed as:
\begin{equation*}
    \min_{\rmU, \{\rmV_i\}_{i=1}^M} \sum_{i=1}^M p_i f_i(\rmU, \rmV_i),
\end{equation*}
where this formulation aligns with Decoupling PFL by setting $\theta := \rmU$ and $v_i := \rmV_i$ in Eq. (3) in the main manuscript.

Although the proposed method, \ours, shares a structural resemblance to federated matrix factorization, particularly in its interpolation-based implementation, there are significant differences in their underlying principles and objectives.
To clarify, consider a specific setting where the optimal personalized model for client $i$ is known and denoted as $\theta_i^* \in \mathbb{R}^{p \times q}$.
Let $\Theta \in \mathbb{R}^{pq \times k}$ represent a shared parameter matrix, and the operator $\text{vec}$ flattens a matrix into a vector.
Then, the underlying objective of federated matrix factorization is:
\begin{equation*}
    \min_{\rmU, \{\rmV_i\}_{i=1}^M} \sum_{i=1}^M p_i \Vert \theta_i^* - \rmU\rmV_i^\top\Vert^2,
\end{equation*}
which aims to discover a shared, latent low-dimensional structure across heterogeneous parameters $\theta_i^*$, as well as client-specific preferences.
In contrast, following the idea of \ours, we aim to identify basis models that span the space of personalized models $\theta_i^*$ with the following objective function:
\begin{equation*}
    \min_{\Theta, \{\rvc_i\}_{i=1}^M} \sum_{i=1}^M p_i \Vert \text{vec}(\theta_i^*) - \Theta\rvc_i\Vert^2.
\end{equation*}
Notably, the dimensions of the optimized parameters in the two objective functions above differ: in the federated matrix factorization problem, both $\rmU \in \R^{m\times r}$ and $\rmV_i\in \R^{n\times r}$ are typically matrices, while in \ours, $\Theta\in \R^{mn\times K}$ is a matrix and $\rvc_i\in\R^{K}$ is a vector.
Consequently, federated matrix factorization seeks to minimize the reconstruction error by finding a shared representation and client-specific preferences across the heterogeneous parameters, while \ours aims to derive a set of basis models from the heterogeneous clients.

Compared to federated matrix factorization, a notable advantage of \ours is its interpretability.
Specifically, the membership vector $\rvc_i$ provides an intuitive representation of how the canonical models are combined to form client $i$'s personalized model, as illustrated in the experiments.
In contrast, interpreting the matrix $\rmV_i$ in federated matrix factorization is particularly challenging, especially in complex neural network settings. 

\subsection{Example: Contrasting Objective Formulations}
As outlined in Section 2.2.1 of the main text, we first present the detailed prediction formulas for the standard linear regression and logistic regression as follows:
\begin{itemize}
    \item \textbf{Linear Regression}: In the linear regression model, the link function $g$ is the identity function. Then, Eqs. (7) and (8) lead to the same prediction $\hat{y}_{il}=\sum_{k=1}^Kc_{ik}\BFx_{il}^\top\theta_k=\BFx^\top_{il}\theta\BFc_i$ for the $l$-th data point in client $i$.
    \item \textbf{Logistic Regression}: For a binary classification problem, the link function $g$ is the logit function. In this instance, the predictions $\hat{y}_{il}$ obtained from Eqs. (7) and (8) are different, denoted as $\hat{y}_{il}=\sum_{k=1}^K c_{ik}\frac{1}{1+\exp(-\BFx_{il}^\top\theta_k)}$ and $\hat{y}_{il}=\frac{1}{1+\exp(-\BFx_{il}^\top\theta\BFc_i)}$ respectively.
\end{itemize}

Typically, Eqs. (7) and (8) yield different outcomes with their own implications.
In Eq. (7), the client membership vector $\BFc_i$ scores the predictions of these canonical models, and the weighted sum is taken as the final prediction.
In contrast, Eq. (8) first constructs the personalized model spanned by $\{\theta_k\}_{k=1}^K$, which is then plugged into the formulation of the GLM.

We now contrast the objective functions of \citet{feng2020learning} and \ours, using logistic regression as an illustrative example.
In the collaborative field, \citet{feng2020learning} propose a collaborative logistic regression model that is formulated as:
\begin{equation}
    \label{eq:feng}
    \begin{aligned}
        \min_{\theta,\{\BFc_i\}_{i=1}^M} \quad &
\sum_{i=1}^M\sum_{l=1}^{n_i} \Big\{y_{il}\log(\frac{1}{1+\exp(-\mathbf{x}_{il}^\top\theta\mathbf{c}_i)}) + (1-y_{il})\log(\frac{1}{1+\exp(\mathbf{x}_{il}^\top\theta\mathbf{c}_i)}) \Big\}+ \frac{\lambda}{2}\sum_{i, j}w_{ij}\Vert \mathbf{c}_i - \mathbf{c}_j\Vert_2^2,\\
\textrm{s.t.} \quad & \BFone^\top\BFc_i=1,\; \forall i\in [M],\\ 
  &\BFc_i \geq 0,\; \forall i \in [M].
    \end{aligned}
\end{equation}    
Compared with the original objective function in Eq. (4) in the main text, this formulation \eqref{eq:feng} is a special case of the proposed \texttt{PPFL} method adopting the interpolation form in (Eq. (12) in the main text). 
Note that \texttt{PPFL} can also adopt the form of the weighted sum of predictions (Eq. (11) in the main text) as
\begin{equation*}    
\begin{aligned}
\min_{\theta,\{\BFc_i\}_{i=1}^M} \quad &
\sum_{i=1}^M\sum_{k=1}^K c_{ik}\sum_{l=1}^{n_i}\Big\{y_{il}\log(\frac{1}{1+\exp(-\mathbf{x}_{il}^\top\theta_k)}) + (1-y_{il})\log(\frac{1}{1+\exp(\mathbf{x}_{il}^\top\theta_k)}) \Big\}+ \frac{\lambda}{2}\sum_{i, j}w_{ij}\Vert \mathbf{c}_i - \mathbf{c}_j\Vert_2^2,\\
\textrm{s.t.} \quad & \BFone^\top\BFc_i=1,\; \forall i\in [M],\\ 
  &\BFc_i \geq 0,\; \forall i \in [M].
\end{aligned}
\end{equation*}

\section{Experimental Setup}

\label{appendix:experiments}

In this section, we present the setup of our experiments, including the details of data sets and models, the experimental pipeline, and the selection of hyperparameters. 

\subsection{Data sets}
\label{subsec:datasets}
Here, we elaborate on the details of all the data sets used in the experiments.
\begin{itemize}
    \item MNIST. We randomly divide $10$ digit classes into four groups in the MNIST data set, where two groups contain $3$ digit classes each, and the other two contain $2$ digit classes each.
    We collect all samples for each group with the contained digit classes as the group-specific data set.
    Besides, each group-specific data set is randomly and evenly partitioned into $25$ parts as clients' local data sets;
    \item CIFAR-10. To simulate the domain-heterogeneous scenario, classes with similar types are grouped in the CIFAR-10 data set.
    There are three groups: the first group contains $6$ animal classes, the second group contains automobile and truck classes, and the last group contains the others.
    Similarly, each group-specific data set is created by collecting all samples with the contained labels and is randomly and evenly partitioned into $26$ parts as clients' local data sets (the first group is divided into $27$ parts);
    \item Synthetic. Firstly, we generate three regression coefficients, each of whose elements follows the same uniform distribution.
    Thus, there are three group-specific data sets in the synthetic data set, and the number of data points is identical.
    Secondly, for each group-specific data set, we generate a prior for the number of data points among clients' data sets from a Dirichlet distribution.
    Finally, based on the prior, we generate the actual number of data points in clients' local data sets from each group-specific data set following a multinomial distribution.
    The generation process of the synthetic data set is in accordance with \citet{marfoq2021federated};
    \item Digits. The data set, created by \citet{li2020fedbn}, consists of 5 distinct sub-datasets: SVHN \citep{netzer2011reading}, USPS \citep{hull1994database}, SynthDigits \citep{ganin2015unsupervised}, MNIST-M \citep{ganin2015unsupervised}, and Sub-MNIST \citep{lecun1998gradient}.
    Each sub-dataset has been processed to contain $7348$ samples.
    As illustrated in Figure \ref{fig:5-digits}, these sub-datasets exhibit heterogeneous visual characteristics and diverse feature distributions, while sharing the same set of labels and a similar label distribution.
    Following \citet{li2020fedbn}, we treat each sub-dataset as an individual client (resulting in a total of 5 clients) to evaluate the performance of \ours under the feature shift scenario.
    \item EMNIST. The data set is created by \citet{caldas2018leaf} and can be found in the TensorFlow Federated package. 
    With the same selection criterion as \citet{pillutla2022federated}, we select $1114$ clients and clients' local data sets contain digits/characters written by themselves;
    \item StackOverflow. The data set is a collection of the answers and questions of users in StackOverflow and is provided by the Tensorflow Federated package.
    Similarly, we evaluate the performance of the same clients as \citet{pillutla2022federated}.
    Besides, the vocabulary only contains the top 10,000 most frequent words in the data sets.
\end{itemize}
Table \ref{tab:summary of dataset} provides the number of samples and the dimension of features for each data set. For the StackOverflow data set, the embedding size is reported as the dimension of features.

\begin{table}[t]
\centering
\caption{Summary of the datasets used in the experiments.}
\label{tab:summary of dataset}
{\begin{tabular}{ccc}
\hline
Data set      & \# Samples & \# Features \\ \hline
MNIST         & $60000$             & $784$                \\
CIFAR-10 & $60000$ & $3072$ \\
Synthetic     & $72758$             & $150$                \\
Digits & $37190$ & $784$ \\
EMNIST        & $814255$            & $784$                \\
StackOverflow & $4964000$ & $10000$\\ \hline
\end{tabular}}
{}
\end{table}

\subsection{Models}
Regarding the employed models, we adopt the same models as those in \citet{marfoq2021federated} and \citet{pillutla2022federated} for pathological and practical data sets, respectively. 
Moreover, we use the personalized output layer \citep{pillutla2022federated} for the decoupling method in practical data sets.
Based on these models, we implement the canonical models as follows:
\begin{itemize}
    \item For the MNIST data set, the canonical model is the two-layer perceptron;
    \item For the CIFAR-10 data set, the canonical model is the last fully connected layer;
    \item For the synthetic data set, the canonical model corresponds to different regression coefficients;
    \item For the EMNIST data set, the canonical model is the last fully connected layer;
    \item For the StackOverflow data set, the canonical model is the last decoder layer, which is the vocabulary table to give the predictive words.
\end{itemize}

\subsection{Experimental pipeline}

First, with a fixed random seed, we generate the data set following the instructions in Section \ref{subsec:datasets}.
Afterward, we generate the parameter $\theta$ for the server with the same seed in all methods and broadcast it to all clients. 
Meanwhile, the parameter $\BFc_i$ is initialized as $[\frac{1}{K}, \dots, \frac{1}{K}]\in \mathbb{R}^{K}$, $i \in [M]$.
Besides, for the decoupling method, the parameters of personalized layers are randomly generated across clients.
Additionally, we collect the frequency of different labels in clients' local data sets and then calculate the pairwise cosine similarity to generate the matrix $\BFW$. 
Finally, we repeat the experiment five times with different initial values of $\theta$ given the random seeds, and then the average accuracy is reported.

In pathological data sets, our method uses the same training pipeline as \citet{marfoq2021federated}.
While in practical data sets, we omit the personalized federated training process \citep{pillutla2022federated} and directly train the shared parameter $\theta$ and the membership vectors $\BFc_i$ together to ensure the consistency between $\theta$ and $\BFc_i$.

\subsection{Hyperparameters}

For each data set, the number of communication rounds, the batch size, the learning rate for $\theta$, the number of selected clients per round, the number of local steps, and the learning rate scheduler are the same for all methods. Moreover, we set these hyperparameters to be identical to those in \citet{marfoq2021federated} and \citet{pillutla2022federated}.

The hyperparameter $\lambda$ in our methods is selected from the set $\{10^{-5}, 10^{-4}, 10^{-3}, 10^{-2}, 10^{-1}\}$ in all data sets, and we tune the number of canonical models $N$ from the set $\{2, 5, 7, 10\}$ in practical data sets. The selected values are the following:
\begin{itemize}
    \item For all pathological data sets, we set the number of canonical models $N$ to be equal to the number of underlying groups and the coefficient $\lambda = 10^{-5}$;
    \item For EMNIST data set, the coefficient $\lambda = 10^{-4}$ and $N=10$;
    \item For StackOverflow data set, the coefficient $\lambda=10^{-5}$ and $N=5$.
\end{itemize}

\section{Additional Results}

\subsection{Time Consumption}
We carry out the experiment on 2080Ti GPUs and report the average time consumption of each method on practical data sets. 
As shown in Table \ref{tab:time comparison}, for the CIFAR-10 and Synthetic data sets, the time taken by our method is slightly more than that of \texttt{FedAvg} method and is much less than the \texttt{FedEM} method. 
For Mnist, due to the absence of $\theta_{\text{com}}$, our method takes about twice as much time as the \texttt{FedAvg} method.
However, compared to \texttt{FedEM}, our method is still efficient.
Besides, cutting down the number of copies of a large model may lead to less execution time of the local method on the CIFAR-10 data set.

\begin{table}[t]
\centering
\caption{Average execution time of all methods on the pathological datasets.}
\label{tab:time comparison}
\begin{tabular}{lrrrrrrr}
\hline
          & Local   & FedAvg & pFedMe & FedEM   & Clustered FL & FedLG  & PPFL    \\ \hline
Mnist     & 1h10min & 1h7min & 1h8min & 4h50min & 1h10min      & 1h3min & 2h20min \\
CIFAR-10   & 1h5min  & 2h6min & 2h7min & 7h33min & 2h20min      & 2h     & 2h33min \\
Synthetic & 16min   & 17min  & 17min  & 40min   & 18min        & 17min  & 22min   \\ \hline
\end{tabular}
\end{table}

\subsection{Theoretical Communication Comparison between \texttt{PPFL} and \texttt{FedEM}}
\label{sec:communication comparison}

In this section, we provide a comparison of the communication costs between \ours and \texttt{FedEM}.
We analyze both the number of floats sent per round and the total communication cost, demonstrating that our method is more communication-efficient due to its common model $\theta_{\text{com}}$ for the canonical models.

Let $d$ and $s$ denote the dimension of the model and canonical model, respectively.
Since canonical models typically comprise one or two fully connected layers, the dimension $s$ is generally much smaller than that of the model $\theta$ ($s \ll d$).

Regarding the number of floats sent per round, since \texttt{FedEM} \citep{marfoq2021federated} has to broadcast all $K$ entire models to clients, the number of floats sent per round of \texttt{FedEM} is $\mathcal{O}(Kd)$.
In contrast, our method's communication cost per round is $\mathcal{O}(d + Ks)$.
Since $s \ll d$, the number of floats sent per round of our method is lower than that of \texttt{FedEM}.

When considering the total communication cost until convergence is reached, \texttt{FedEM} also has the $\mathcal{O}(\frac{1}{\sqrt{T}})$ convergence rate (shown in Theorem 3.2 in \citet{marfoq2021federated}) and incurs a total communication cost as $\mathcal{O}(\frac{Kd}{\epsilon^2})$.
While, the total communication cost of our method is $\mathcal{O}(\frac{d + Ks}{\epsilon^2})$, which is smaller than $\mathcal{O}(\frac{Kd}{\epsilon^2})$ attributed to $s \ll d$.
Hence, \ours has a lower communication cost than \texttt{FedEM}.

\subsection{Additional Results on CIFAR-10 and Synthetic Data Sets}
\label{sec:additonal results}

\begin{figure}[t]
    \centering
    \caption{(a) Heatmaps of the matrix $\BFC$ during training on the CIFAR-10 data set; (b) Heatmaps of the final $\BFC$ matrix for different values of $N$ on CIFAR-10 data set.}
    \subfloat[]{\label{fig:C matrix on CIFAR10}\includegraphics[scale=0.5]{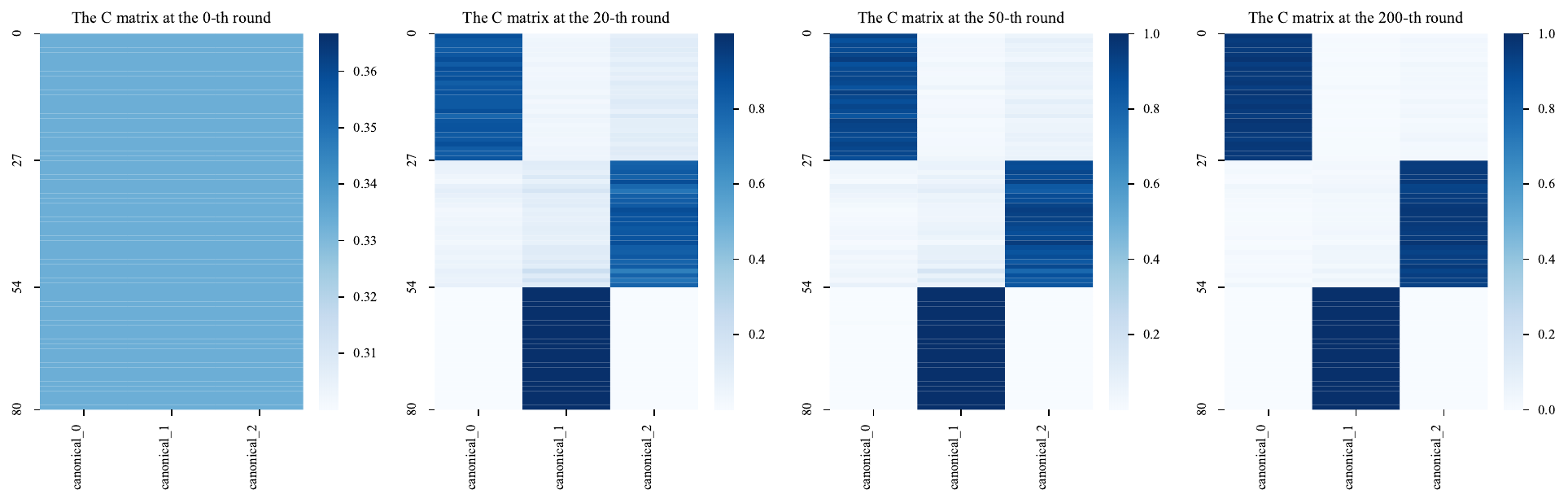}}\\ \vspace{3pt}
    \subfloat[]{\label{fig:compare CIFAR10}\includegraphics[scale=0.5]{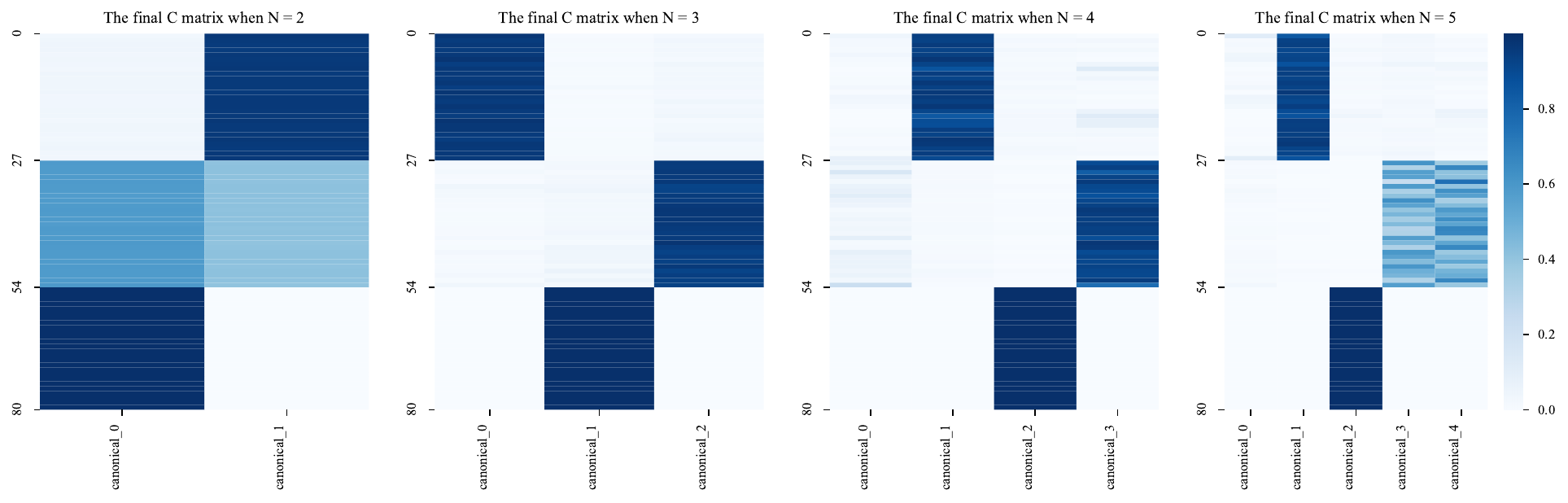}}
    \label{fig:total C CIFAR10}
\end{figure}

Figures \ref{fig:C matrix on CIFAR10} and \ref{fig:compare CIFAR10} present the variation of the matrix $\BFC$ during training when $N=K$ and the final heatmap of $\BFC$ for different $N$, respectively.
In Figure \ref{fig:C matrix on CIFAR10}, each row of the matrix denotes the vector $\BFc_i$ for a specific client, and darker-colored elements have values close to $1$. 
During the training process, $\BFc_i$ gradually converges to a one-hot vector, consistent with the pathological setup.
As shown in Figure \ref{fig:compare CIFAR10}, when $N\geq K$, $\BFc_i$ could identify the client's preference.
These results further validate the effectiveness and interpretability of \ours.

\begin{figure}[t]
\centering
\includegraphics[scale=0.4]{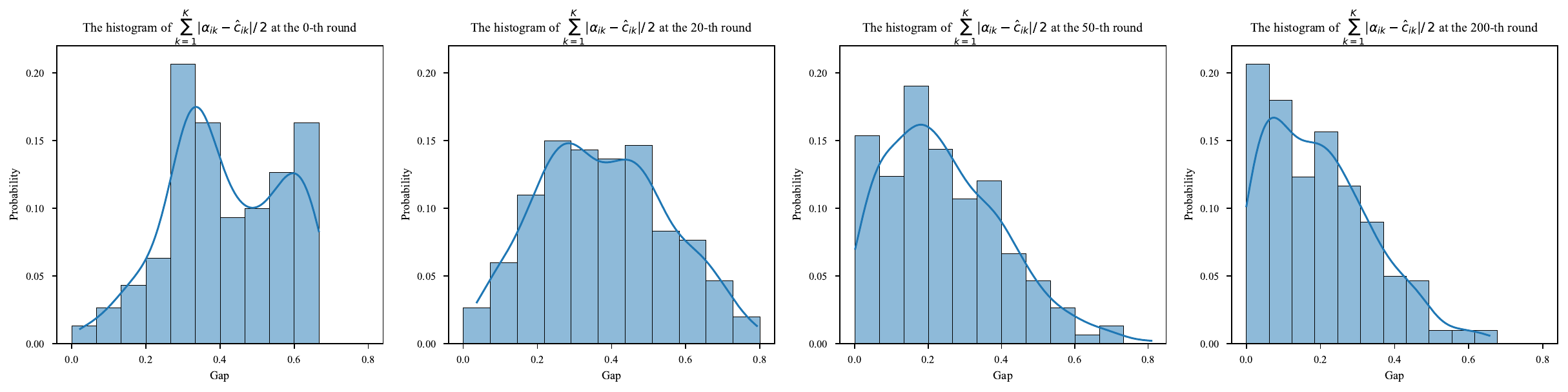}
\caption{The histogram of $\frac{\sum_{k=1}^K \vert \alpha_{ik} - \hat{c}_{ik}\vert}{2}$ at different communication rounds for the Synthetic data set}. 
\label{fig:exhibit of behavior pattern}
\end{figure}

For the Synthetic data set, let the vector $\bm{\alpha}_i$ represent the ground-truth mixture weight in client $i$ and $\hat{\BFc}_i$ be the estimated membership vector.
Generally, we hope $\hat{\BFc}_i$ is close to the mixture weight $\bm{\alpha}_i$ (e.g., if client $i$ has a significant proportion of samples from the $k$-th source, we expect the value of $\hat{c}_{ik}$ to be large).
Then, we employ the difference $\frac{\sum_{k=1}^K \vert \alpha_{ik} - \hat{c}_{ik}\vert}{2}$ to evaluate the gap between the ground-truth mixture weight $\bm{\alpha}_i$ and the learned vector $\hat{\mathbf{c}}_i$.
Thus, the closer the value is to $0$, the more closely the estimated vector $\hat{\mathbf{c}}_i$ approaches the ground-truth weight $\bm{\alpha}_i$. 
As shown in Figure \ref{fig:exhibit of behavior pattern}, the estimated $\hat{\mathbf{c}}_i$ gradually approaches the ground-truth mixture weight $\bm{\alpha}_i$ along with the increase in communication rounds, which demonstrates that \texttt{PPFL} can learn the client's underlying preference in the mixture distribution scenario.

\subsection{Performance on Covariate-shift Data Set} 

\begin{figure}[t]
\centering
\includegraphics[scale=0.4]{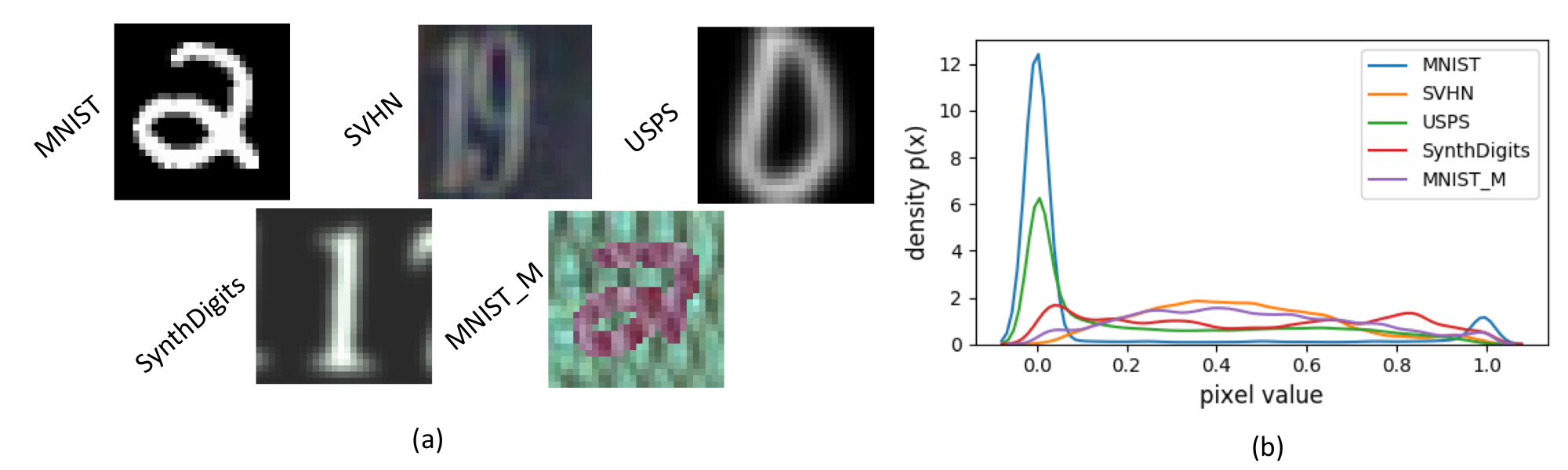}
\caption{Illustration of the Digit dataset from \citet{li2020fedbn}. (a) Examples from each data set (client). (b) Feature distributions across the data sets (over 100 random samples for each data set).
}
\label{fig:5-digits}
\end{figure}

This section investigates the performance of \texttt{PPFL} compared to baseline methods on a covariate-shift data set.
Unlike the domain-heterogeneity setting, where $p(y)$ differs among clients, covariate-shift refers to the situation where the distribution of $p(x)$ varies among clients while the conditional distribution $p(x\vert y)$ remains the same.
Specifically, we utilize the Digit data set as described by \citet{li2020fedbn}.
This data set comprises five sub-datasets: SVHN, USPS, SynthDigits, MNIST-M, and Sub-MNIST.
As shown in Figure \ref{fig:5-digits}, these sub-datasets share the same label distribution but exhibit heterogeneous appearances due to different distributions of pixel values.

According to \citet{li2020fedbn}, we set five clients in this setting, with each corresponding to a component of the Digit data set. 
We present the overall performance on the whole data set as well as the performance on each sub-data set in Table \ref{tab:merged_experiment}. 
The results show that \texttt{PPFL} achieves the best overall performance.
It also achieves the highest test accuracy on four of the five sub-datasets. 
This demonstrates the effectiveness of \texttt{PPFL} in the covariate-shift setting.

\begin{table}[t]
\centering
\caption{Comparison of PPFL with other methods on the covariate-shift data set. The average test accuracy ($\%$) over five independent trials and the standard deviation ($\%$) are reported.
}
\label{tab:merged_experiment}
\resizebox{\textwidth}{!}{
{\renewcommand\arraystretch{1.2}\begin{tabular}{ccccccccc}
\hline
 & Local & FedAvg & pFedMe & FedEM & Clustered FL & FedLG & PPFL1 & PPFL2 \\ 
\hline
  Digits     & $83.24_{0.30}$ & $85.63_{0.13}$  & $79.73_{0.33}$  & $85.57_{0.08}$ & $79.21_{0.74}$        & $83.24_{0.37}$ & $86.33_{0.23}$  & $\bm{86.49}_{0.14}$                                        \\  \cline{2-9}
 Sub-MNIST     & $95.93_{0.13}$ & $97.12_{0.10}$  & $95.95_{0.19}$  & $97.30_{0.04}$ & $95.86_{0.03}$        & $96.65_{0.22}$ & $97.33_{0.05}$  & $\bm{97.41}_{0.06}$                                          \\ 
 SVHN & $69.51_{0.80}$ & $75.82_{0.36}$  & $67.58_{0.61}$  & $75.76_{0.30}$ & $64.17_{0.02}$        & $66.77_{0.85}$ & $76.98_{0.36}$  & $\bm{77.20}_{0.17}$                                          \\ 
 USPS & $96.50_{0.11}$ & $97.10_{0.16}$  & $95.50_{0.16}$  & $97.27_{0.11}$ & $96.16_{0.24}$        & $96.06_{0.15}$ & $\bm{97.46}_{0.22}$  & $\bm{97.46}_{0.25}$                                          \\ 
 SYNTH & $84.03_{0.40}$ & $86.27_{0.13}$  & $80.13_{0.43}$  & $85.92_{0.08}$ & $0.80_{0.10}$        & $84.42_{0.45}$ & $ 86.86_{0.26}$  & $\bm{87.02}_{0.17}$                                          \\ 
 MNIST\_M & $82.75_{0.37}$ & $82.12_{0.26}$  & $75.70_{0.73}$  & $83.82_{0.19}$ & $74.5_{0.13}$        & $\bm{84.23}_{0.39}$ & $83.42_{0.34}$  & $83.64_{0.38}$                                          \\ 
\hline
\end{tabular}}}
\end{table}

\subsection{Varying the Number of Canonical Models}
\label{sec:vayring the number of canonical models}

\begin{figure}[t]
\centering
\includegraphics[scale=0.5]{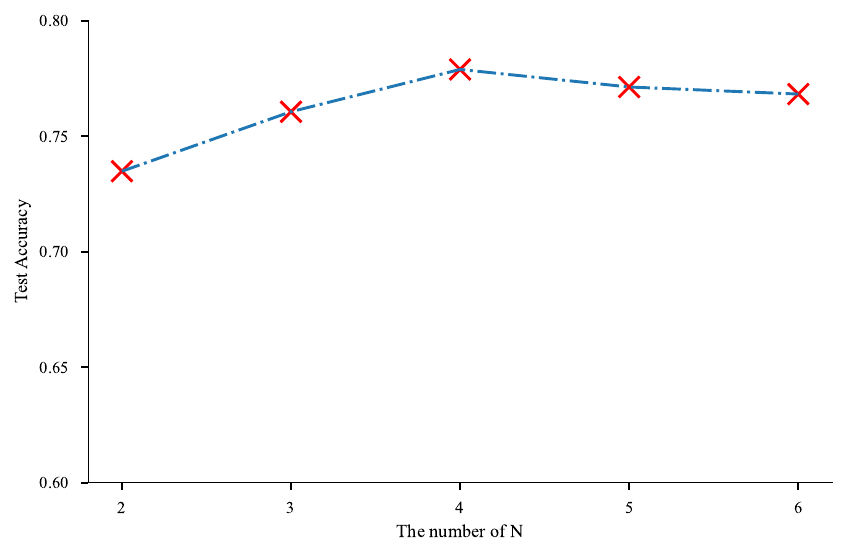}
\caption{Average test accuracy across different values of $N$ on the Synthetic data set.}
\label{fig:N effect}
\end{figure}

\begin{figure}[t]
\centering
{\subfloat[$\alpha=0.2$]{\includegraphics[scale=0.38]{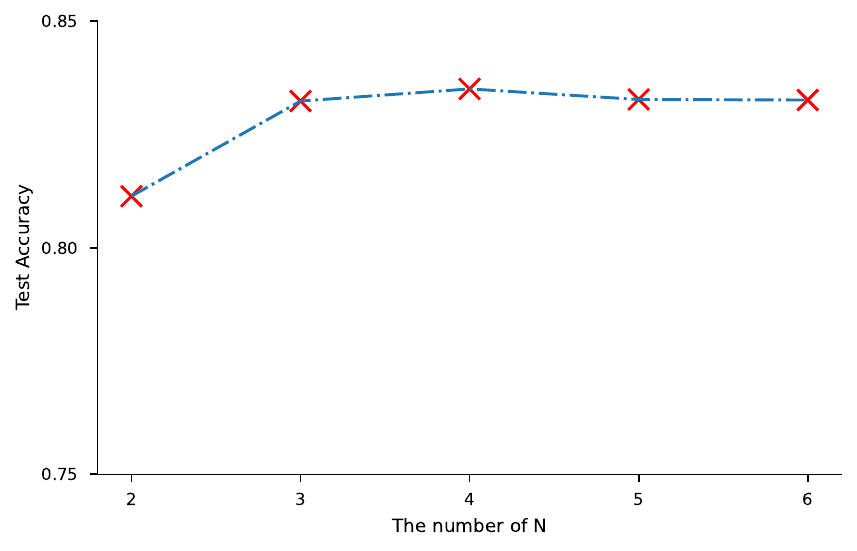}}
\subfloat[$\alpha=0.5$]{\includegraphics[scale=0.38]{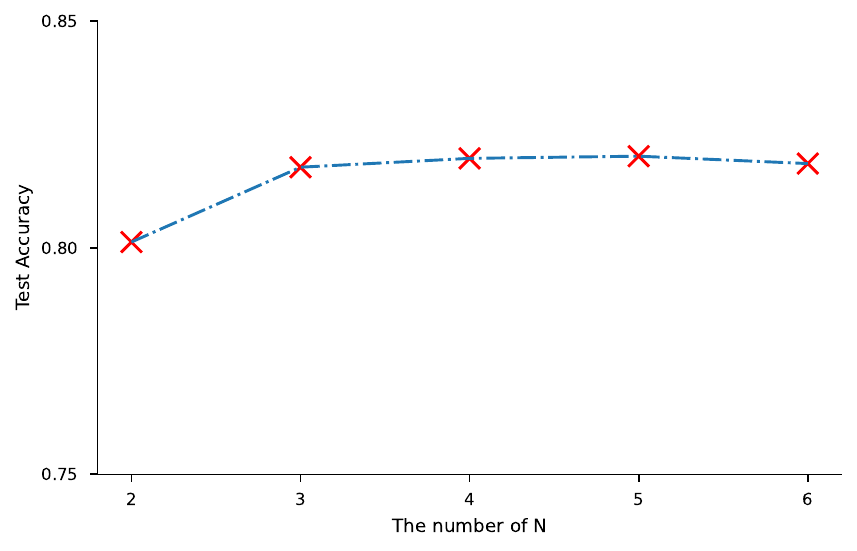}}
\subfloat[$\alpha=1$]{\includegraphics[scale=0.38]{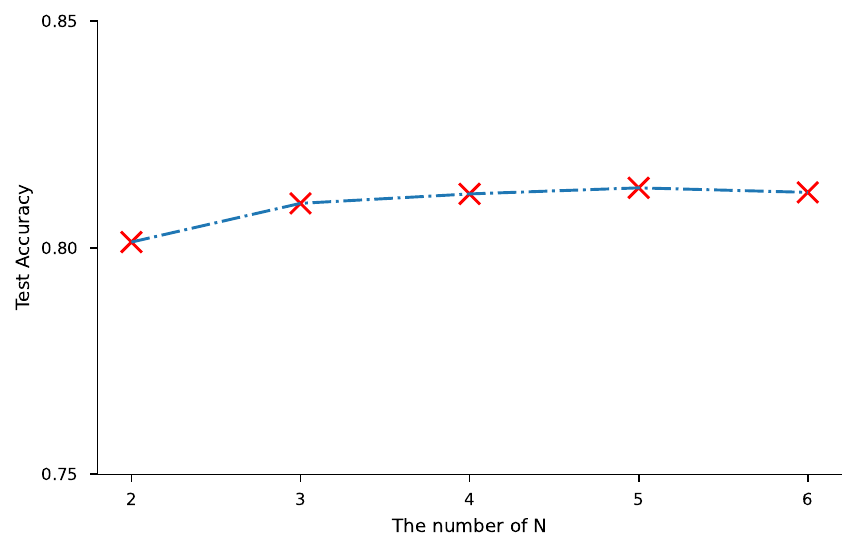}}}
\caption{Average test accuracy across different values of $N$ on CIFAR-10 data set. The optimal numbers of canonical models are $4$, $4$, and $5$, respectively.}
\label{fig:simulation2}
\end{figure}

As shown in Figure \ref{fig:N effect}, we plot the average accuracy against $N$ on the Synthetic data set with the number of latent data sources set to $K=3$.
In this setting, we also find that when $K>N$, the performance of \texttt{PPFL} surpasses that observed when $K\leq N$.
As shown in Figure \ref{fig:simulation2}, we also plot the results of \ours on the CIFAR-10 data set. 
The results also confirm the conclusion in Section 4 in the main text that employing a relatively large number of canonical models could obtain acceptable performance.

\subsection{Effect of the number of local steps}

We explore the performance of \ours across varying the number of local steps (i.e., $E$).
Table \ref{tab:effect-E} presents the performance of \ours under different local step settings, where we progressively increase the number of local steps by a factor of $t$ compared to the original configuration ($1\times E$).
Here, $1\times E$ denotes the original number of local steps used in our previous experiments, $t\times E$ indicates a t-fold increase in local steps compared to $1\times E$, and the centralized setting in Table \ref{tab:effect-E} corresponds to $E=1$.
The results show that increasing $E$ does not consistently lead to improved performance, which aligns with the analysis of \citet{li2019convergence}.
Nonetheless, all tested configurations outperform the centralized version.
Moreover, the optimal number of local steps $E$ varies by data sets, with more complex tasks often requiring larger values of $E$.
For instance, in more complex data sets like CIFAR-10 and Synthetic, increasing $E$ (e.g., $2\times E$ and $3\times E$) improves the performance of \ours.
In contrast, for the simpler data set MNIST, a larger $E$ (e.g., $2\times E$) actually leads to a decline in performance.

\begin{table}[bt]
\centering
\caption{Comparison of \ours with different values of local steps $E$.}
\label{tab:effect-E}
{\begin{tabular}{lccc}
\hline
        & MNIST & CIFAR-10 & Synthetic \\
\hline
Central & 98.01 & 81.29   & 71.61     \\
\hline
$1\times \text{E}$      & 99.01 & 85.33   & 77.60     \\
$2\times \text{E}$     & 98.97 & 86.06   & 78.52     \\
$3\times \text{E}$     & 98.92 & 86.21   & 78.38     \\
$4\times \text{E}$    & 98.92 & 85.84   & 78.22    \\
\hline
\end{tabular}}
\end{table}

\subsection{Train Accuracy on Practical Data Set}
\label{subsec:train accuracy}

\begin{figure}[t]
\centering
\caption{Average Train Accuracy $\&$ Communication Cost. The average train accuracy on (a) MNIST data set, (b) CIFAR-10 data set, and (c) Synthetic data set.}
\subfloat[MNIST]{\label{fig:train-mnist}\includegraphics[scale=0.34]{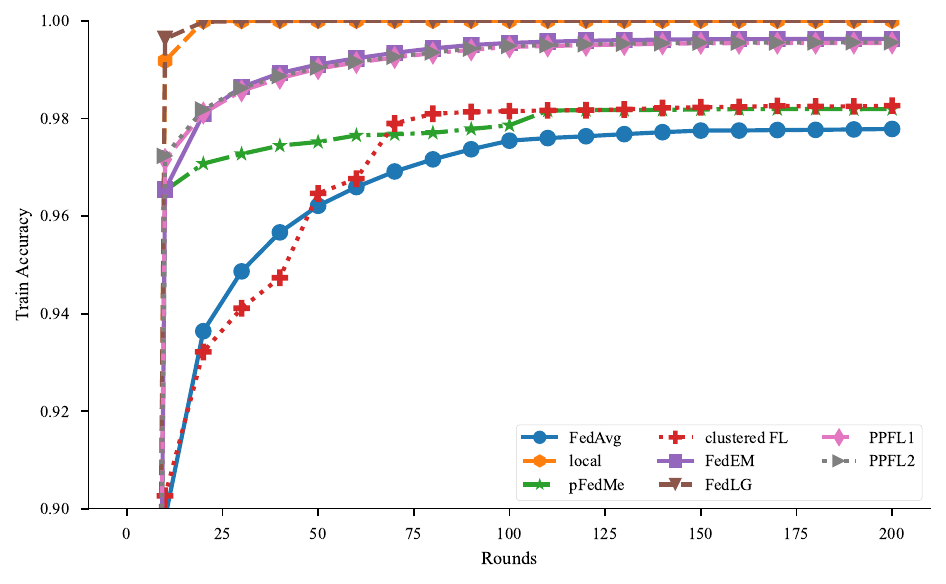}}
\subfloat[CIFAR-10]{\label{fig:train-cifar}\includegraphics[scale=0.34]{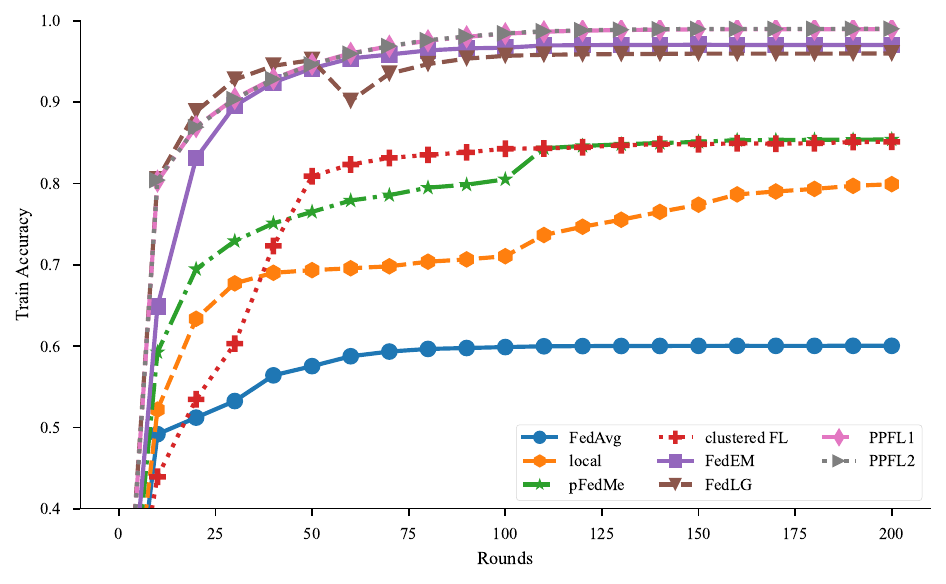}}
\subfloat[Synthetic]{\label{fig:train-synthetic}\includegraphics[scale=0.34]{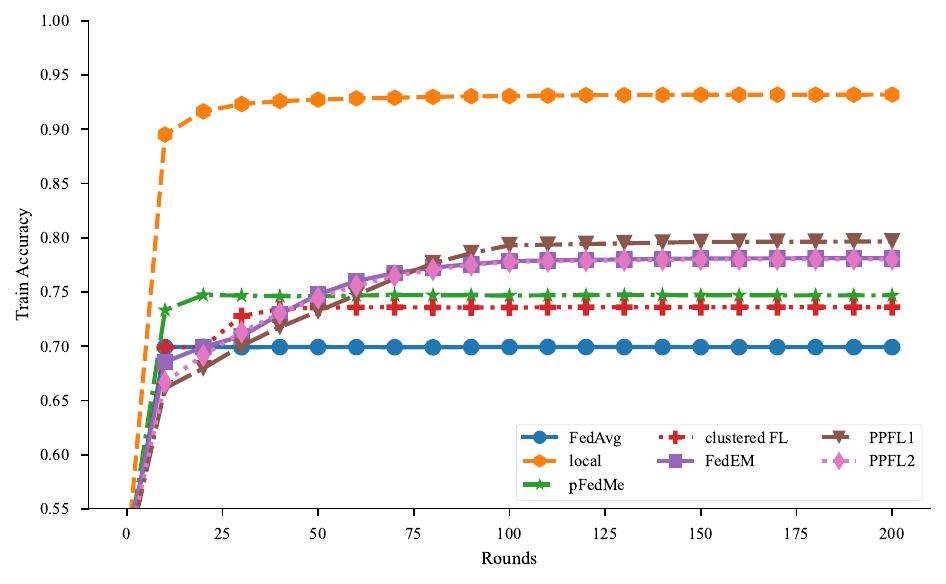}}
\label{fig:train curve of manipulated data sets}
\end{figure}

In addition to test accuracy, we also include the training accuracy curves for the pathological data sets.
As shown in Figures \ref{fig:train-mnist} and \ref{fig:train-cifar}, the proposed method \ours achieves faster convergence compared to other baselines, highlighting its advantage in heterogeneous settings.
On the Synthetic dataset, as shown in Figure \ref{fig:train-synthetic}, \ours attains the second-highest training accuracy (lower than \texttt{FedEM}), whereas \texttt{FedEM} is overfitting, as evidenced by its degraded test set accuracy; this contrast suggests that \ours generalizes more effectively in settings with mixed client distributions.

\subsection{Difference Between \ppflone and \ppfltwo}
\label{sec:variants difference}
Specifically, when the architecture of canonical models can capture the heterogeneity across clients, our insight is that these canonical models tend to lie in distant basins in the loss landscape under conditions of high heterogeneity.
In this high-heterogeneity scenario, permutations in the membership vector $\rvc_i$ within the parameter space can lead to substantial shifts in function values and predictions for \ppfltwo, which utilizes interpolation to construct personalized models.
This sensitivity may result in inferior performance for \ppfltwo, while \ppflone may perform better due to its resilience to such permutations.
Conversely, lower heterogeneity can bring the basins of canonical models closer together, and these models may even share a single basin in the loss landscape. 
Under conditions of low heterogeneity, the interpolation approach may also enhance generalization by leveraging the proximity of these basins, allowing \ppfltwo to outperform \ppflone.
These insights are further validated by our experiments. 
In the highly heterogeneous setting shown in Table 1, \ppflone demonstrates superior performance to \ppfltwo. Similarly, Table 2 shows that \ppflone performs better under high heterogeneity, while \ppfltwo achieves better results when heterogeneity is low.
Besides, when the canonical model's structure fails to capture the clients' heterogeneity, the resulting loss landscape becomes less distinct, which may reduce the effectiveness of the interpolation approach (i.e., \ppfltwo). 
For example, in the StackOverflow dataset in Table 3, where we use the last embedding layer as the canonical model, \ppfltwo performs worse than \ppflone, likely due to the unexplored heterogeneity that this module fails to capture.
Therefore, when the structure of the canonical model can capture the heterogeneity among clients, \ppflone may be preferable for high heterogeneity scenarios, while \ppfltwo may yield better performance in low heterogeneity settings. In contrast, when it is unclear whether the model structure can effectively capture heterogeneity, \ppflone might offer greater stability due to its less distinct loss landscape.

\end{document}